\documentclass[11pt]{article}
\usepackage{sustyle}

\usepackage[utf8]{inputenc} 
\usepackage[T1]{fontenc}    
\usepackage[hidelinks]{hyperref}
\usepackage{url}            
\usepackage{booktabs}       
\usepackage{amsfonts}       
\usepackage{nicefrac}       
\usepackage{microtype}      
\usepackage{xcolor}         

\usepackage{bigstrut}

\usepackage[export]{adjustbox}
\usepackage{subcaption}
\usepackage{graphicx}
\usepackage{makecell}
\usepackage{amsmath}
\usepackage{url}

\title{A Law of Next-Token Prediction in Large Language Models}

\author{Hangfeng He\thanks{University of Rochester. Email: \texttt{hangfeng.he@rochester.edu}.} \and Weijie J.~Su\thanks{University of Pennsylvania. Email: \texttt{suw@wharton.upenn.edu}.}\\[0.5em]}

\date{August 2024; Revised August 2025}

\begin{document}

\maketitle

\begin{abstract}

Large language models (LLMs) have been widely employed across various application domains, yet their black-box nature poses significant challenges to understanding how these models process input data internally to make predictions. In this paper, we introduce a precise and quantitative law that governs the learning of contextualized token embeddings through intermediate layers in pre-trained LLMs for next-token prediction. Our findings reveal that each layer contributes equally to enhancing prediction accuracy, from the lowest to the highest layer---a universal phenomenon observed across a diverse array of open-source LLMs, irrespective of their architectures or pre-training data. We demonstrate that this law offers new perspectives and actionable insights to inform and guide practices in LLM development and applications, including model scaling, pre-training tasks, and interpretation.

\end{abstract}

\renewcommand{\figurename}{Fig.}

\section{Introduction}
\label{sec:introdu-}

The rapid advancement of large language models (LLMs) has profoundly impacted various fields, including mathematical discovery \cite{romera2024mathematical}, medical diagnosis \cite{clusmann2023future}, genomic research \cite{huang2024crispr,hao2024large}, and education \cite{Tu2024What}. Despite their transformative and widespread adoption, the deployment of LLMs is often impeded by a lack of understanding of how these enormous, complex black-box models internally process data to generate predictions \cite{radhakrishnan2024mechanism}. Without understanding the prediction mechanisms, practitioners face challenges in interpreting these predictions for decision-making. For LLM developers, this lack of transparency hinders the development of general and robust design principles for LLMs. Consequently, these challenges constrain the full realization of the potential offered by LLM methodologies.

The primary difficulty arises from the hierarchical nature of LLM architectures, such as Transformer \cite{vaswani2017attention}, RWKV \cite{peng2023rwkv}, and Mamba \cite{gu2024mamba}. These architectures are composed of multiple layers, each corresponding to simple functions such as linear or quadratic transformations---better known as the attention mechanism---or a nonlinear combination of both. While the input and output can be observed and specified, the internal transformation of data by each layer becomes elusive due to the hierarchical composition. Specifically, in the case of generative pre-trained transformers (GPT) \cite{radford2018improving}, it is unclear how the embeddings of the input text are progressively transformed into features across different layers for next-token prediction, where a token refers to a word or subword. This generative nature of LLMs introduces additional complexity compared to traditional classification tasks in multilayer perceptrons (MLP): the vocabulary size typically exceeds the embedding dimension, and models undergo relatively few training epochs. The goal of this paper is to shed light on the inner workings of LLMs as they process token embeddings across all layers. In particular, we aim to identify universal and quantitative patterns that can provide useful principles and insights to refine training processes and enhance interpretability in LLMs.

\begin{figure*}[!htp]
        \captionsetup[subfigure]{labelformat=empty}
		\centering
		 	\subfloat[GPT-1]{
			\centering
		\includegraphics[scale=0.24]{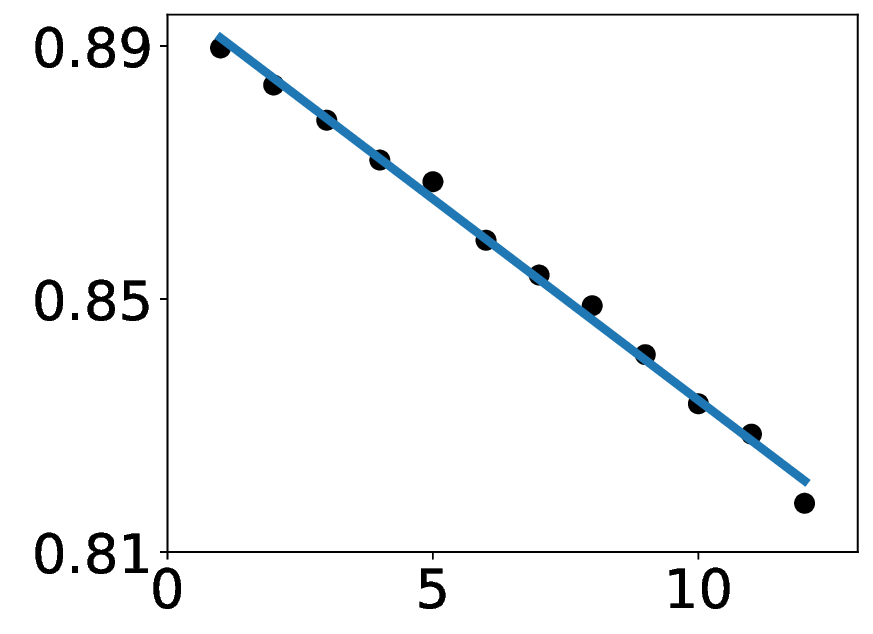}}
			\label{fig:openai-gpt}\hfill
        	\subfloat[GPT-2 XL]{
			\centering
		\includegraphics[scale=0.24]{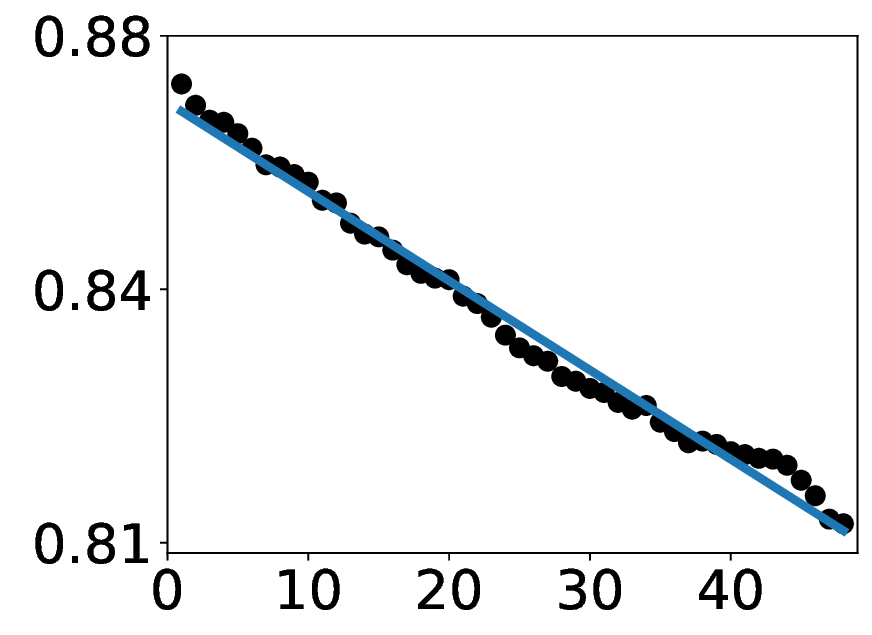}}
			\label{fig:gpt2-xl}\hfill
     	\subfloat[Llama-1-13B]{
			\centering
		\includegraphics[scale=0.24]{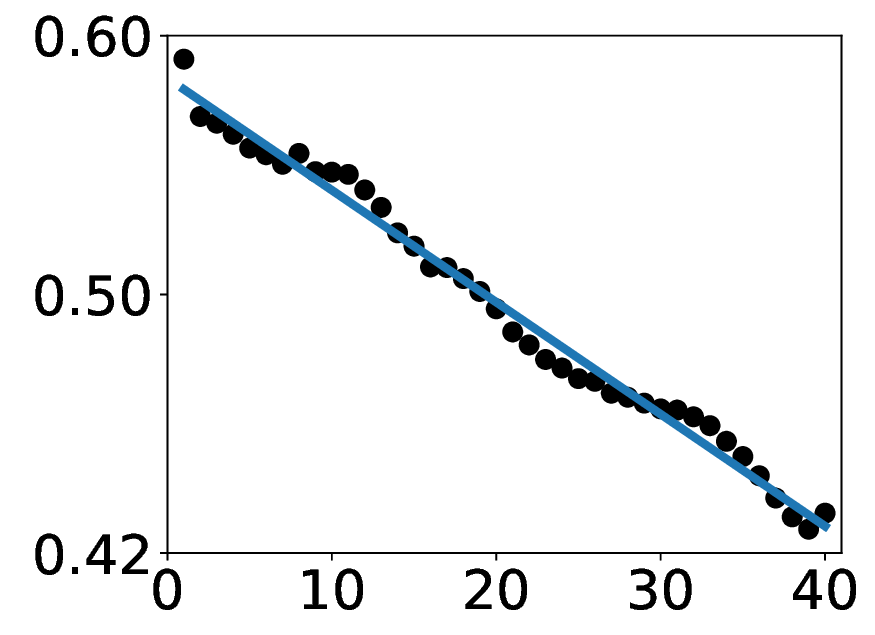}}
			\label{fig:llama-13b}\hfill
		    \subfloat[Llama-2-13B]{
			\centering
	   \includegraphics[scale=0.24]{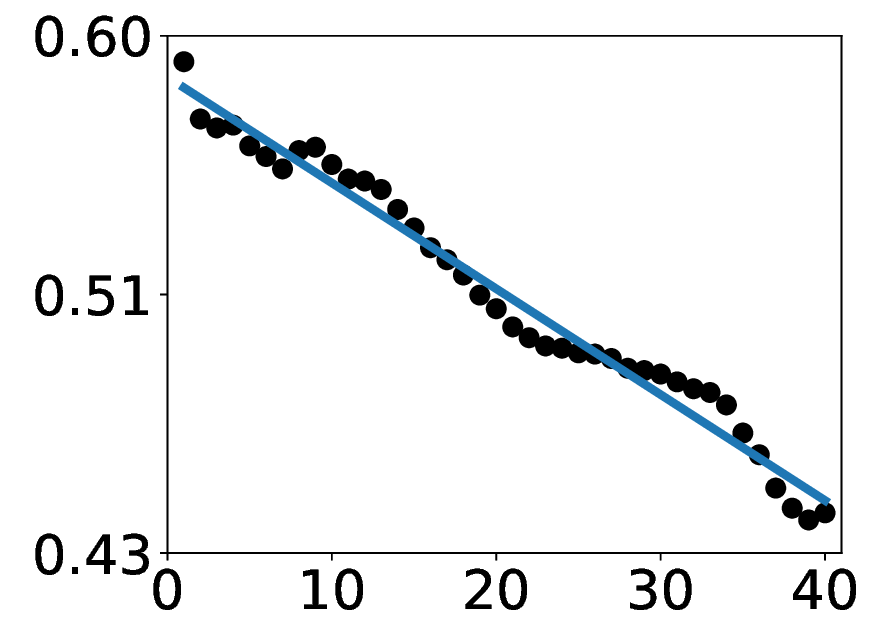}}
			\label{fig:Llama-2-13b}\hfill
   
        	\subfloat[Llama-2-13B-Chat]{
			\centering
	   \includegraphics[scale=0.24]{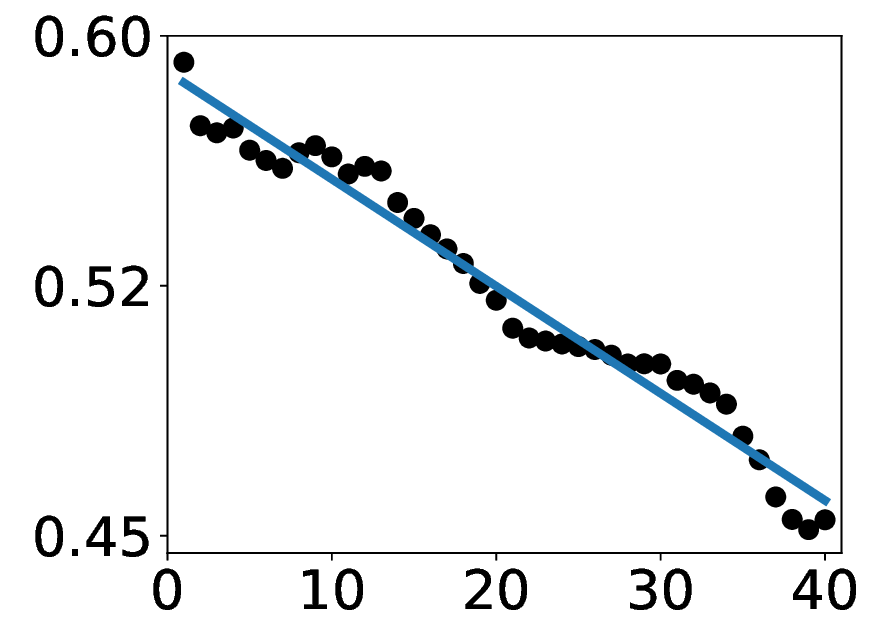}}
			\label{fig:Llama-2-13b-chat}\hfill
     	\subfloat[Llama-3-8B]{
			\centering
		\includegraphics[scale=0.24]{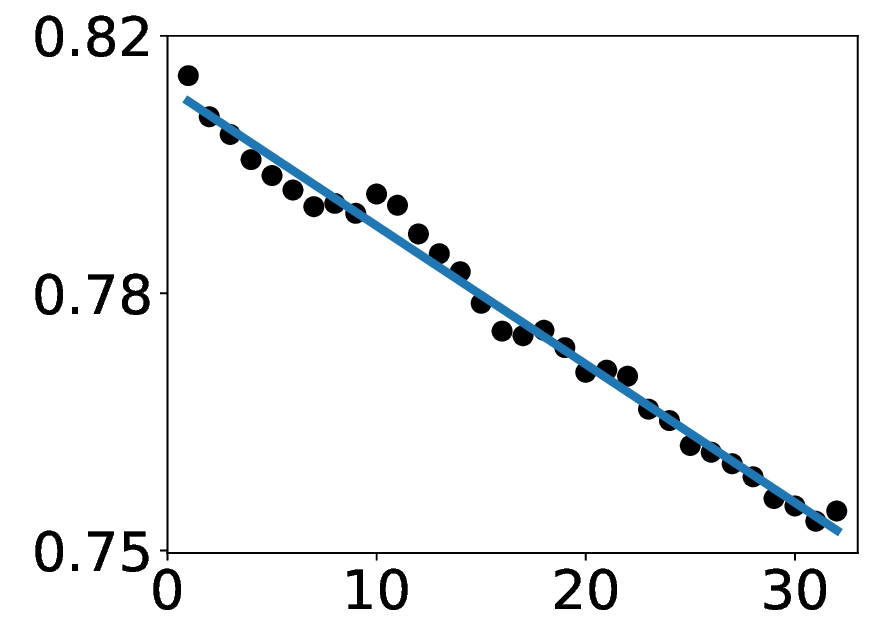}}
			\label{fig:Llama-3-8B}\hfill
   \subfloat[Llama-3-8B-Instruct]{
			\centering
		\includegraphics[scale=0.24]{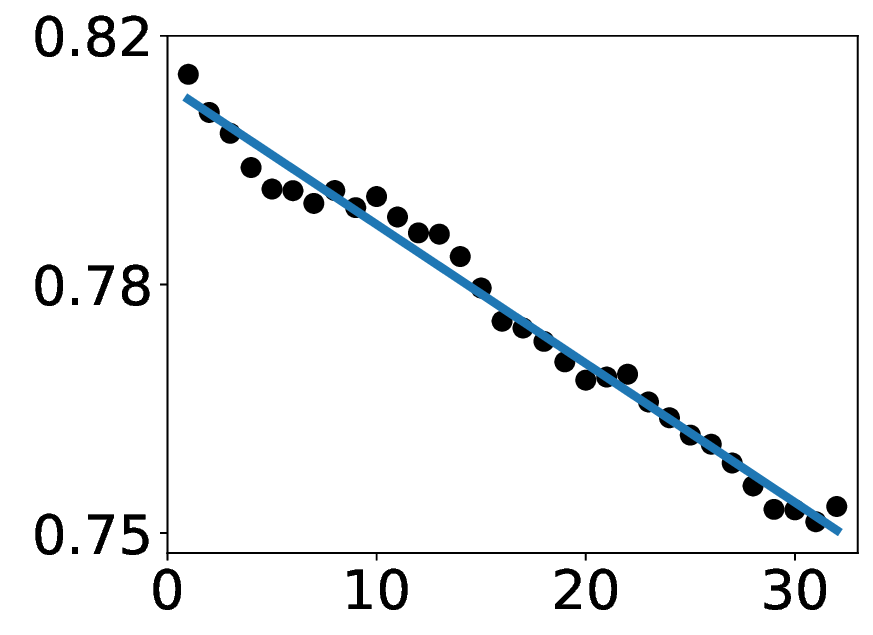}}
			\label{fig:Llama-3-8b-Instruct}\hfill
        	\subfloat[Mistral-7B-v0.1]{
			\centering
		\includegraphics[scale=0.24]{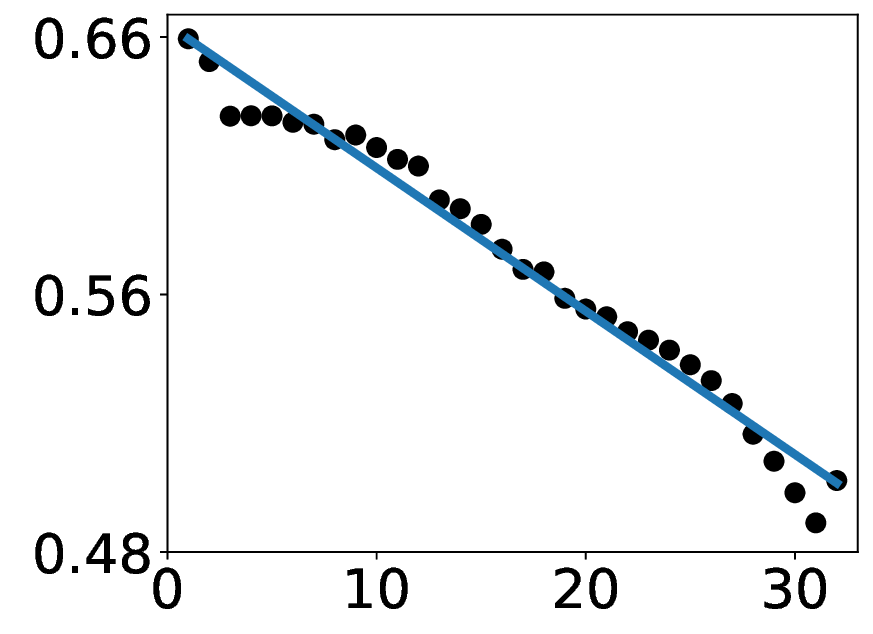}}
			\label{fig:Mistral-7B-v0.1}\hfill
   
     	\subfloat[Mistral-7B-Instruct-v0.1]{
			\centering
		\includegraphics[scale=0.24]{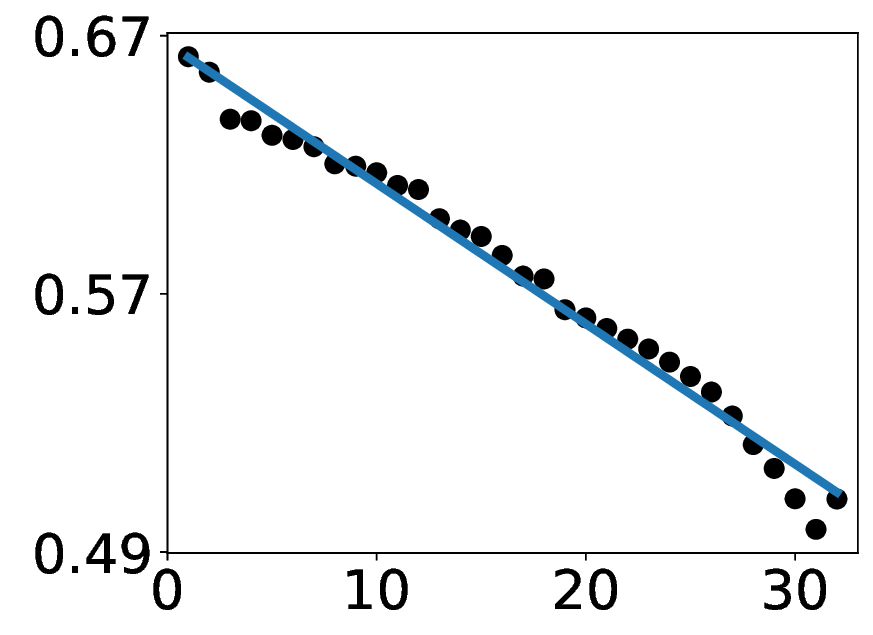}}
			\label{fig:Mistral-7B-Instruct-v0.1}\hfill
		    \subfloat[Mistral-7B-v0.2]{
			\centering
		\includegraphics[scale=0.24]{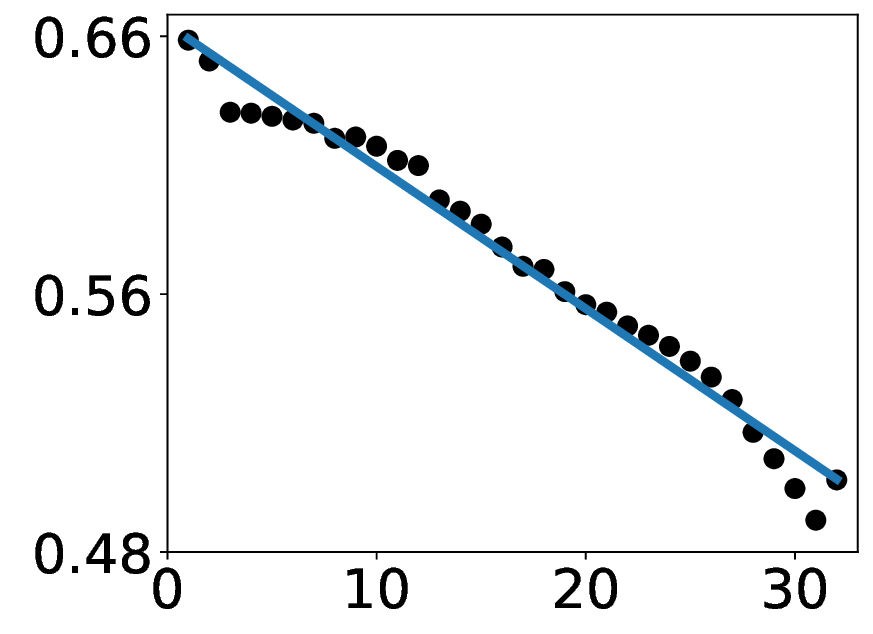}}
			\label{fig:Mistral-7B-v0.2}\hfill
        	\subfloat[Mistral-7B-Instruct-v0.2]{
			\centering
		\includegraphics[scale=0.24]{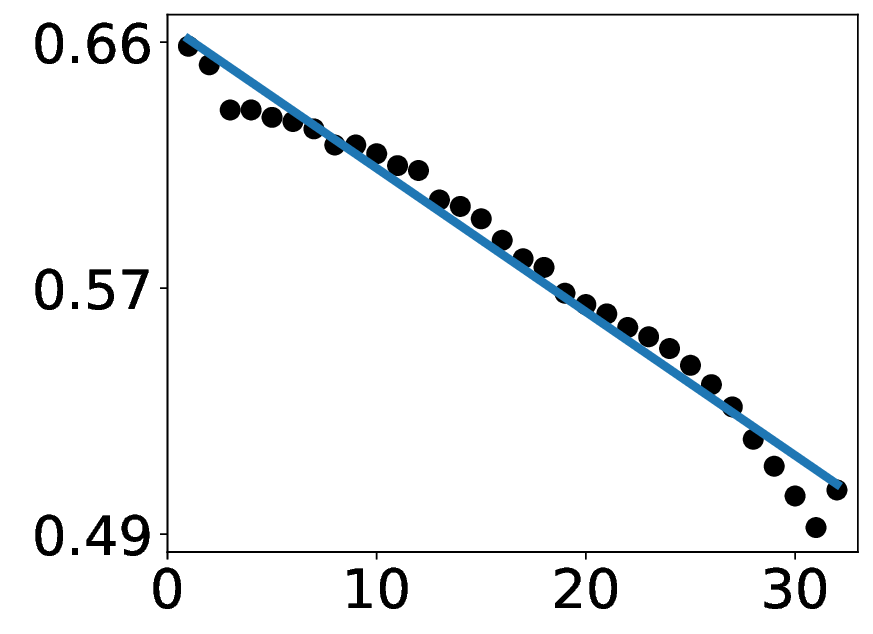}}
			\label{fig:Mistral-7B-Instruct-v0.2}\hfill
            \subfloat[Mistral-7B-v0.3]{
			\centering
		\includegraphics[scale=0.24]{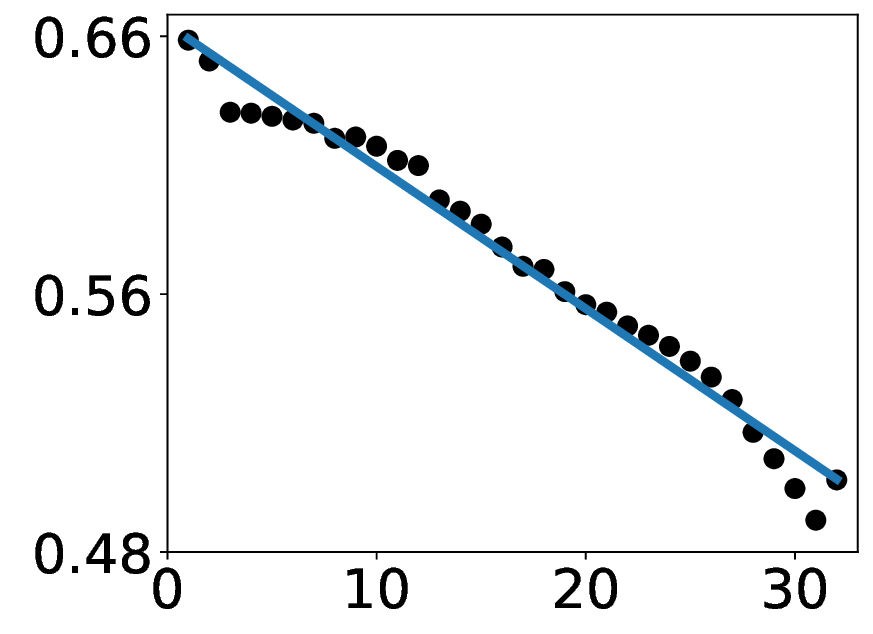}}
			\label{fig:Mistral-7B-v0.3}\hfill
   
        	\subfloat[Mistral-7B-Instruct-v0.3]{
			\centering
		\includegraphics[scale=0.24]{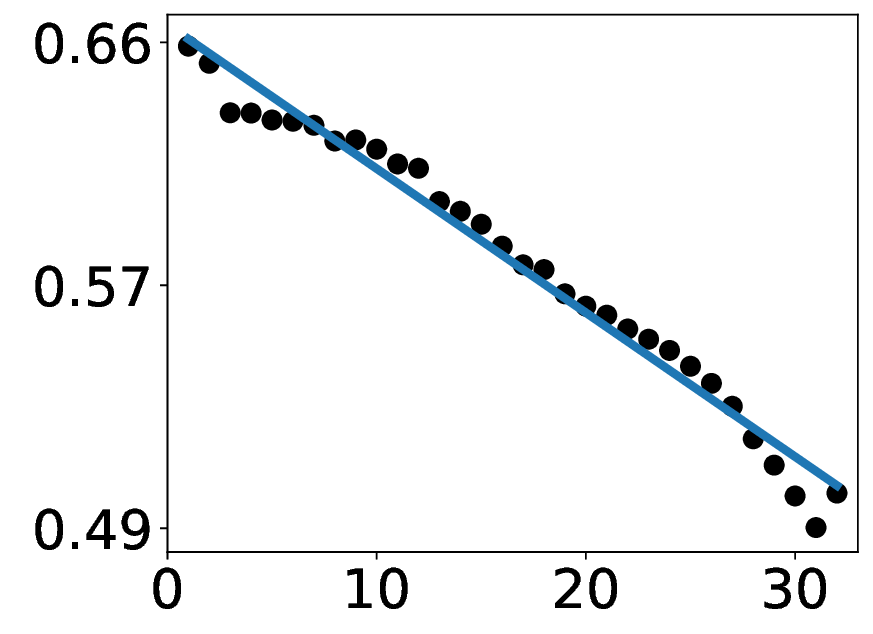}}
			\label{fig:Mistral-7B-Instruct-v0.3}\hfill
     	\subfloat[phi-1.5]{
			\centering
		\includegraphics[scale=0.24]{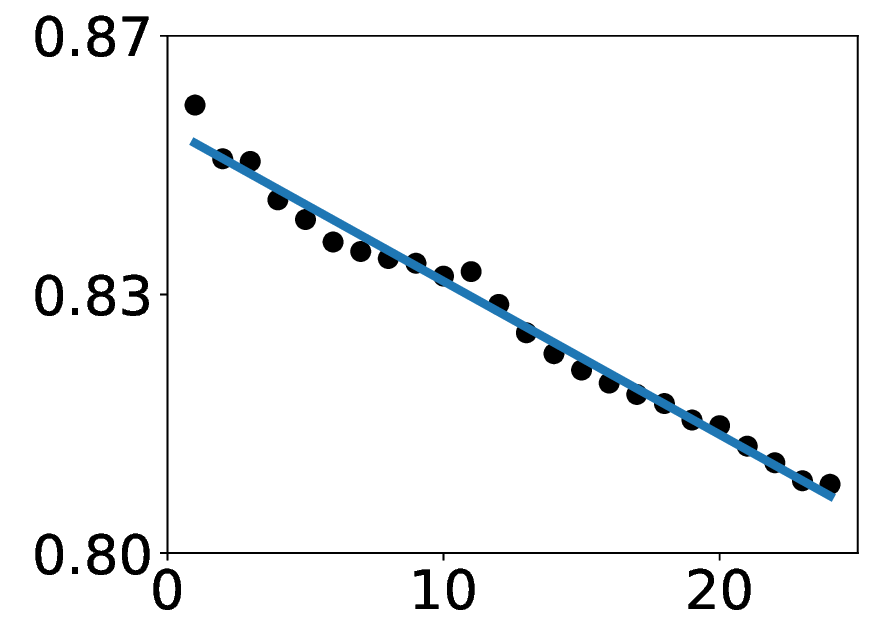}}
			\label{fig:phi-1.5}\hfill
		    \subfloat[phi-2]{
			\centering
	   \includegraphics[scale=0.24]{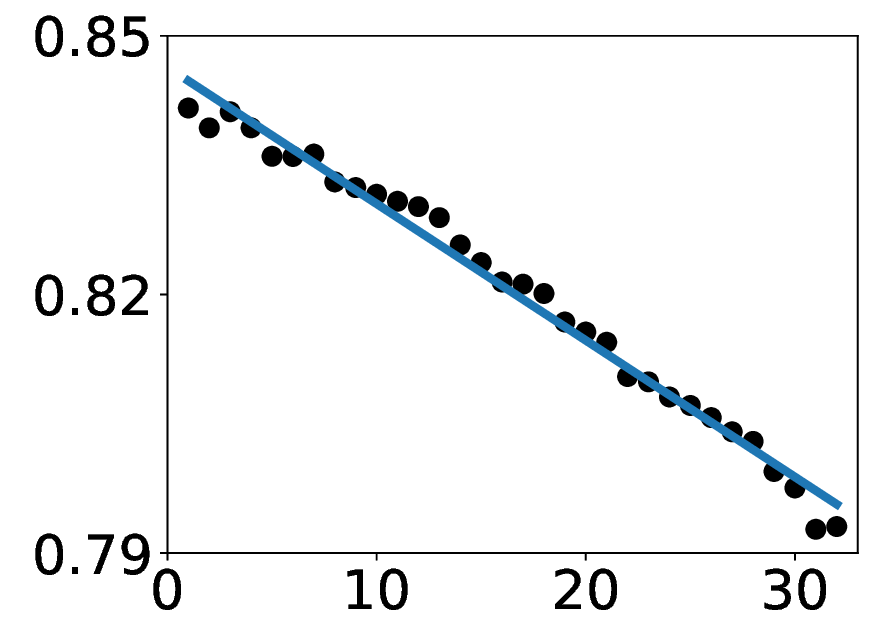}}
			\label{fig:phi-2}\hfill
        	\subfloat[
phi-3-medium-4k-instruct]{
			\centering
	   \includegraphics[scale=0.24]{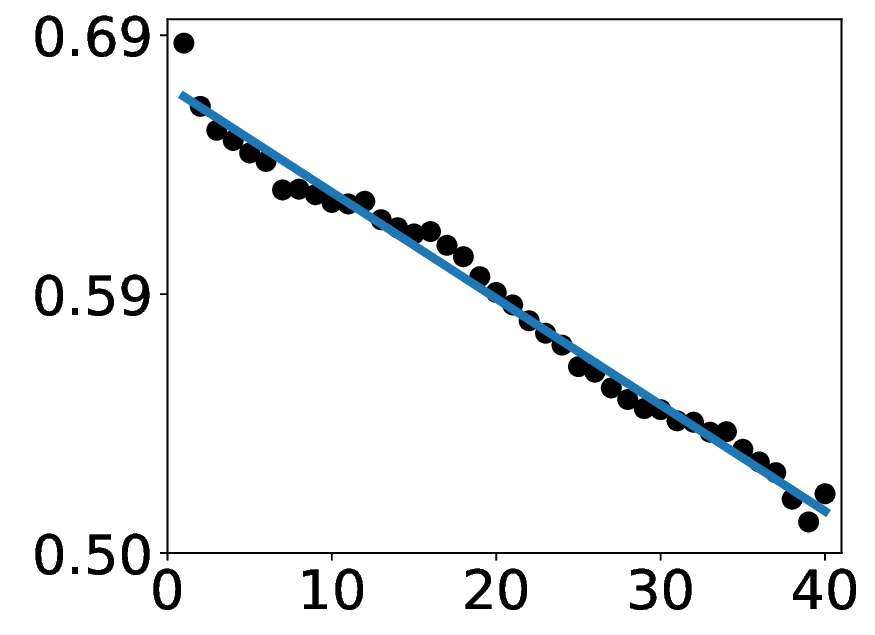}}
			\label{fig:Phi-3-medium-4k-instruct}\hfill
   
     	\subfloat[phi-3-medium-128k-instruct]{
			\centering
		\includegraphics[scale=0.24]{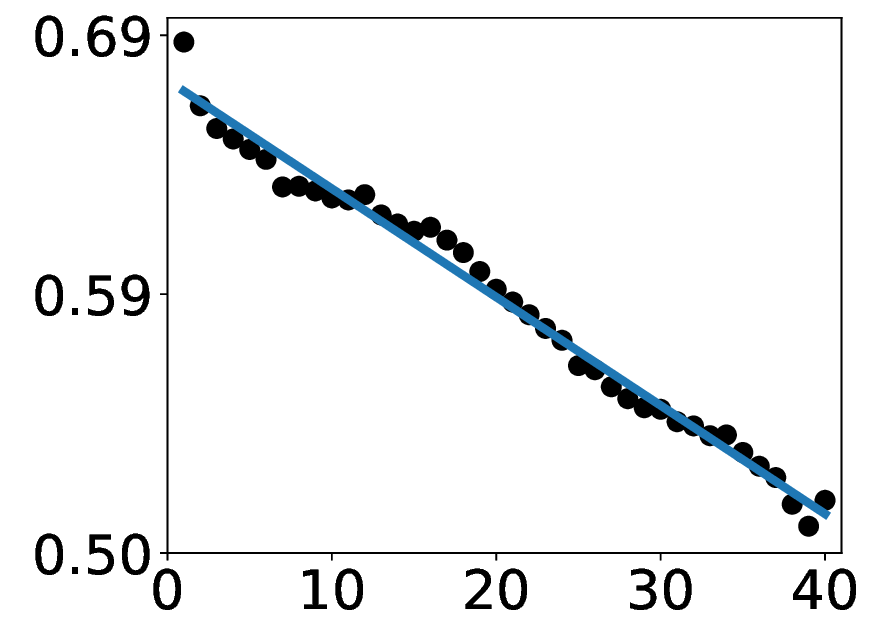}}
			\label{fig:Phi-3-medium-128k-instruct}\hfill
            \subfloat[RWKV-14B]{
			\centering
		\includegraphics[scale=0.24]{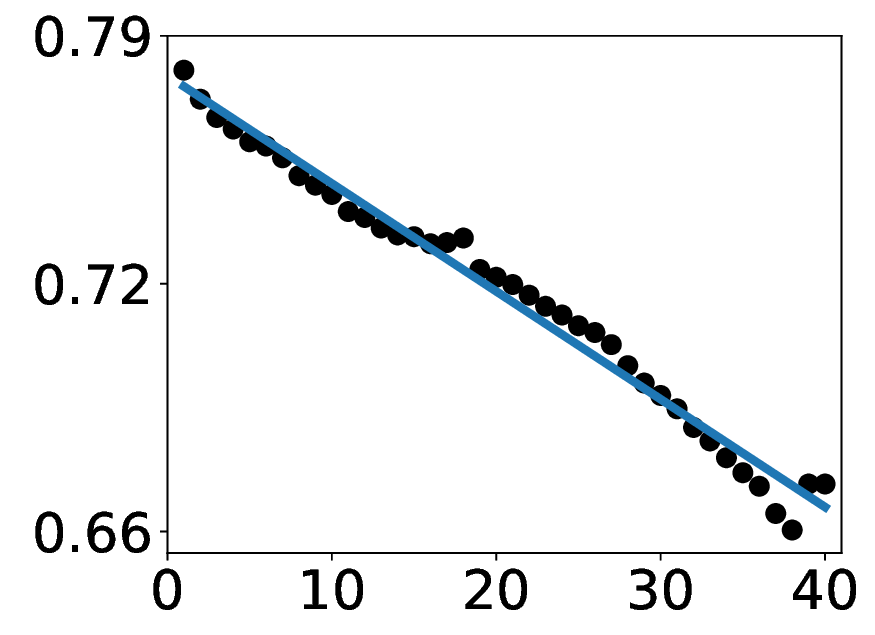}}
            \label{fig:rwkv-14b}\hfill
            \subfloat[RWKV-Raven-14B]{
			\centering
		\includegraphics[scale=0.24]{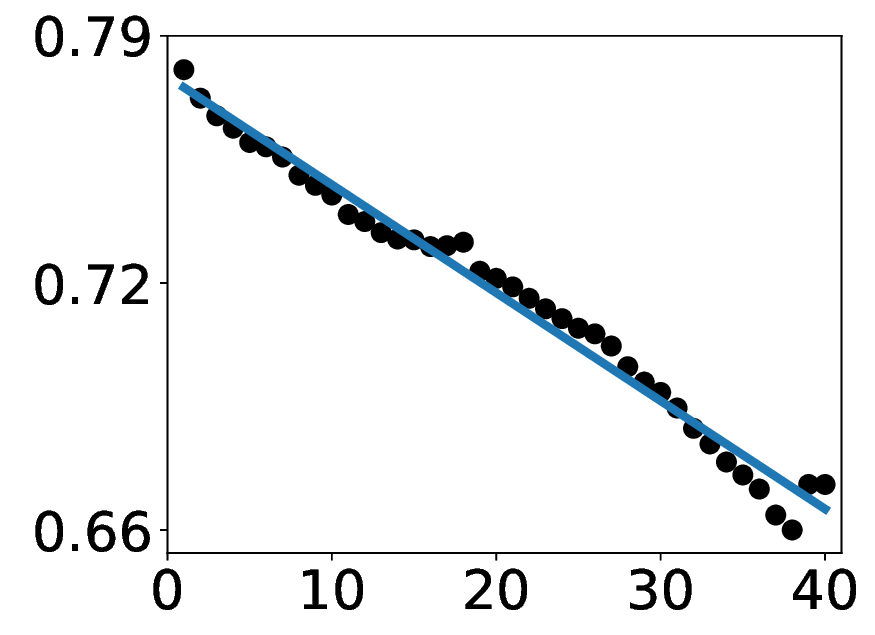}}
            \label{fig:rwkv-raven-14b}\hfill
             \subfloat[Mamba-2.8B]{
			\centering
		\includegraphics[scale=0.24]{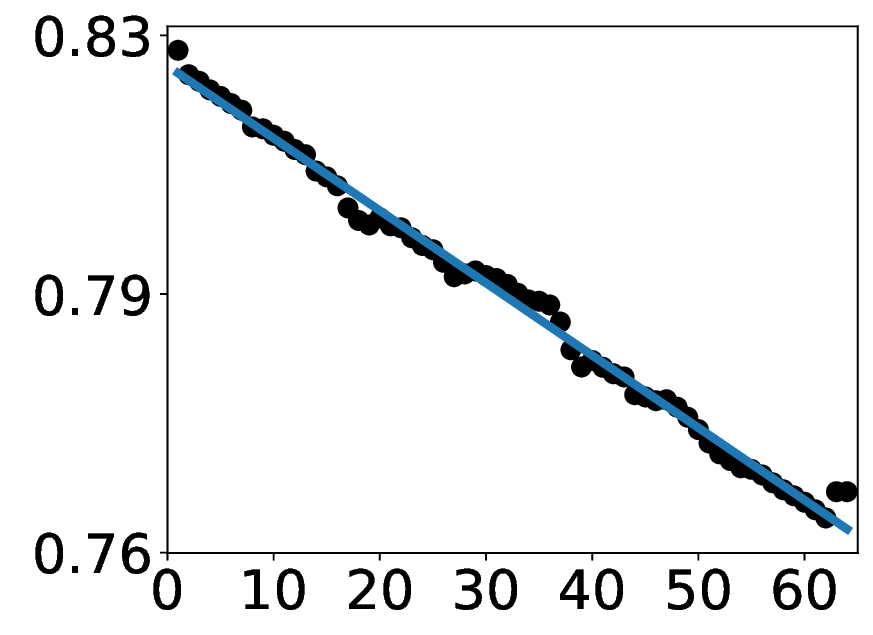}}
            \label{fig:mamba-2.8b}
		\caption{The law of equi-learning in large language models. Throughout the paper, the x axis represents the layer index, and the y axis, on a logarithmic scale, represents the prediction residual (PR) defined in Eq.~\ref{eq:PR}, unless otherwise specified. The Pearson correlation coeﬀicients, by row first, are -0.997, -0.994, -0.994, -0.988; -0.983, -0.993, -0.992, -0.988; -0.991, -0.989, -0.988, -0.989; -0.988, -0.994, -0.993, -0.992; -0.992, -0.991, -0.991, -0.997. More details can be found in the Supplementary Materials.
		}
		\label{fig:law}
\end{figure*}

\begin{figure*}[!t]
    \captionsetup[subfigure]{labelformat=empty}
		\centering
            \subfloat[Layer=1]{
			\centering
	\includegraphics[scale=0.24]{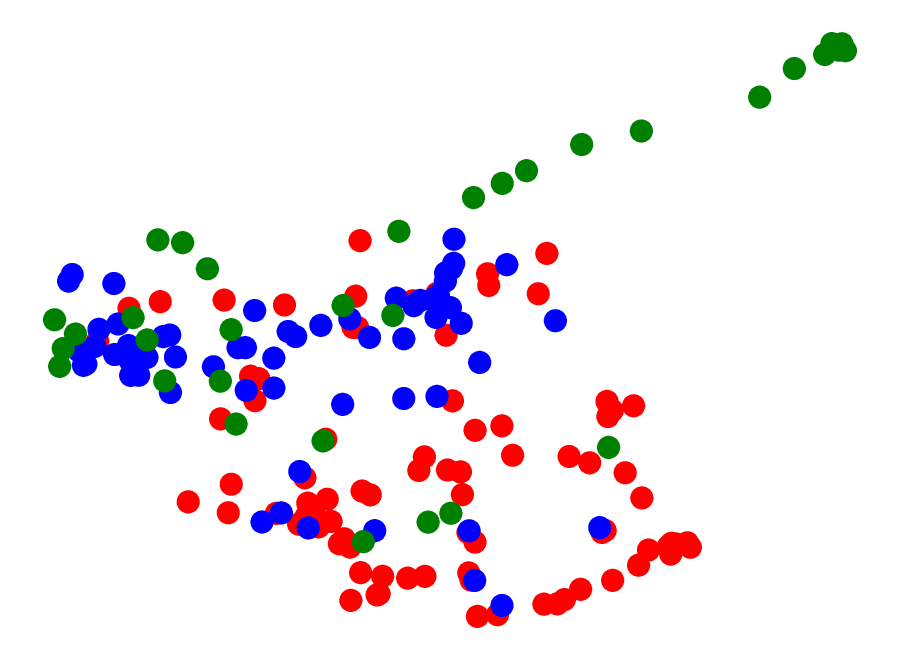}}
            \label{fig:medicine-layer1}\hfill
     	\subfloat[Layer=2]{
			\centering
	\includegraphics[scale=0.24]{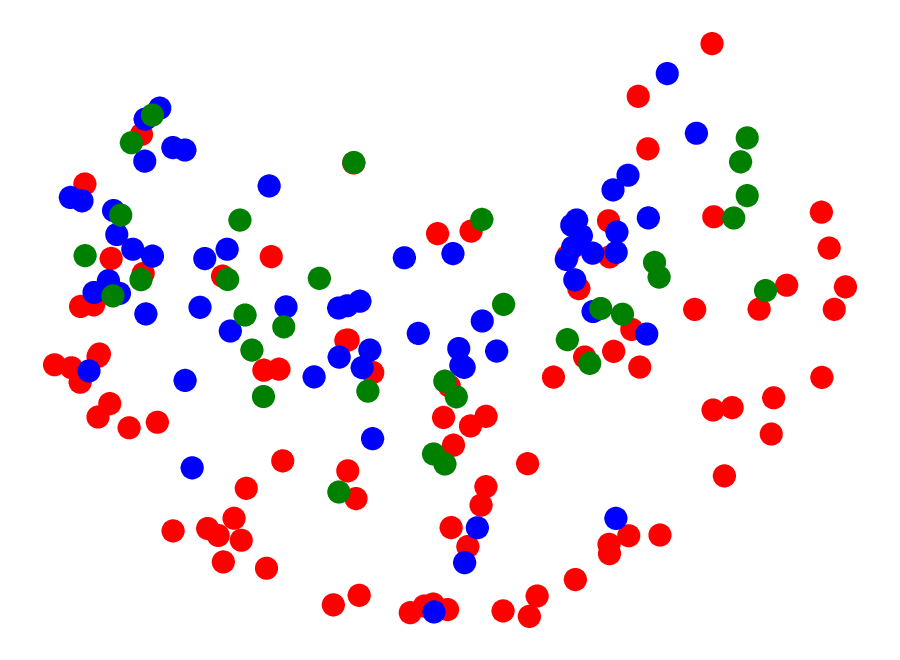}}
            \label{fig:medicine-layer2}\hfill
        \subfloat[Layer=3]{
			\centering
	\includegraphics[scale=0.24]{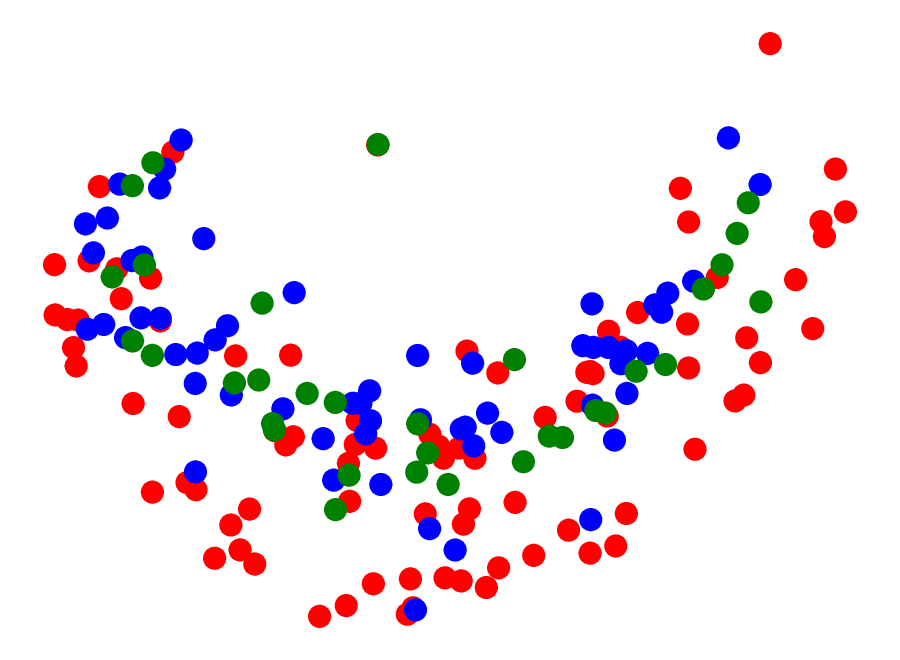}}
            \label{fig:medicine-layer3}\hfill
            \subfloat[Layer=4]{
			\centering
	\includegraphics[scale=0.24]{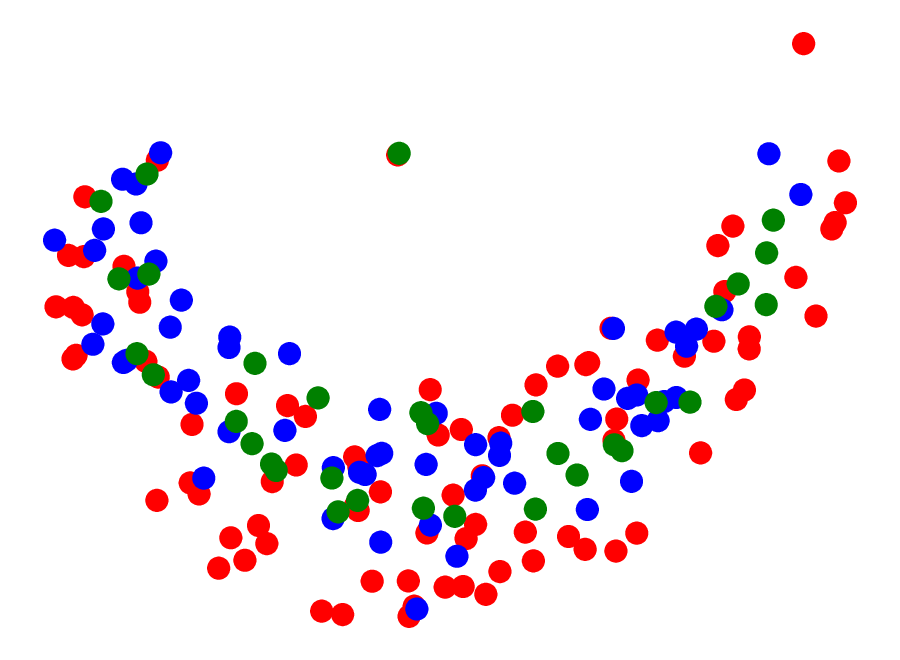}}
            \label{fig:medicine-layer4}
            
     	\subfloat[Layer=5]{
			\centering
	\includegraphics[scale=0.24]{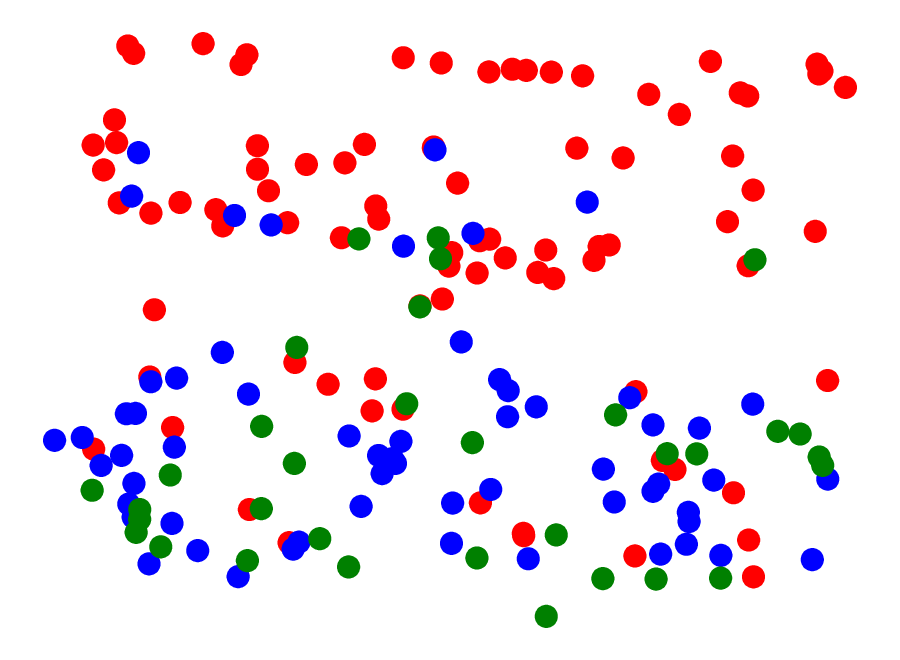}}
            \label{fig:medicine-layer5}\hfill
            \subfloat[Layer=6]{
			\centering
	\includegraphics[scale=0.24]{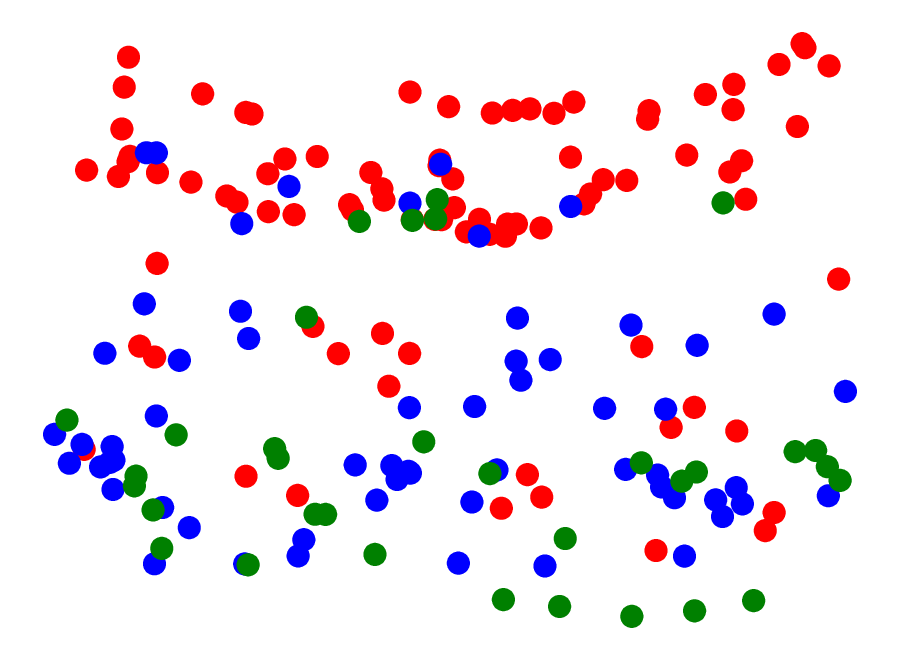}}
            \label{fig:medicine-layer6}\hfill
        \subfloat[Layer=7]{
			\centering
	\includegraphics[scale=0.24]{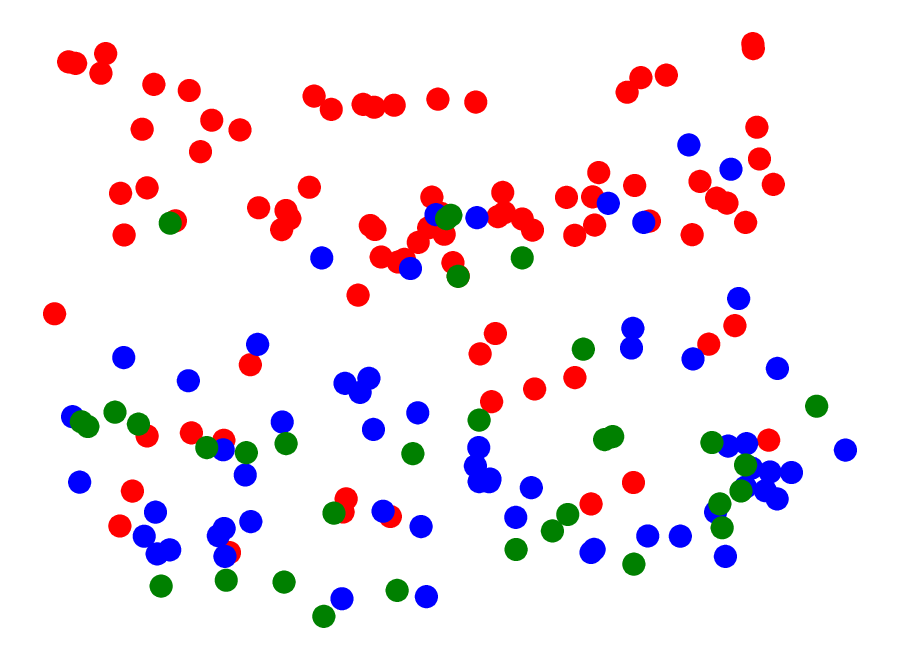}}
            \label{fig:medicine-layer7}\hfill   
        \subfloat[Layer=8]{
			\centering
	\includegraphics[scale=0.24]{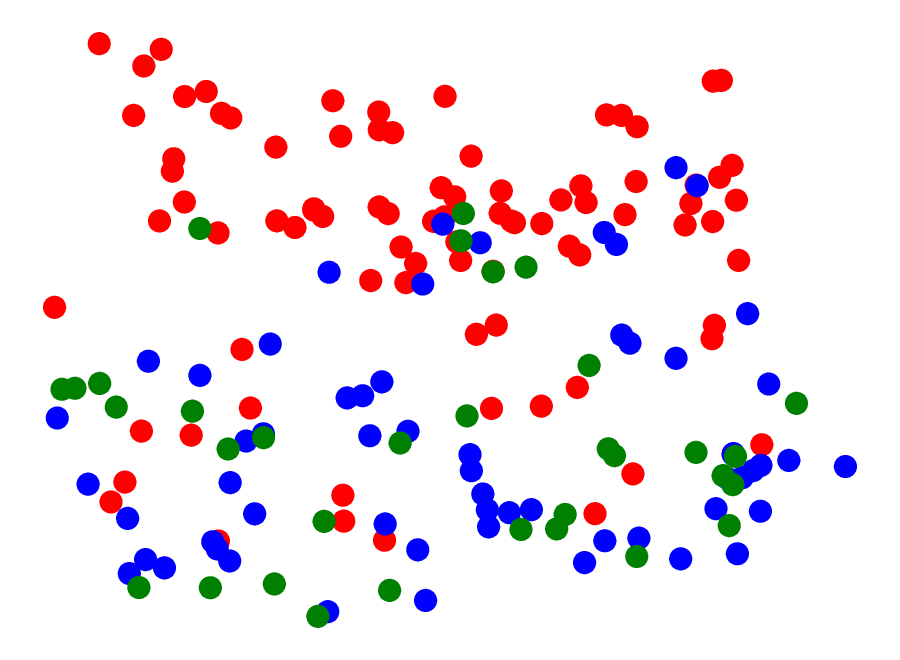}}
            \label{fig:medicine-layer8}
            
        \subfloat[Layer=9]{
			\centering
	\includegraphics[scale=0.24]{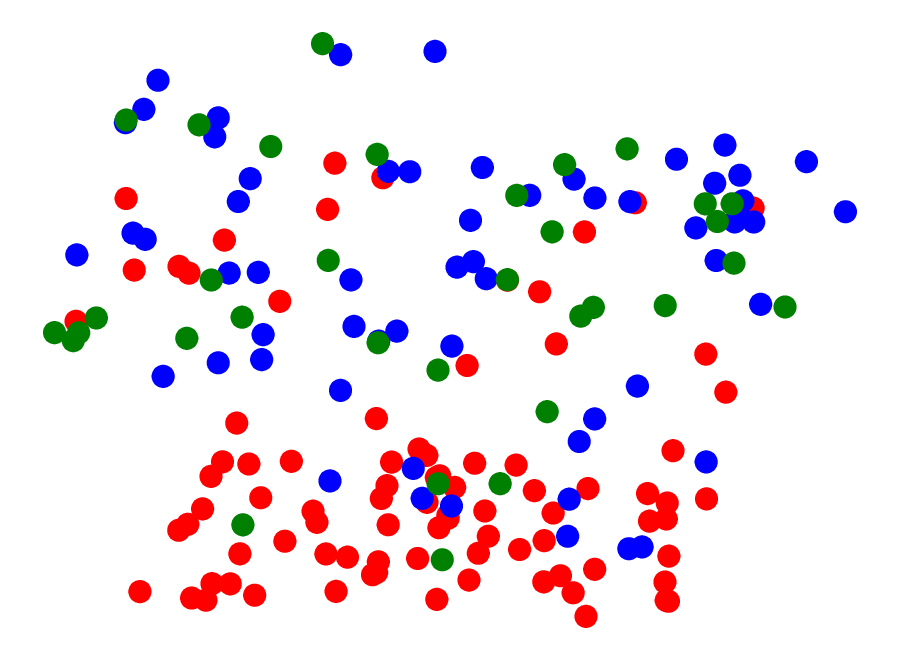}}
            \label{fig:medicine-layer9}\hfill
        \subfloat[Layer=10]{
			\centering
	\includegraphics[scale=0.24]{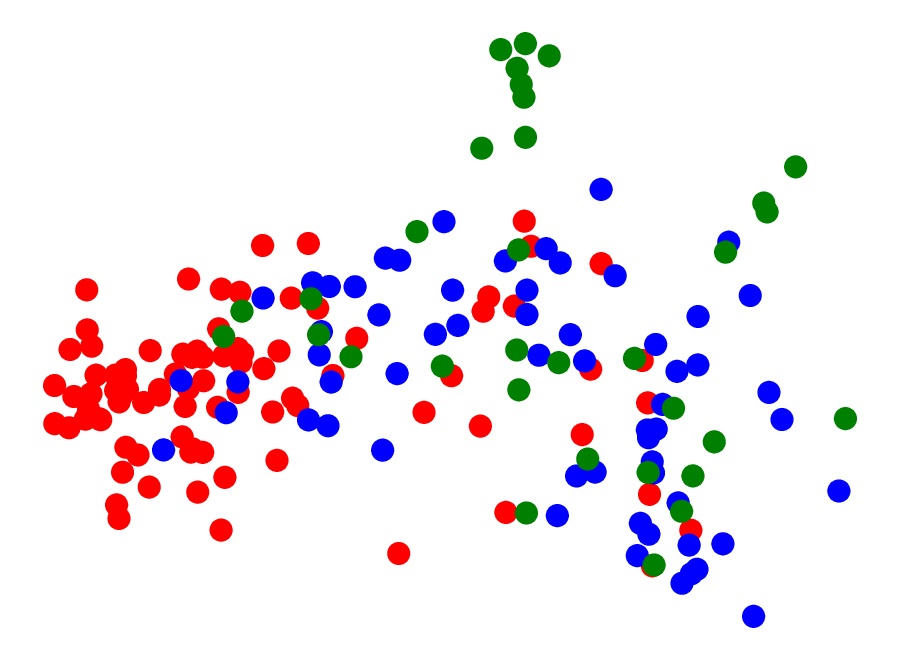}}
            \label{fig:medicine-layer10}\hfill
        \subfloat[Layer=11]{
			\centering
	\includegraphics[scale=0.24]{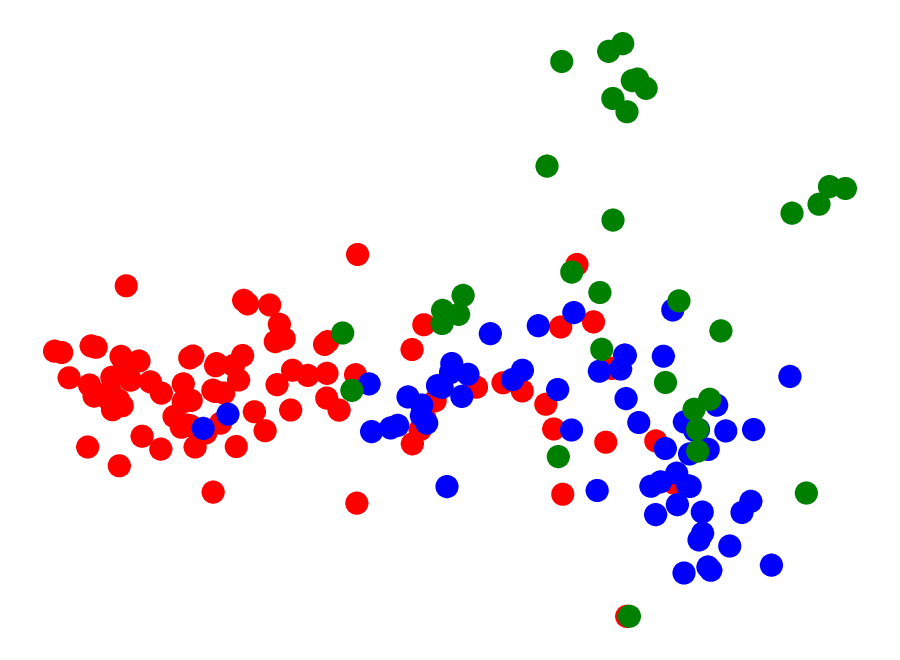}}
            \label{fig:medicine-layer11}\hfill 
        \subfloat[Layer=12]{
			\centering
	\includegraphics[scale=0.24]{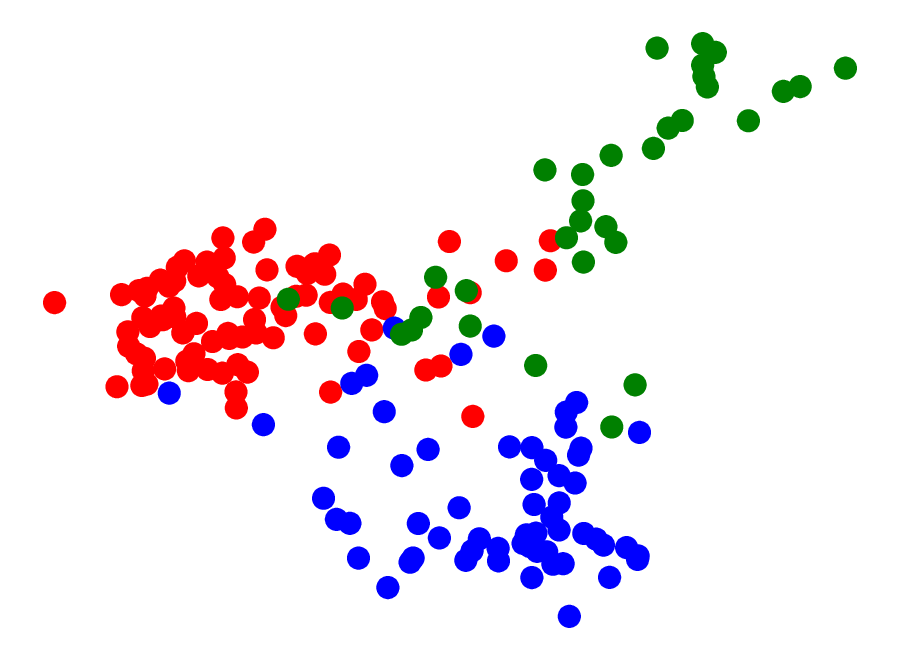}}
            \label{fig:medicine-layer12}\hfill
		\caption{Intermediate-layer contextualized token embeddings for \texttt{patients</w>} (red), \texttt{cells</w>} (blue), and \texttt{disorder</w>} (green) plotted on the plane of the first two principal components. The embeddings are extracted from the GPT-1 model using $200$ sentences sampled from the MedRAG-Textbooks dataset \cite{xiong2024benchmarking} in the medicine domain. As the model progresses from lower to upper layers, the contextualized token embeddings exhibit a clear and progressively increasing separation. The x-axis and y-axis represent the first and second principal components, respectively. More details can be found in the Supplementary Materials.
		}
		\label{fig:pca-visualization-medicine}
\end{figure*}

In this paper, we introduce a quantitative and precise characterization of how LLMs learn contextualized token embeddings for next-token prediction across all layers. As illustrated in Fig.~\ref{fig:law}, our extensive experiments demonstrate that LLMs enhance their ability to predict the next token according to an exponential law, where each layer improves token prediction by approximately an equal multiplicative factor from the first layer to the last. We refer to this as the law of equi-learning. This law is consistently observed across a wide range of open-source LLMs, including those based on the Transformer architecture and more recent architectures like Mamba and RWKV. Specifically, our experiments reveal the emergence of this law in GPT-1 \cite{radford2018improving}, GPT-2 \cite{radford2019language}, Llama-1 \cite{touvron2023llama}, Llama 2 and its fine-tuned variant Llama 2-Chat \cite{touvron2023llama2}, Llama 3 and its instruction-fine-tuned version Llama 3 Instruct \cite{dubey2024llama}, Mistral 7B and its fine-tuned version Mistral 7B-Instruct \cite{jiang2023mistral}, phi-1.5 \cite{li2023textbooks}, phi-2 \cite{javaheripi2023phi}, phi-3 \cite{abdin2024phi}, RWKV and its chat version RWKV-Raven \cite{peng2023rwkv}, and Mamba \cite{gu2024mamba}.

While one might intuitively expect that different token embeddings would be progressively differentiated across the layers of LLMs (an example is shown in Fig.~\ref{fig:pca-visualization-medicine}), it is remarkable that a universal and geometric law governing tens of thousands of tokens emerges in models of such immense complexity. Moreover, the equi-learning law is perhaps the simplest geometric pattern that could arise across intermediate layers, and what is striking is that this pattern is indeed the observed reality. Notably, the law emerges naturally during the training process without any explicit constraints designed to induce its appearance.

The equi-learning law suggests that every layer should be considered equally important in characterizing the formation of features from input embeddings. In particular, the quantitative nature of this law implies that the layer at the midpoint of the model is precisely where the LLM has achieved half of its overall capability in predicting the next token. This finding challenges the view that feature learning can be disproportionately attributed to certain layers over others \cite{tenney2019bert,liu2019linguistic}. Moreover, the equi-learning law provides practical guidelines and insights into several empirical aspects of LLM training. For instance, this law enables a fine-grained understanding of how the overall capabilities of an LLM relate to its depth, leading to a more nuanced perspective on model scaling that goes beyond what is captured by test loss alone. Additionally, this law sheds light on the superiority of next-token prediction---the currently dominant pre-training task---over alternative training approaches employed in models such as BERT~\cite{devlin2019bert}, RoBERTa~\cite{liu2019roberta} and T5~\cite{raffel2020exploring}.

\section{Main Results}
\label{sec:main}

LLMs are nonlinear models designed to predict the subsequent token given a sequence of preceding tokens. Formally, a model processes an input sequence of tokens $x_1, x_2, \ldots, x_t, \ldots$. Each token $x_t$ is initially mapped to a vector $\mathbf{h}_{t,0}$ in the embedding space. These embeddings subsequently undergo transformation through a hierarchical stack of model layers---typically comprising attention mechanisms and various operations---yielding a sequence of contextualized token embeddings $\mathbf{h}_{t,\ell}$ at each layer $1 \leq \ell \leq L$. Notably, $\mathbf{h}_{t,\ell}$ is computed using the outputs from the preceding layer $\{\mathbf{h}_{j,\ell-1} \mid 1 \leq j \leq t\}$, which, due to the autoregressive nature of LLMs, contains information exclusively from tokens at the current position and all preceding positions. Finally, the LLM uses the embedding from the last layer, denoted $\mathbf{h}_{t, \text{last}} := \mathbf{h}_{t,L}$, to predict the subsequent discrete token $x_{t+1}$. Aggregating all such pairs across the entire training corpus yields the dataset $\mathcal{D} := \{(\mathbf{h}_{t, \text{last}}^{s}, x^{s}_{t+1}) \mid 1 \leq s \leq S\}$.

To assess the capability of the LLM in predicting the next token, we evaluate how well a linear regression model fits on the dataset $\mathcal{D}$. For this purpose, we identify $x$ with its index in the token vocabulary. Let $\hat x_{\text{next}} = \mathbf{w} \cdot \mathbf{h} + b$ denote the least-squares fit on $\mathcal{D}$. This suggests using the following metric, which we term the prediction residual (PR), to quantify the LLM's next-token prediction capability:
\begin{equation}\label{eq:PR}
\text{PR} :=  \frac{\sum (x_{\text{next}}- \hat x_{\text{next}})^2}{\sum (x_{\text{next}} - \bar x_{\text{next}})^2},
\end{equation}
where the sum is over all $x_{\text{next}} = x^{s}_{t+1}$, $\hat x_{\text{next}} = \mathbf{w} \cdot \mathbf{h}_{t, \text{last}}^{s} + b$, and $\bar x_{\text{next}}$ represents the mean of all $x^{s}_{t+1}$. In statistical terms, this measure is known as the fraction of variance unexplained, equivalent to one minus the coefficient of determination \cite{weisberg2005applied}. It serves as a canonical metric for evaluating the proportion of variance in the dependent variable that remains unaccounted for by the independent variables. A high PR value indicates limited predictive power of the token embeddings, while a low value suggests strong predictive capability. Thus, PR inherently captures how effectively the token embeddings explain next-token prediction\footnote{See more elaboration on this metric in Section \ref{sec:disc}.}, aligning with the linear probing paradigm widely employed to analyze the structural properties of contextualized representations in LLMs \cite{hewitt2019structural, liu2019linguistic}.

To investigate how the predictive power of an LLM with depth $L$ evolves across its layers, we calculate the PR for the next-token prediction task at each intermediate layer. Let $\text{PR}_l$ denote this value for the $l$-th layer, where $1 \le l \le L$. Specifically, instead of using the last-layer embedding \( \mathbf{h}_{t,\text{last}} \equiv \mathbf{h}_{t,L} \), we use the embedding of the current token at layer \( \ell \), denoted \( \mathbf{h}_{t,\ell} \), to predict the next token when computing \( \text{PR}_\ell \). The law of equi-learning (see Fig.~\ref{fig:law}) states that the dynamics of predictive power across layers follow the relationship
\[
\text{PR}_l \approx \rho^{l-1} \times \text{PR}_1 
\]
for some decay ratio $0 < \rho < 1$. Since the token embeddings at the input layer (the 0-th layer) are not yet contextualized, they are not included. This implies that $\log \text{PR}_{l+1} - \log \text{PR}_l \approx - \log \frac{1}{\rho}$, indicating a roughly constant reduction in the logarithm of the PR value across each layer, hence the name equi-learning. The Pearson correlation coefficients between the logarithm of the PR value and the layer index range from $-0.983$ to $-0.997$ in Fig.~\ref{fig:law}. For GPT-1, for example, the decay ratio $\rho$ is approximately $0.993$ (as shown in the top-left plot of Fig.~\ref{fig:law}). In general, $\rho$ depends on various factors, including the model architecture, pre-training data, model depth, feature dimension, and pre-training time.

This law provides what may be the first precise geometric characterization of the learning process for contextualized token embeddings within the intermediate layers of LLMs. Notably, the pre-training objective focuses solely on the last-layer embeddings, aiming to minimize a loss function associated with the last-layer value of PR. Surprisingly, this training dynamics inherently ensure that each layer contributes equally, rather than allowing some layers to carry a disproportionate amount of the workload. 

While related phenomena have been observed in MLPs for classification tasks \cite{he2023law,papyan2020prevalence}, the law presented in this work makes distinct contributions through its focus on LLMs trained for next-token prediction. This task fundamentally differs from classification in several key aspects: it processes sequential data with progressively increasing text lengths and operates within an inherently probabilistic framework without ground truth labels for token predictions. Furthermore, whereas previous studies primarily examined the terminal phase of MLP training, our work demonstrates that the law of equi-learning manifests well before the terminal phase in LLMs, which typically undergo very few epochs of training. This early emergence is particularly noteworthy given the substantial complexity of Transformer-based architectures compared to MLPs. Importantly, this law illuminates the internal mechanisms of the sophisticated LLM architectures \cite{wu2024linguistic}.

The universality of this law is demonstrated by its emergence across a diverse range of open-source LLMs, spanning different architectures, model sizes, and pre-training datasets. Our investigation employs a comprehensive collection of probing datasets to calculate PR and evaluate the law of equi-learning, including BookCorpus \cite{zhu2015aligning}, C4 \cite{raffel2020exploring}, OpenWebText \cite{Gokaslan2019OpenWeb}, Wikipedia \cite{wikidump}, peS2o \cite{peS2o}, The Pile \cite{gao2020pile}, Redpajama \cite{together2023redpajama}, and OSCAR \cite{suarez2020monolingual}.

\begin{figure*}[!htp]
     \captionsetup[subfigure]{labelformat=empty}
		\centering
        	\subfloat[Step=0]{
			\centering
		\includegraphics[scale=0.24]{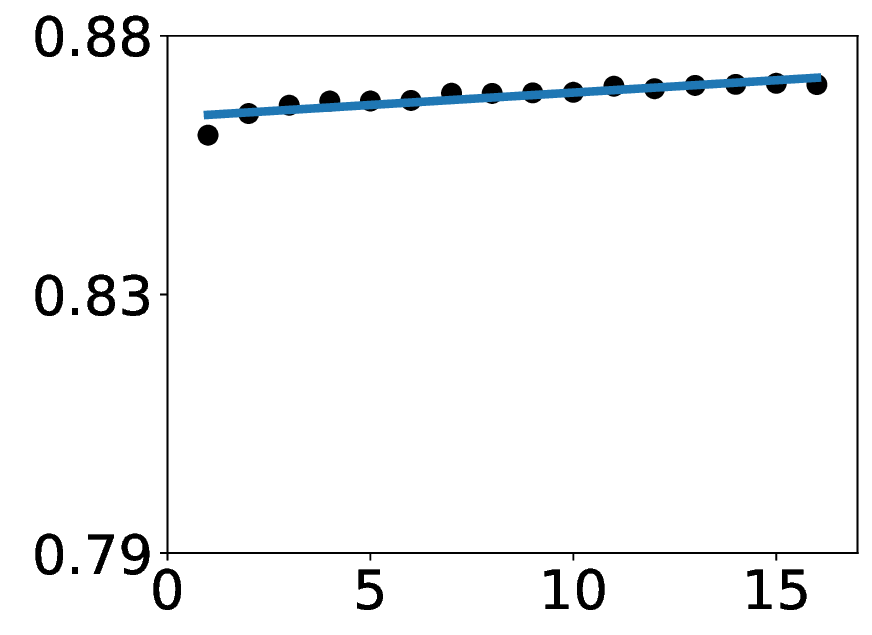}}
            \label{fig:pythia-1b-step0}\hfill
            \subfloat[Step=4]{
			\centering
		\includegraphics[scale=0.24]{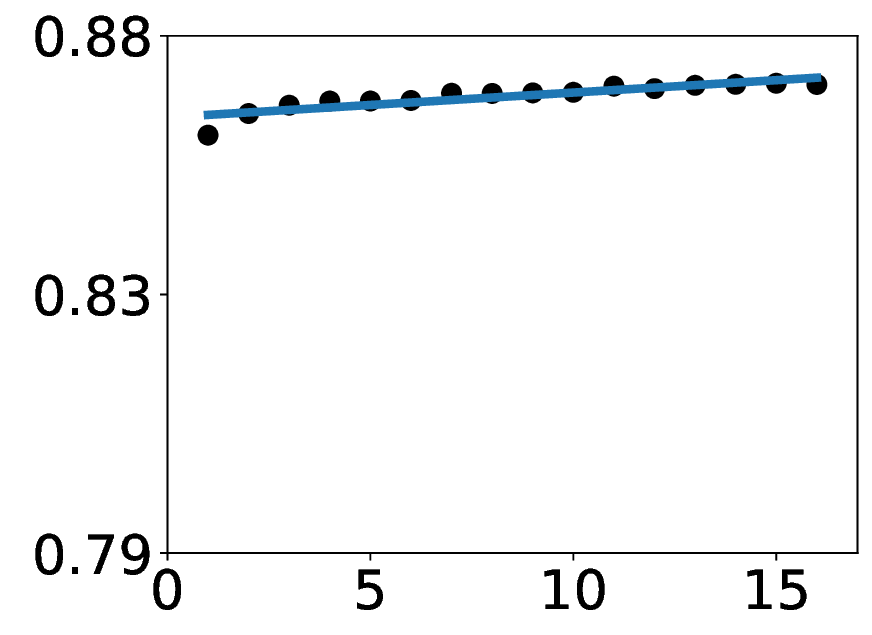}}
            \label{fig:pythia-1b-step4}\hfill
     	\subfloat[Step=8]{
			\centering
		\includegraphics[scale=0.24]{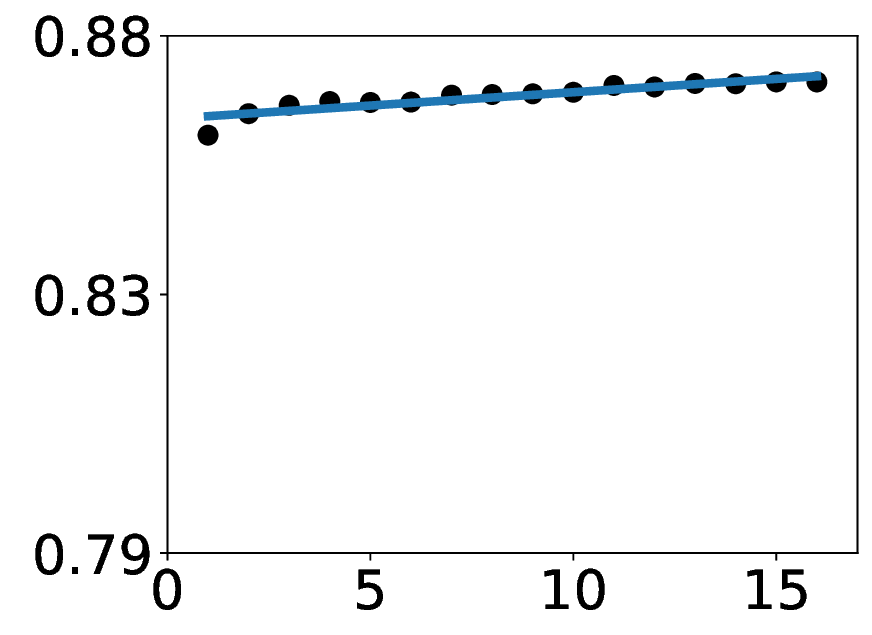}}
            \label{fig:pythia-1b-step8}\hfill
        \subfloat[Step=16]{
			\centering
		\includegraphics[scale=0.24]{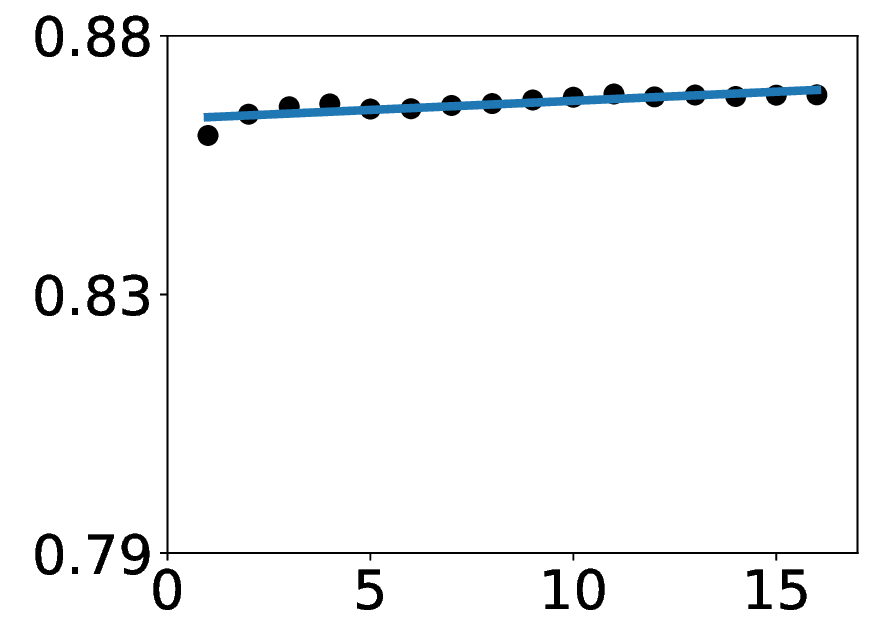}}
            \label{fig:pythia-1b-step16}\hfill
            
            \subfloat[Step=32]{
			\centering
		\includegraphics[scale=0.24]{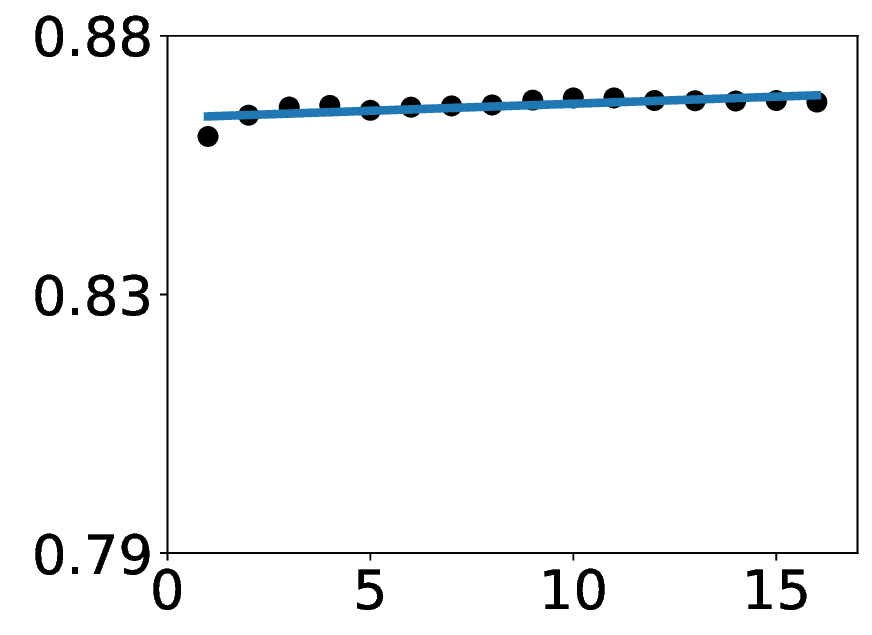}}
            \label{fig:pythia-1b-step32}\hfill
            \subfloat[Step=64]{
			\centering
		\includegraphics[scale=0.24]{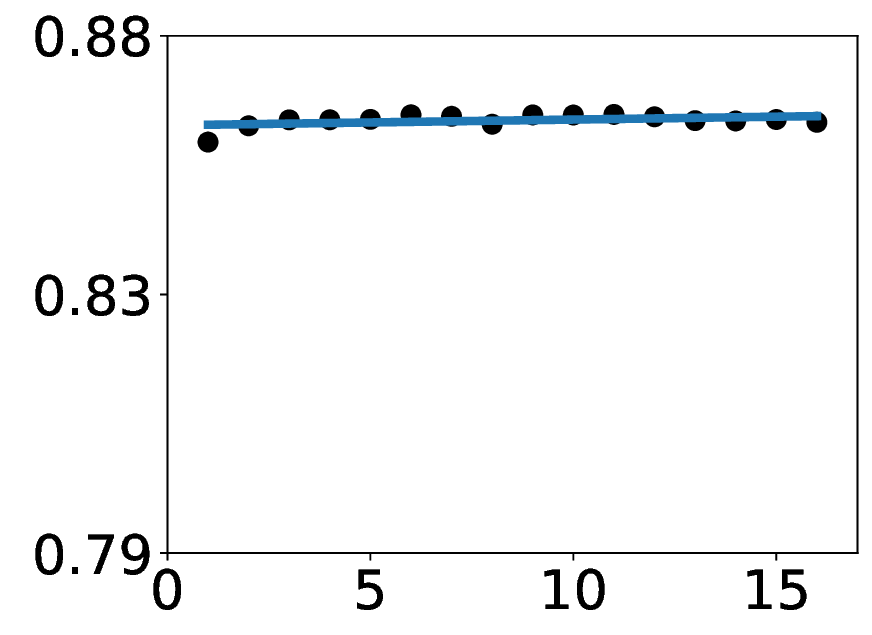}}
            \label{fig:pythia-1b-step64}\hfill
            \subfloat[Step=128]{
			\centering
		\includegraphics[scale=0.24]{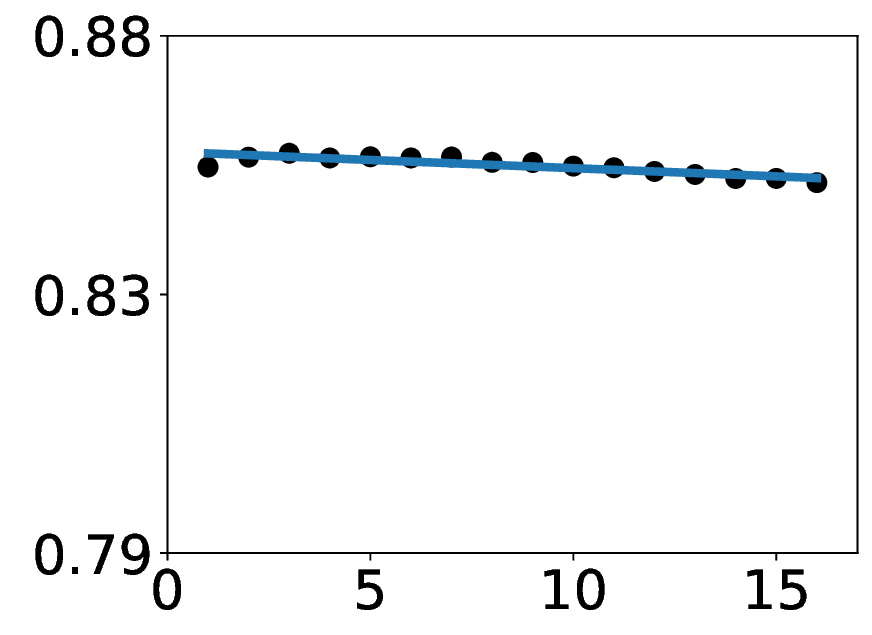}}
            \label{fig:pythia-1b-step128}\hfill
     	\subfloat[Step=256]{
			\centering
		\includegraphics[scale=0.24]{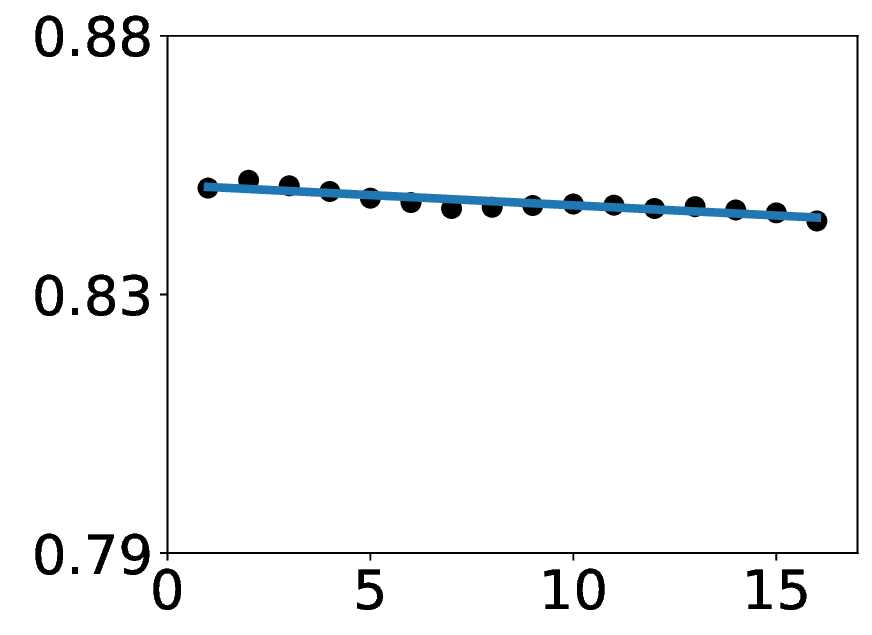}}
            \label{fig:pythia-1b-step256}\hfill
            
            \subfloat[Step=512]{
			\centering
		\includegraphics[scale=0.24]{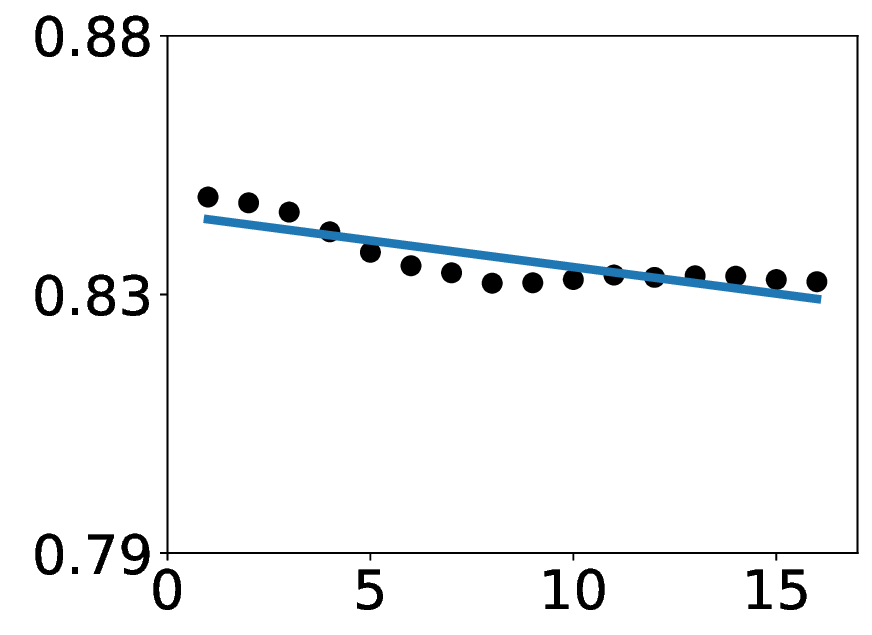}}
            \label{fig:pythia-1b-step512}\hfill
            \subfloat[Step=1000]{
			\centering
		\includegraphics[scale=0.24]{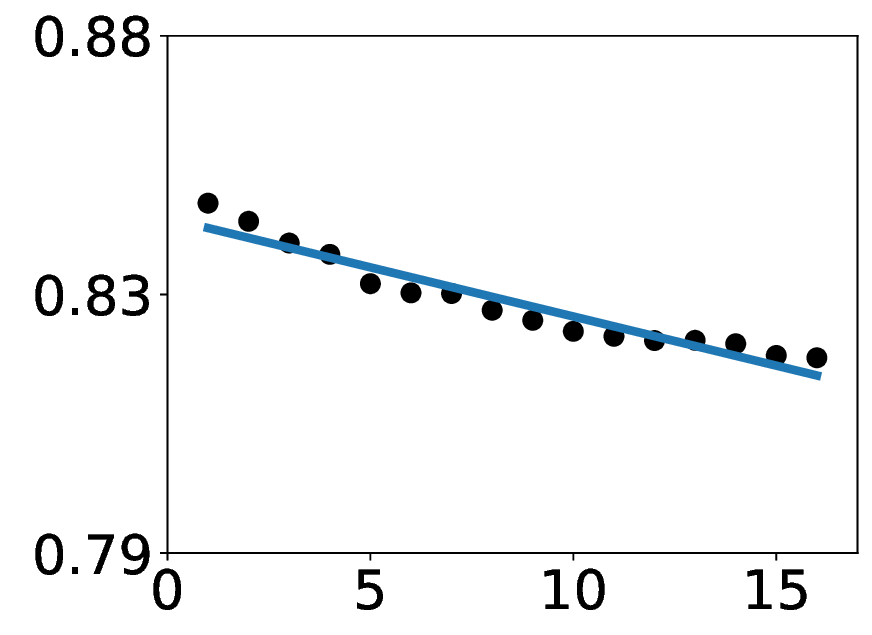}}
            \label{fig:pythia-1b-step1000}\hfill
            \subfloat[Step=2000]{
			\centering
		\includegraphics[scale=0.24]{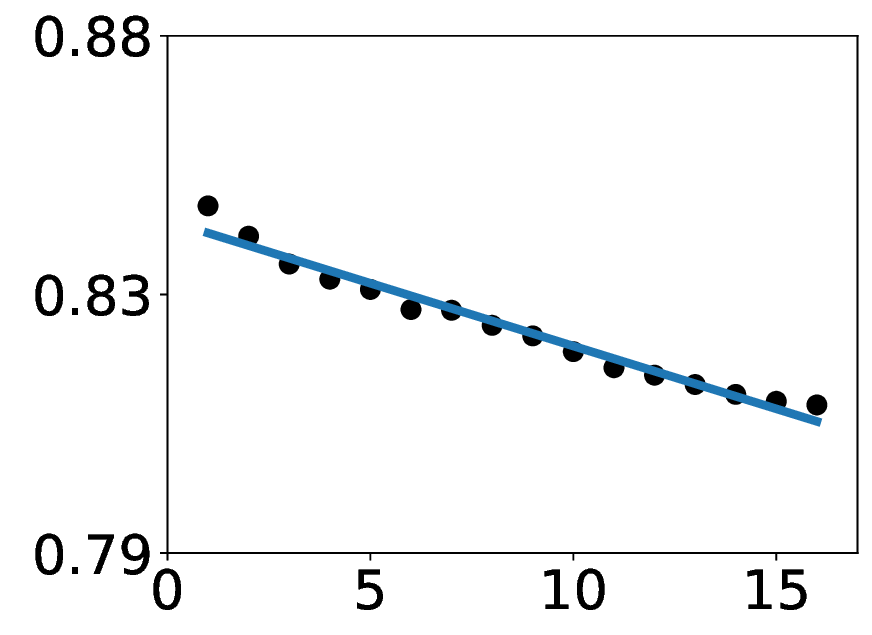}}
            \label{fig:pythia-1b-step2000}\hfill
            \subfloat[Step=4000]{
			\centering
		\includegraphics[scale=0.24]{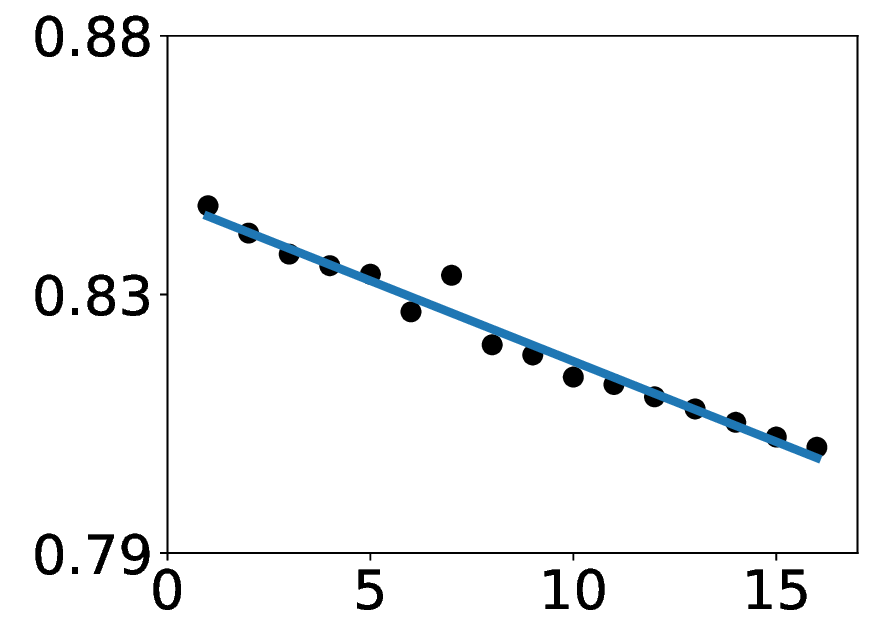}}
            \label{fig:pythia-1b-step4000}\hfill
            
            \subfloat[Step=8000]{
			\centering
		\includegraphics[scale=0.24]{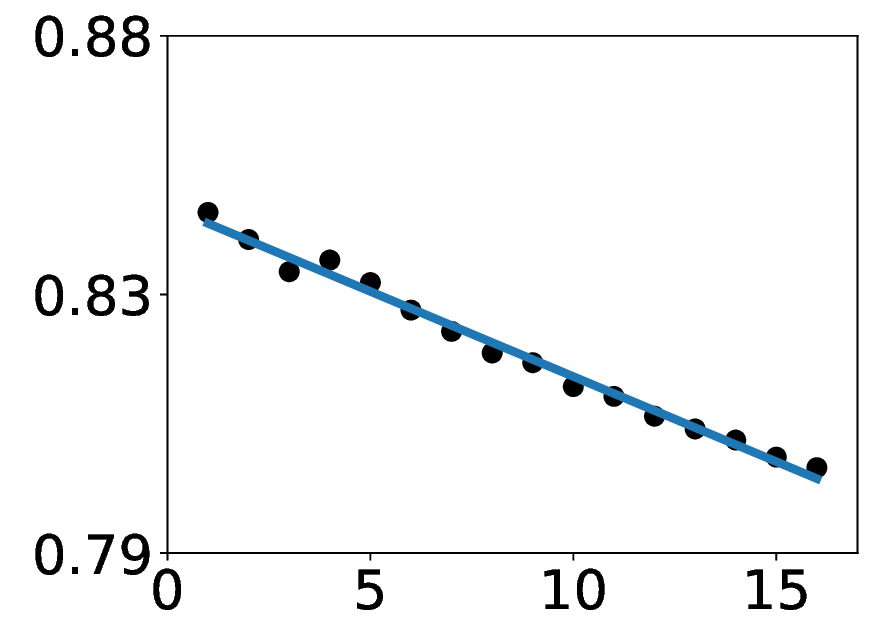}}
            \label{fig:pythia-1b-step8000}\hfill
        \subfloat[Step=16000]{
			\centering
		\includegraphics[scale=0.24]{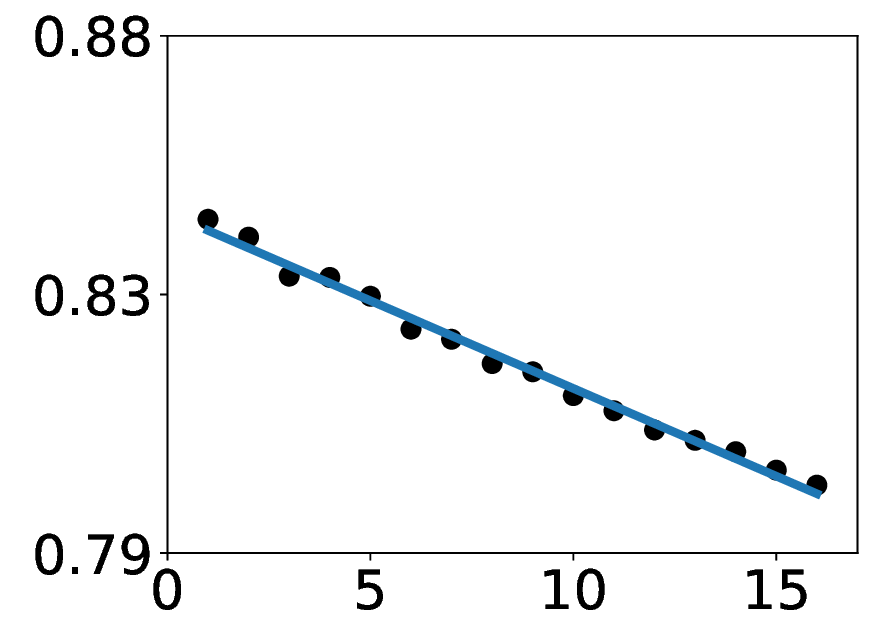}}
            \label{fig:pythia-1b-step16000}\hfill
            \subfloat[Step=128000]{
			\centering
		\includegraphics[scale=0.24]{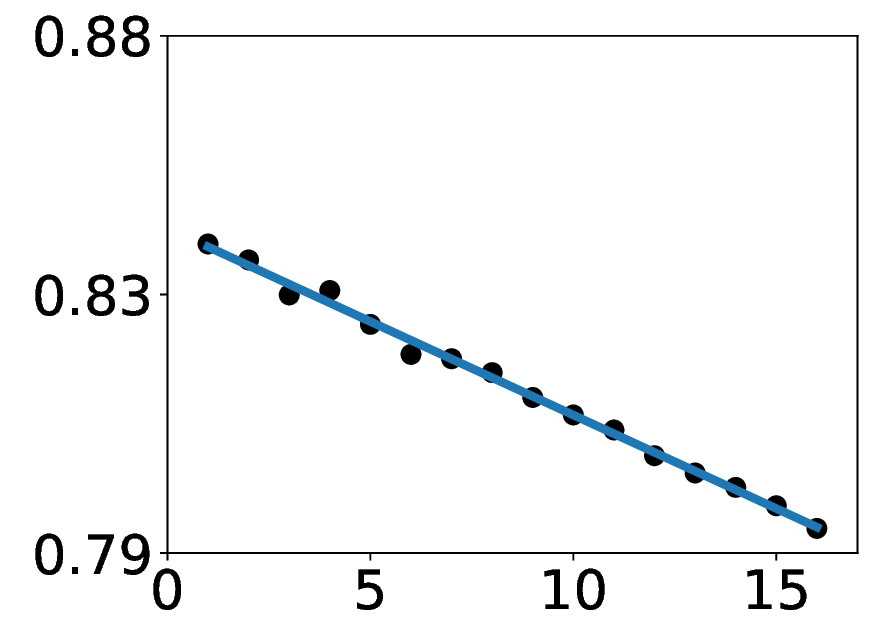}}
            \label{fig:pythia-1b-step128000}\hfill
     	\subfloat[Step=143000]{
			\centering
		\includegraphics[scale=0.24]{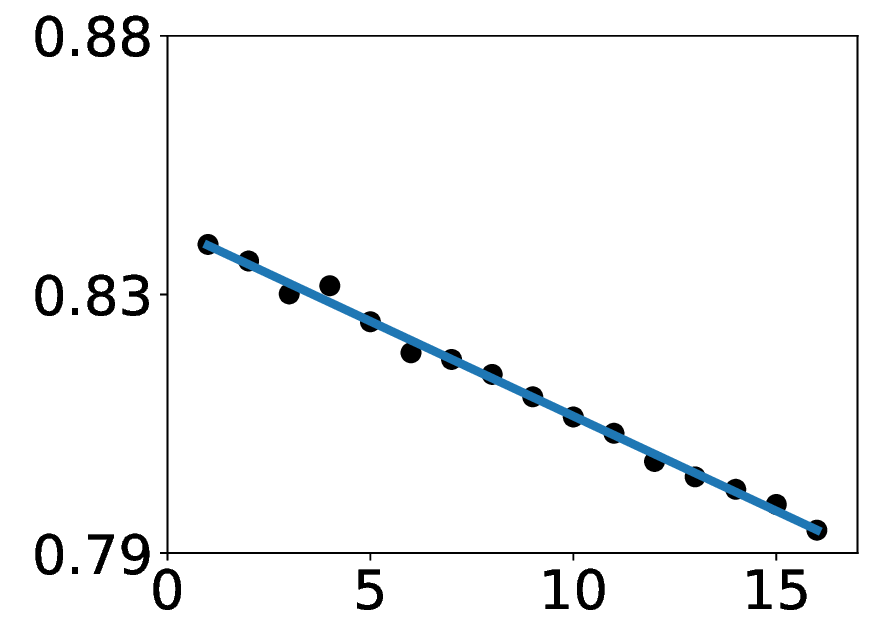}}
            \label{fig:pythia-1b-step143000}
		\caption{Pythia-1B \cite{biderman2023pythia} trained on the Pile dataset for various training steps, using a batch size of 2 million, with the total number of steps reaching 143,000. The x-axis denotes the layer index, while the y-axis (log scale) shows the prediction residual (PR) as defined in Eq.~\ref{eq:PR}. Refer to Fig.~\ref{fig:pythia-1b-step0-original} in the Supplementary Materials for an enlarged version at initialization (Step=0).
		}
		\label{fig:training-step}
\end{figure*}

\textbf{When does the law emerge?}
To deepen our understanding of the law's dynamics during training, we investigate the effects of three key factors---training steps, training epochs, and data repetition---on its emergence throughout the process. As illustrated in Fig.~\ref{fig:training-step}, the progression of this law as a function of training steps closely resembles the behavior of the equi-separation law observed in MLPs \cite{he2023law}. At model initialization, the PR of contextualized token embeddings for next-token prediction may exhibit an upward trend from lower to higher layers. However, after a certain amount of training (e.g., around 8,000 steps), the law of equi-learning becomes apparent. Beyond this point, the decay ratio continues to decrease until convergence, with the law consistently manifesting during this phase.

\begin{figure*}[!htp]
    \captionsetup[subfigure]{labelformat=empty}
		\centering
        	\subfloat[Epoch=1]{
			\centering
		\includegraphics[scale=0.24]{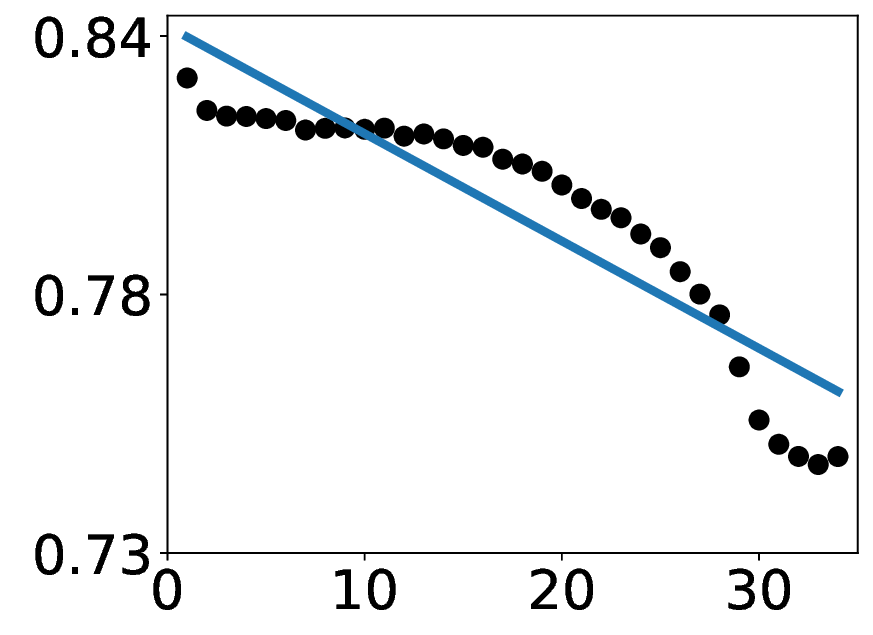}}
            \label{fig:c4-2.8b-epoch1}\hfill
            \subfloat[Epoch=2]{
			\centering
		\includegraphics[scale=0.24]{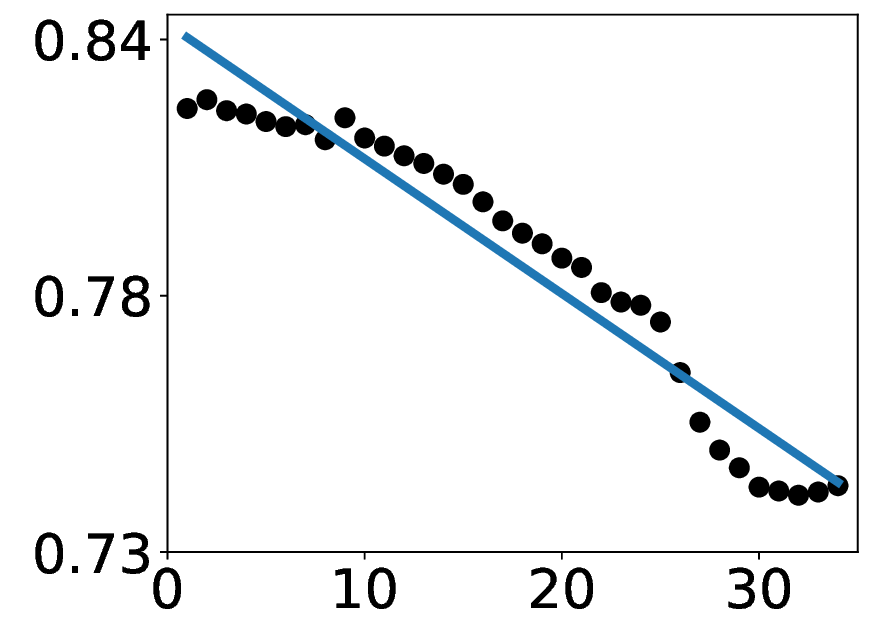}}
            \label{fig:c4-2.8b-epoch2}\hfill
     	\subfloat[Epoch=3]{
			\centering
		\includegraphics[scale=0.24]{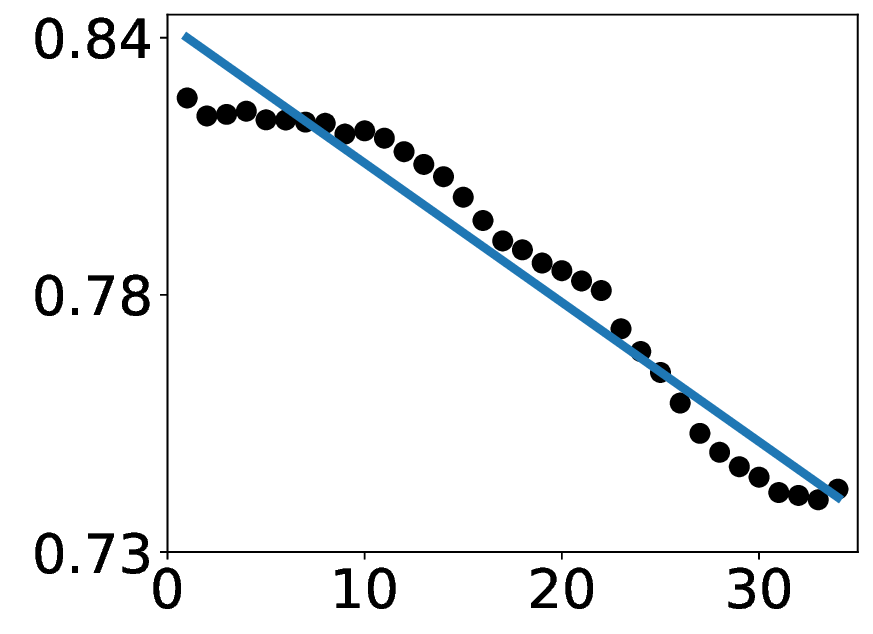}}
            \label{fig:c4-2.8b-epoch3}\hfill
        \subfloat[Epoch=4]{
			\centering
		\includegraphics[scale=0.24]{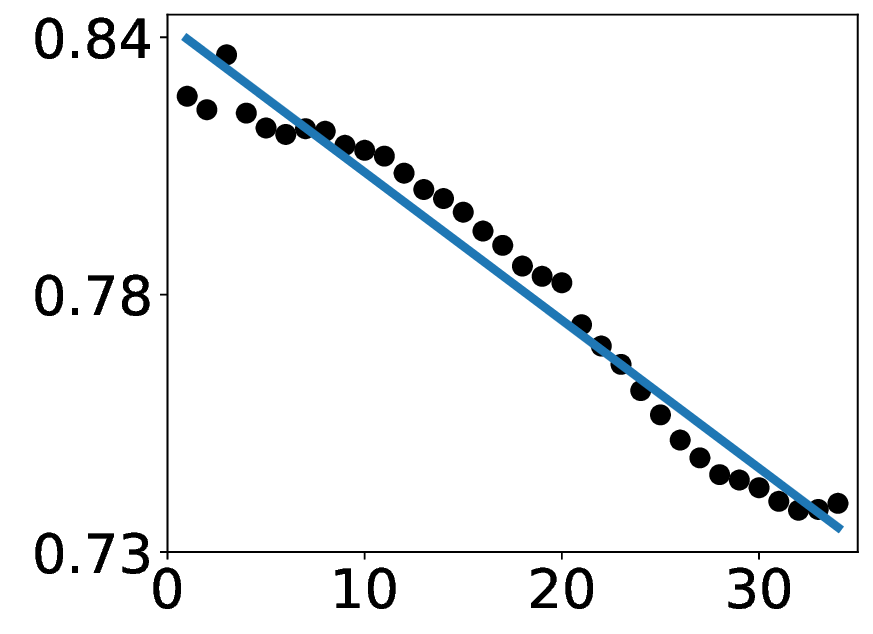}}
            \label{fig:c4-2.8b-epoch4}\hfill
            
            \subfloat[Epoch=6]{
			\centering
		\includegraphics[scale=0.24]{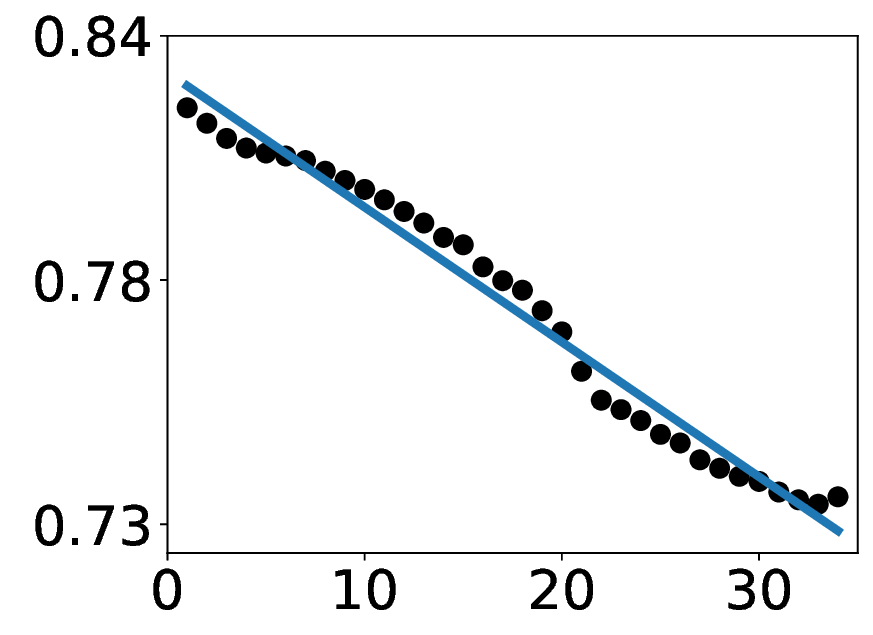}}
            \label{fig:c4-2.8b-epoch6}
     	\subfloat[Epoch=8]{
			\centering
		\includegraphics[scale=0.24]{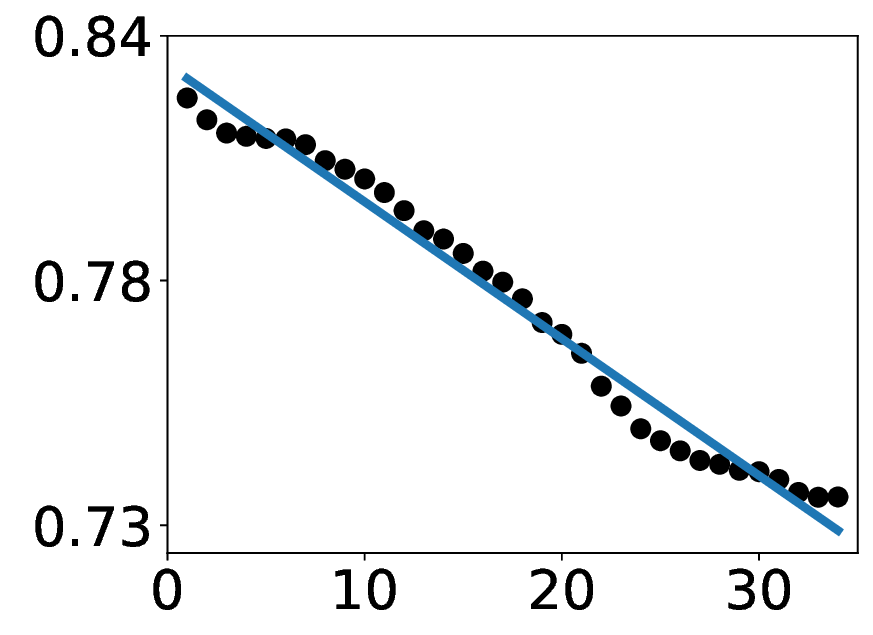}}
            \label{fig:c4-2.8b-epoch8}
            \subfloat[Epoch=10]{
			\centering
		\includegraphics[scale=0.24]{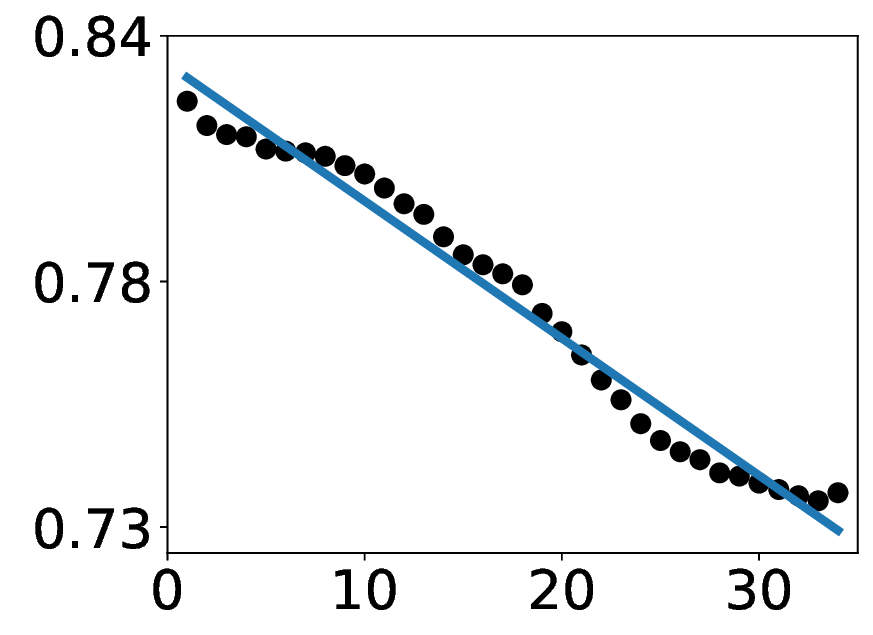}}
            \label{fig:c4-2.8b-epoch10}
		\caption{A 2.8B GPT-2 model pre-trained on a 4 billion token subset of C4 over multiple epochs. The x-axis denotes the layer index, while the y-axis (log scale) shows the prediction residual (PR) as defined in Eq.~\ref{eq:PR}.
		}
		\label{fig:training-epoch}
\end{figure*}

\begin{figure*}[!htp]
    \captionsetup[subfigure]{labelformat=empty}
		\centering
        	\subfloat[Repeat=1]{
			\centering
		\includegraphics[scale=0.24]{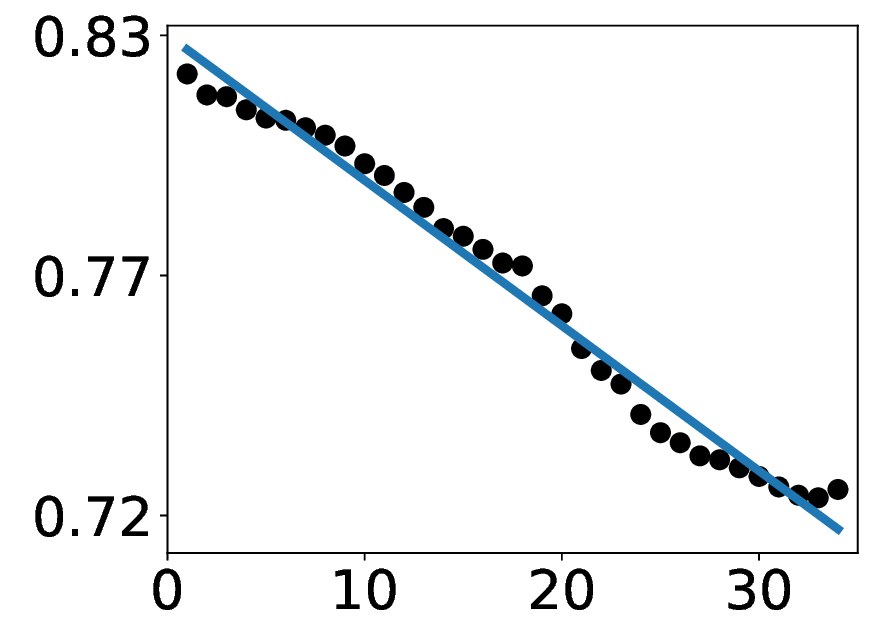}}
            \label{fig:c4-2.8b-repeat1}\hfill
            \subfloat[Repeat=2]{
			\centering
		\includegraphics[scale=0.24]{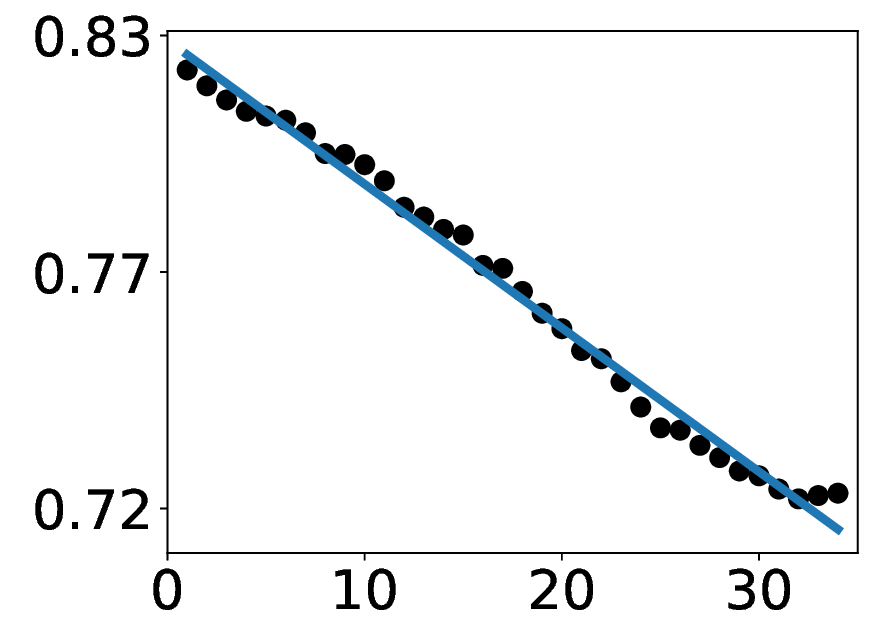}}
            \label{fig:c4-2.8b-repeat2}\hfill
     	\subfloat[Repeat=3]{
			\centering
		\includegraphics[scale=0.24]{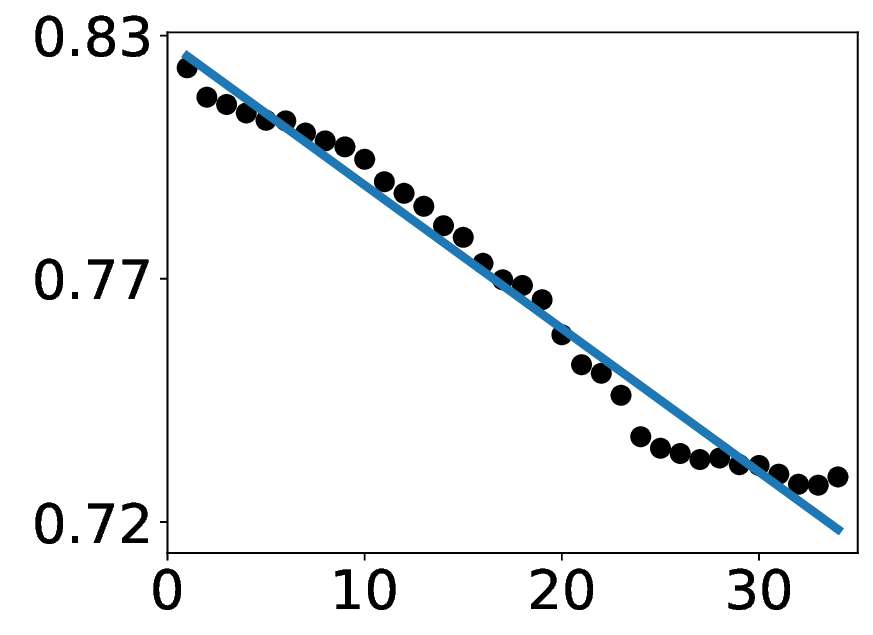}}
            \label{fig:c4-2.8b-repeat3}\hfill
        \subfloat[Repeat=4]{
			\centering
		\includegraphics[scale=0.24]{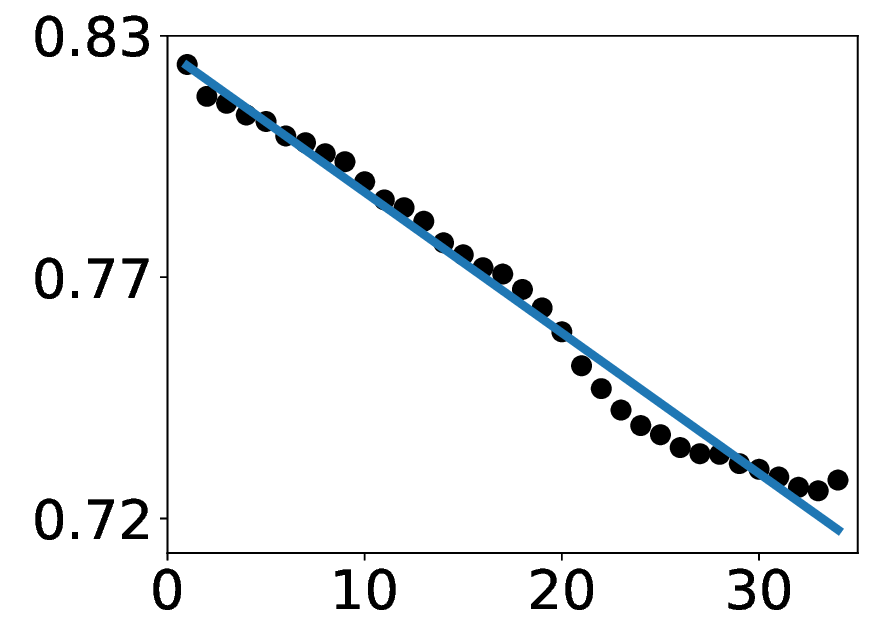}}
            \label{fig:c4-2.8b-repeat4}\hfill
            
            \subfloat[Repeat=5]{
			\centering
		\includegraphics[scale=0.24]{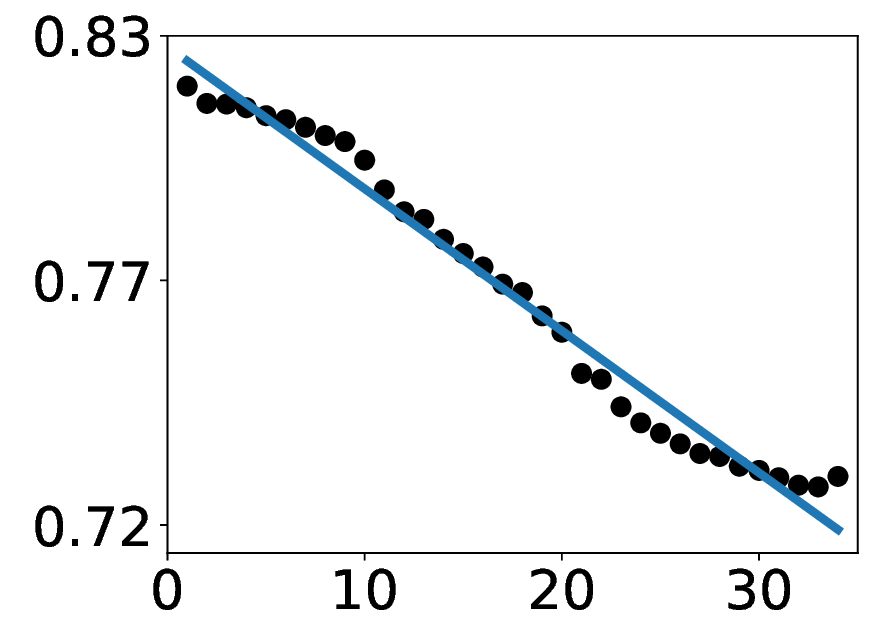}}
            \label{fig:c4-2.8b-repeat5}\hfill
     	\subfloat[Repeat=6]{
			\centering
		\includegraphics[scale=0.24]{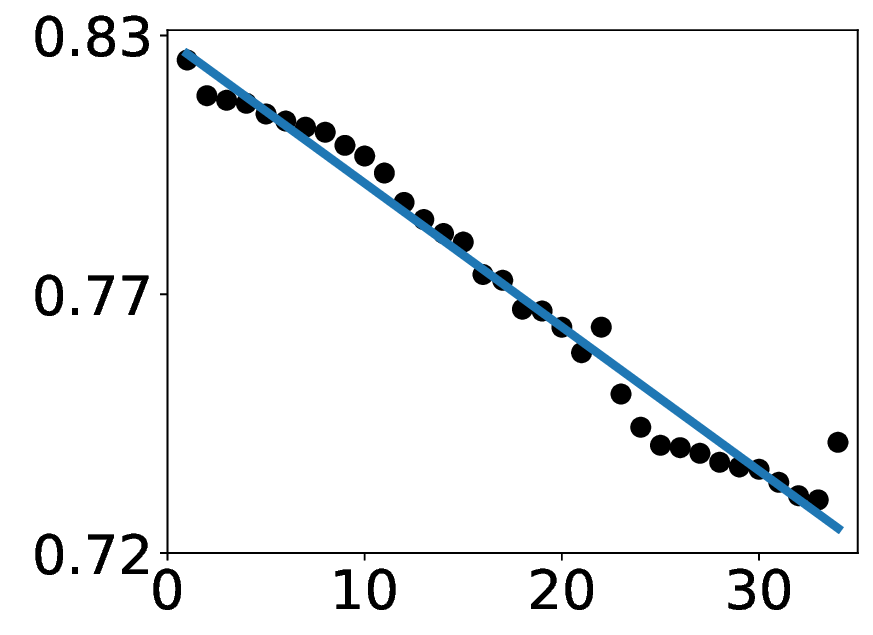}}
            \label{fig:c4-2.8b-repeat6}\hfill
            \subfloat[Repeat=14]{
			\centering
		\includegraphics[scale=0.24]{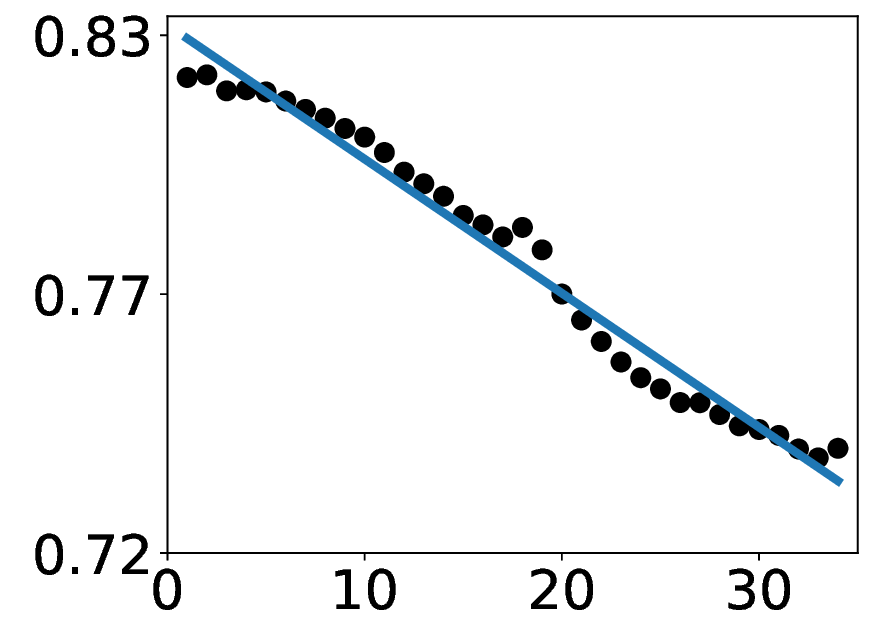}}
            \label{fig:c4-2.8b-repeat14}\hfill
            \subfloat[Repeat=44]{
			\centering
		\includegraphics[scale=0.24]{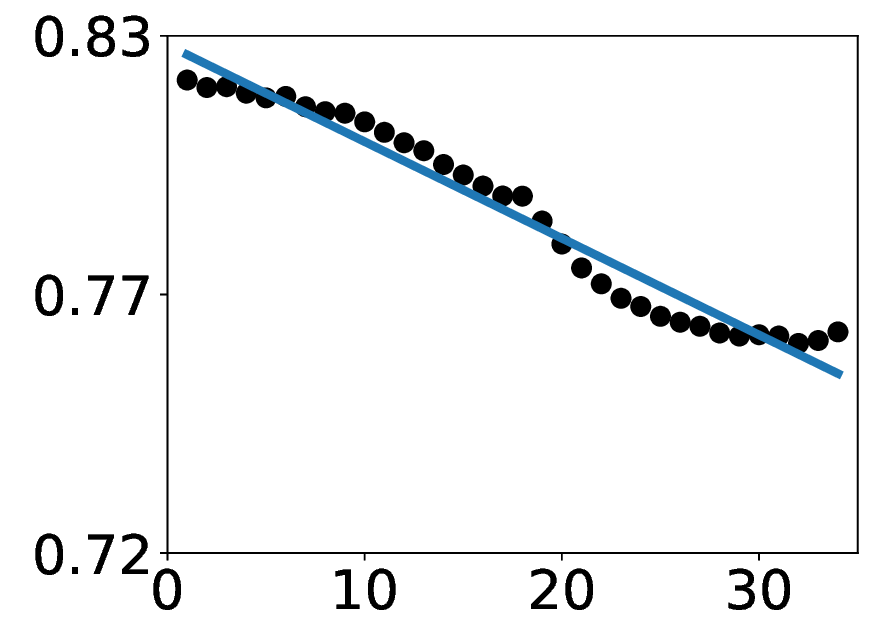}}
            \label{fig:c4-2.8b-repeat44}
		\caption{A 2.8B GPT-2 model pre-trained on varying numbers of unique tokens from the C4 dataset, with a constant total of 55 billion training tokens. Note that the number of unique tokens equals 55 billion divided by the number of repeats. The x-axis denotes the layer index, while the y-axis (log scale) shows the prediction residual (PR) as defined in Eq.~\ref{eq:PR}.
		}
		\label{fig:repeat-number}
\end{figure*}

Although LLMs are typically trained over a limited number of epochs or even a single epoch, we also explore the training dynamics in data-constrained regimes, anticipating that the availability of text data may soon be limited by the finite volume of Internet content \cite{muennighoff2024scaling}. Specifically, we utilize various pre-trained 2.8B GPT-2 models released by \cite{muennighoff2024scaling} to examine the impact of training epochs and data repetition. As depicted in Fig.\ref{fig:training-epoch} and Fig.\ref{fig:repeat-number}, the law of equi-learning emerges as long as the number of epochs is not too small and the number of repeats is not excessively high. Considering these three factors---training steps, training epochs, and data repetition---We observe that a sufficient total number of tokens facilitates the emergence of the equi-learning law, provided that the number of unique tokens is adequate. This finding serves as a crucial condition for training effective LLMs and offers insights into optimizing the training process.

\section{Perspectives from the Law}

The universality of the equi-learning law provides fine-grained perspectives that are applicable to the practical development of LLMs. These perspectives offer new insights into the training processes of LLMs and contribute to advancing transparency in these black-box models. We illustrate the impact of this law on key aspects such as model scaling, pre-training tasks, and information flow. Additional findings, including the impact of pre-training data quality, are discussed in the Supplementary Materials.

\begin{figure*}[!t]
    \captionsetup[subfigure]{labelformat=empty}
		\centering
         \rotatebox[y=1.5cm]{90}{GPT-2}\quad
        \subfloat[\footnotesize GPT-2]{
			\centering
		\includegraphics[scale=0.24]{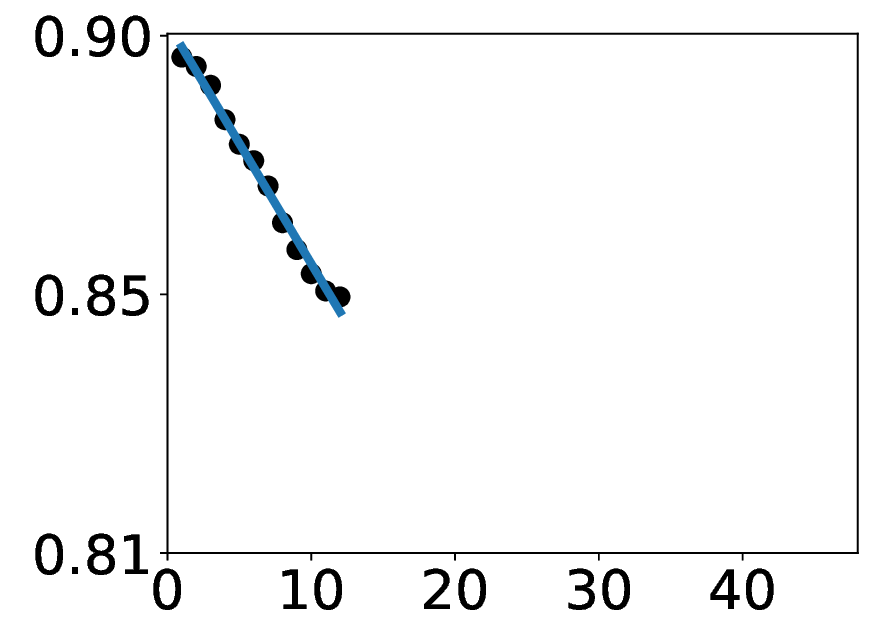}}
            \label{fig:gpt2}\hfill
        \subfloat[GPT-2 Medium]{
			\centering
		\includegraphics[scale=0.24]{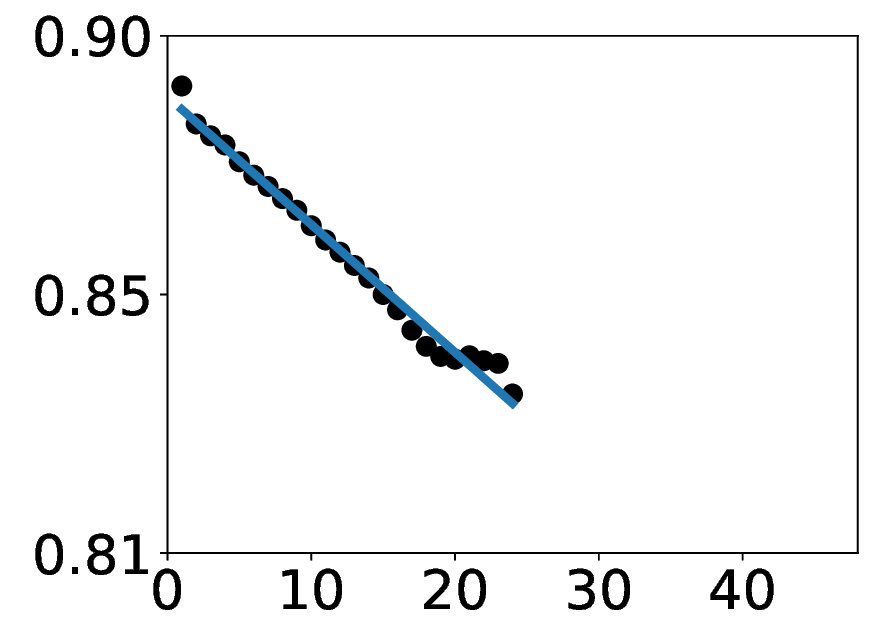}}
            \label{fig:gpt2-medium}\hfill
            \subfloat[GPT-2 Large]{
			\centering
		\includegraphics[scale=0.24]{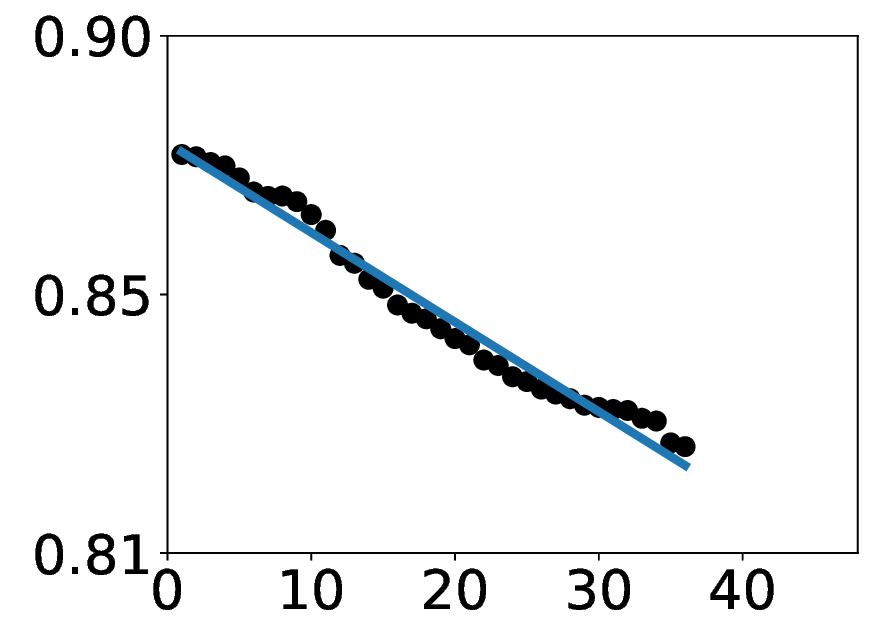}}
            \label{fig:gpt2-large}\hfill
            \subfloat[GPT-2 XL]{
			\centering
		\includegraphics[scale=0.24]{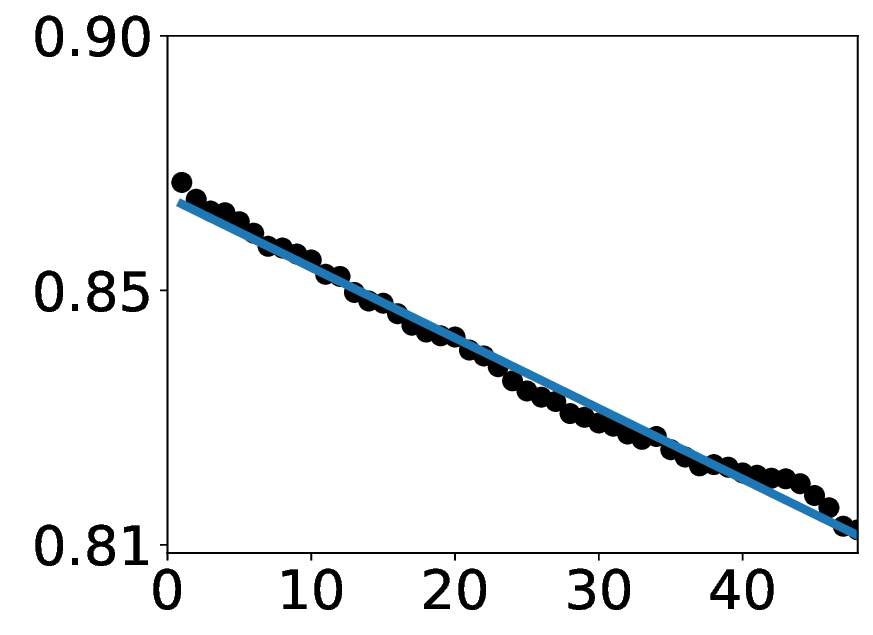}}
            \label{fig:gpt2-xl-size}\hfill 

             \rotatebox[y=1.5cm]{90}{\footnotesize RWKV-Raven}\quad
     	\subfloat[RWKV-Raven-1.5B]{
			\centering
		\includegraphics[scale=0.24]{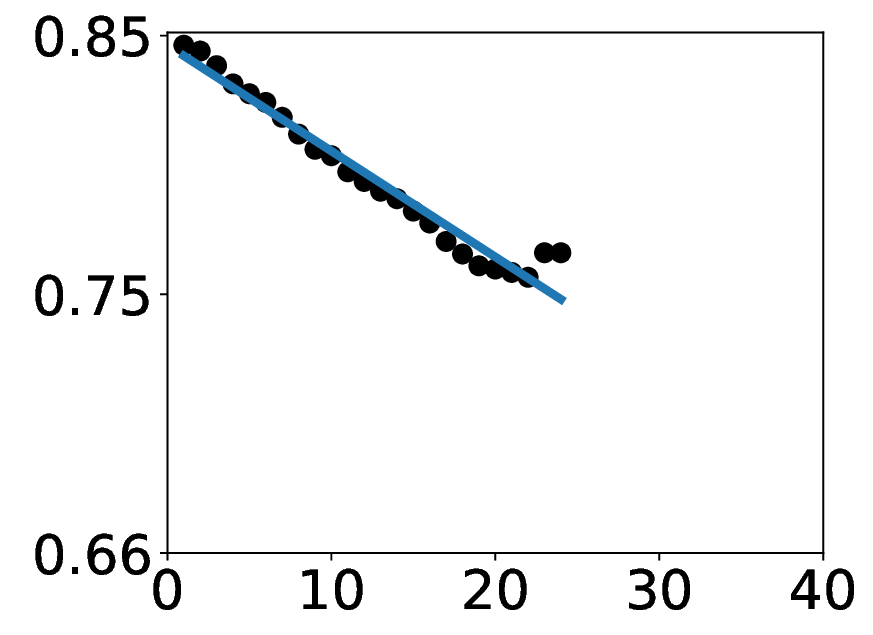}}
            \label{fig:rwkv-raven-1b5}\hfill
            \subfloat[RWKV-Raven-3B]{
			\centering
		\includegraphics[scale=0.24]{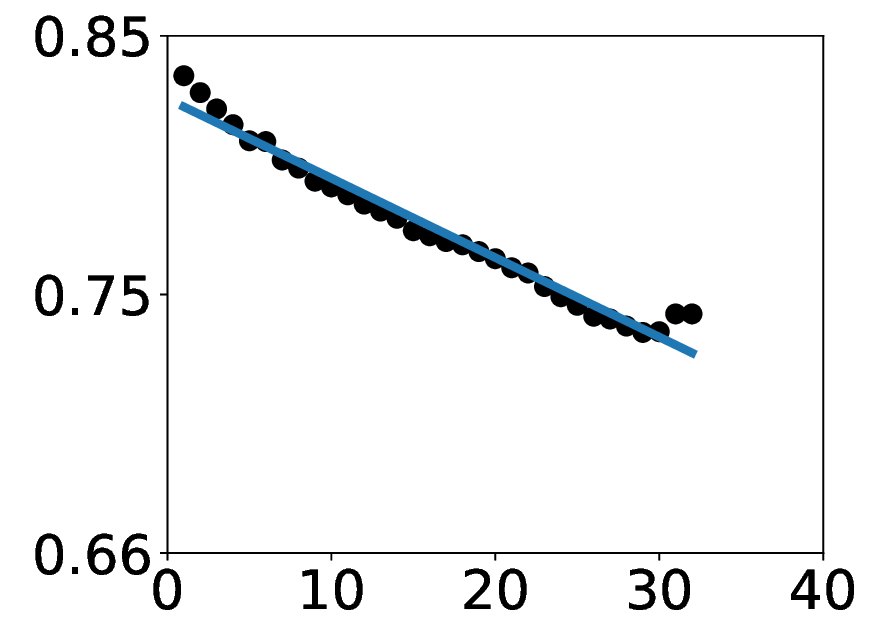}}
            \label{fig:rwkv-raven-3b}\hfill
            \subfloat[RWKV-Raven-7B]{
			\centering
		\includegraphics[scale=0.24]{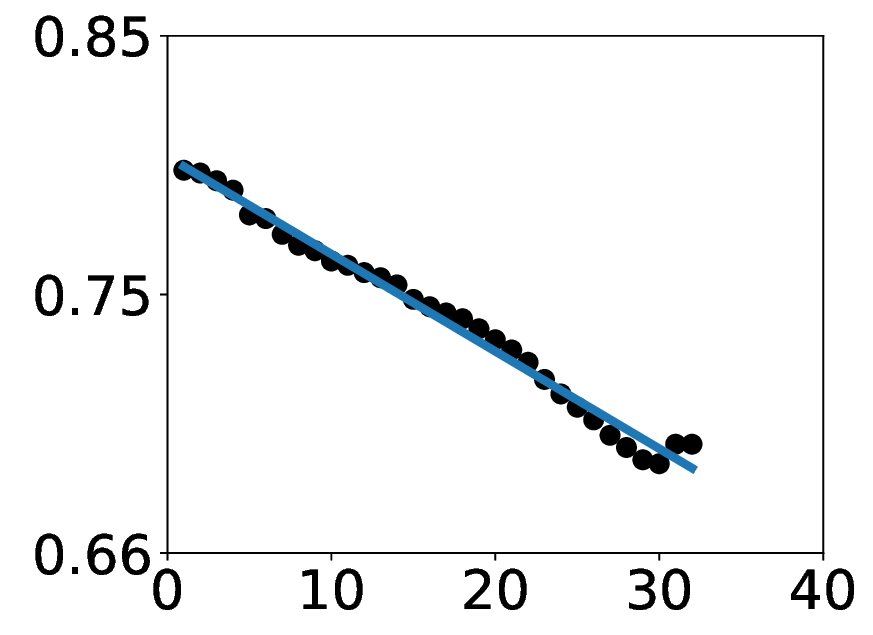}}
            \label{fig:rwkv-raven-7b}\hfill
            \subfloat[RWKV-Raven-14B]{
			\centering
		\includegraphics[scale=0.24]{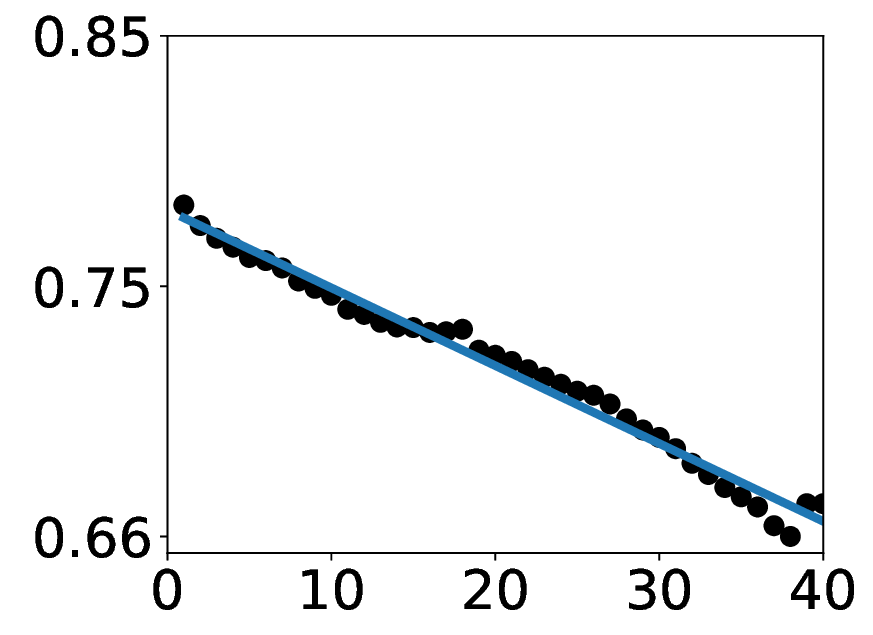}}
            \label{fig:rwkv-raven-14b-size}\hfill

             \rotatebox[y=1.5cm]{90}{\footnotesize Mamba}\quad
            \subfloat[Mamba-370M]{
			\centering
		\includegraphics[scale=0.24]{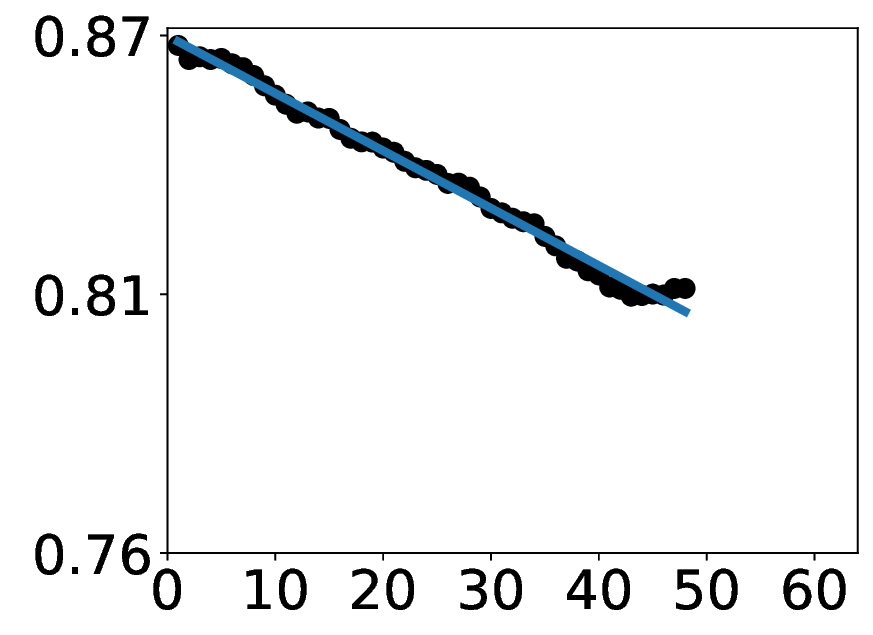}}
            \label{fig:mamba-370m}\hfill
            \subfloat[Mamba-790M]{
			\centering
		\includegraphics[scale=0.24]{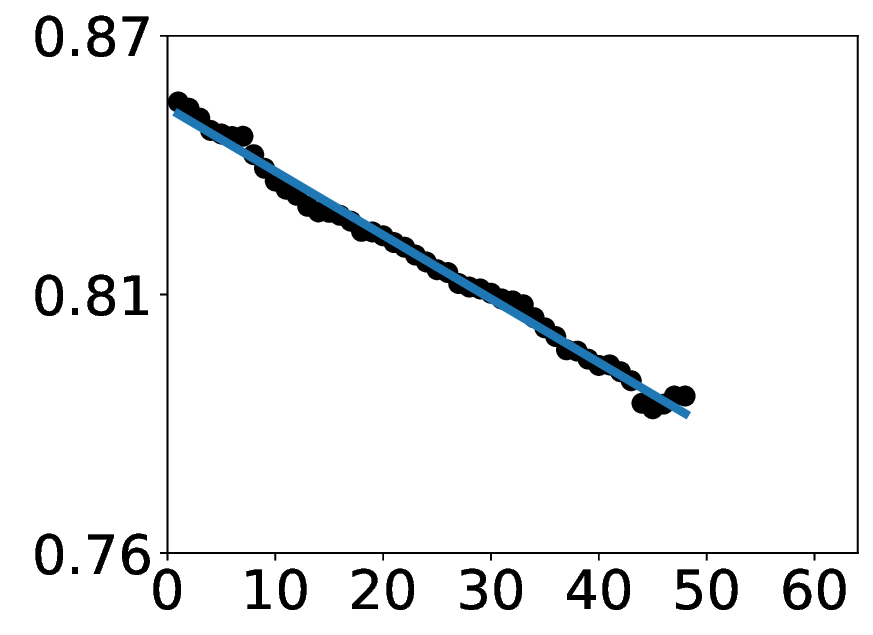}}
            \label{fig:mamba-790m}\hfill
     	\subfloat[Mamba-1.4B]{
			\centering
		\includegraphics[scale=0.24]{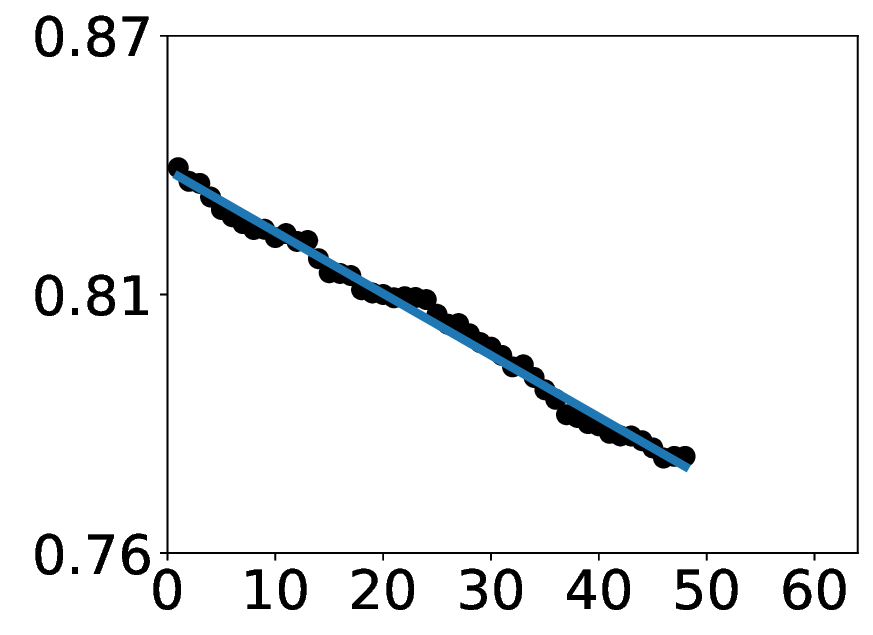}}
            \label{fig:mamba-1.4b}\hfill
            \subfloat[Mamba-2.8B]{
			\centering
		\includegraphics[scale=0.24]{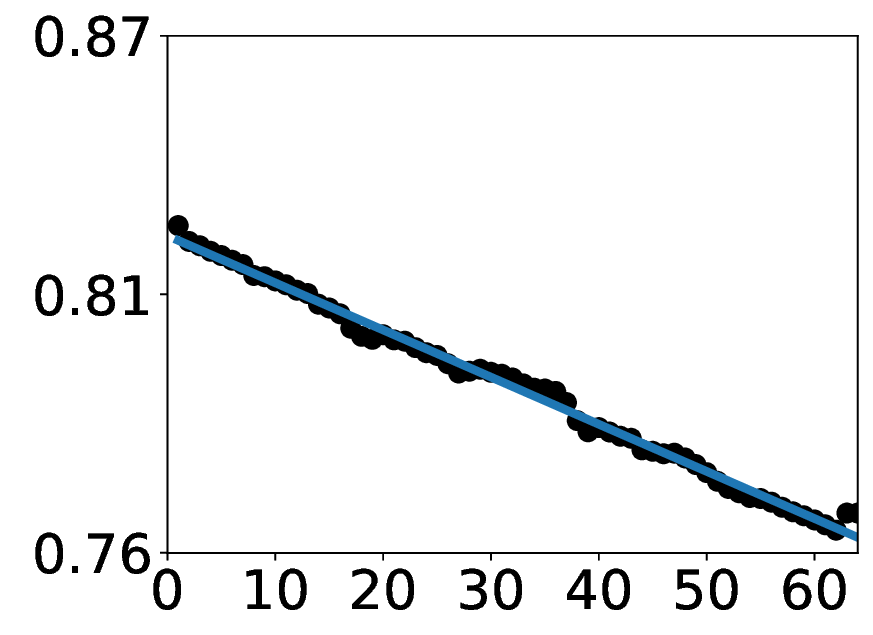}}
            \label{fig:mamba-2.8b-size}\hfill
		\caption{Illustration of the law of equi-learning with varying model sizes. The x-axis denotes the layer index, while the y-axis (log scale) shows the prediction residual (PR) as defined in Eq.~\ref{eq:PR}. Please note that the x axis and y axis share the same ranges within the same model series (same row). }
		\label{fig:model-size}
\end{figure*}

\textbf{Model scaling.} It is well established that larger models generally yield improved performance~\cite{kaplan2020scaling}. In this part, we explore the impact of model scaling on the observed law. As depicted in Fig.~\ref{fig:model-size}, our analysis reveals four consistent trends across three model series, each based on a distinct network architecture: (1) larger models exhibit lower PR values for first-layer token embeddings; (2) larger models demonstrate smaller PR values at the final layer, indicating enhanced next-token prediction capabilities; (3) larger models demonstrate an increased layer-wise decay ratio ($\rho$); and (4) larger models exhibit a reduced overall decay ratio, defined as the PR of last-layer token embeddings divided by the PR of first-layer token embeddings. The first trend may be attributed to the increased size or improved quality of first-layer token embeddings. The other trends suggest that while larger models typically display enhanced feature learning capabilities, resulting in more refined last-layer token embeddings, their performance at the individual layer level may be less effective compared to that of smaller models. These findings indicate that the law of equi-learning provides a more nuanced perspective on LLM behavior, particularly in the context of model scaling, offering insights that extend beyond the limitations of test loss alone.

\begin{figure*}[!htp]
    \captionsetup[subfigure]{labelformat=empty}
		\centering
   \rotatebox[y=1.5cm]{90}{\footnotesize NTP}\quad
        	\subfloat[BERT-base (uncased)]{
			\centering
		\includegraphics[scale=0.24]{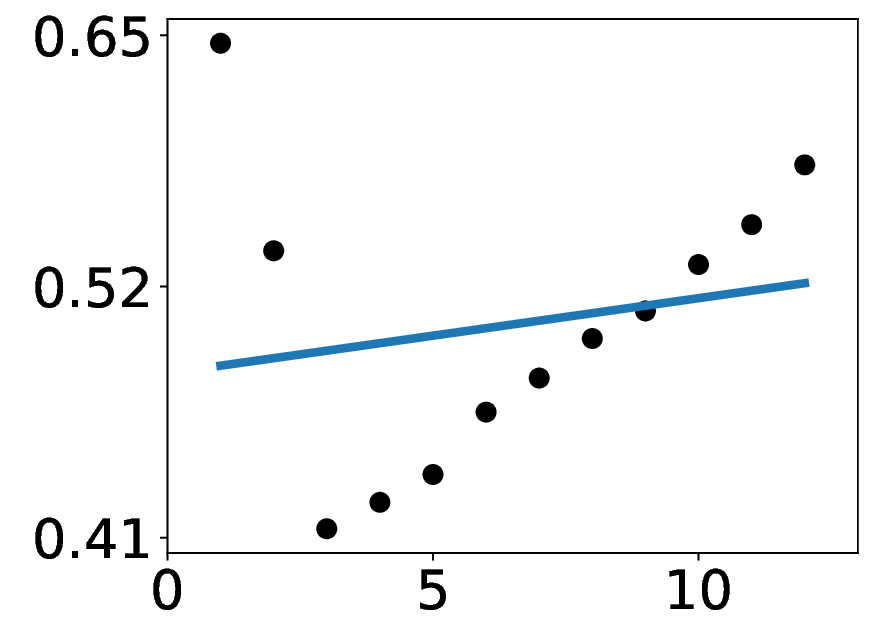}}
            \label{fig:bert-base-ntp}\hfill
            \subfloat[BERT-large (uncased)]{
			\centering
		\includegraphics[scale=0.24]{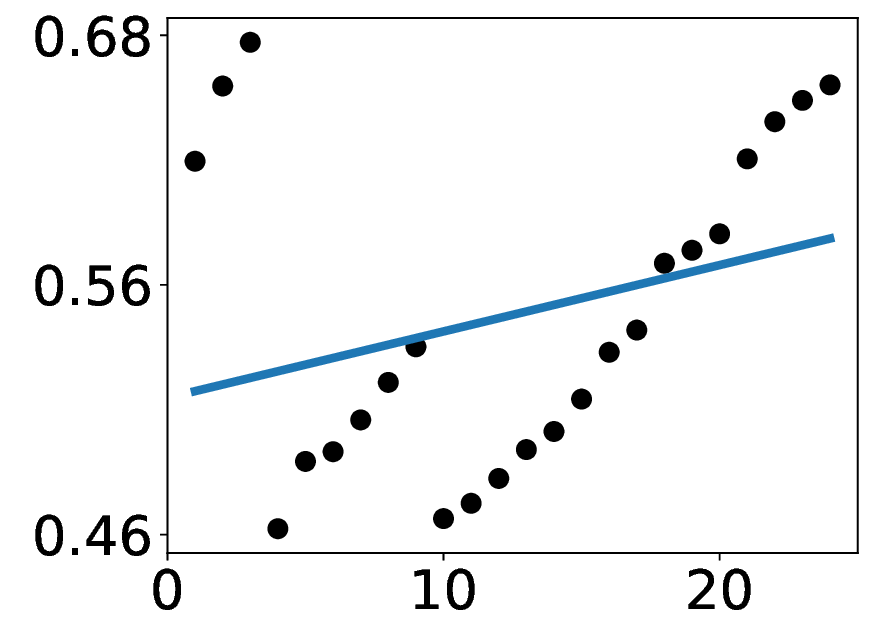}}
            \label{fig:bert-large-ntp}\hfill
     	\subfloat[RoBERTa-base]{
			\centering
		\includegraphics[scale=0.24]{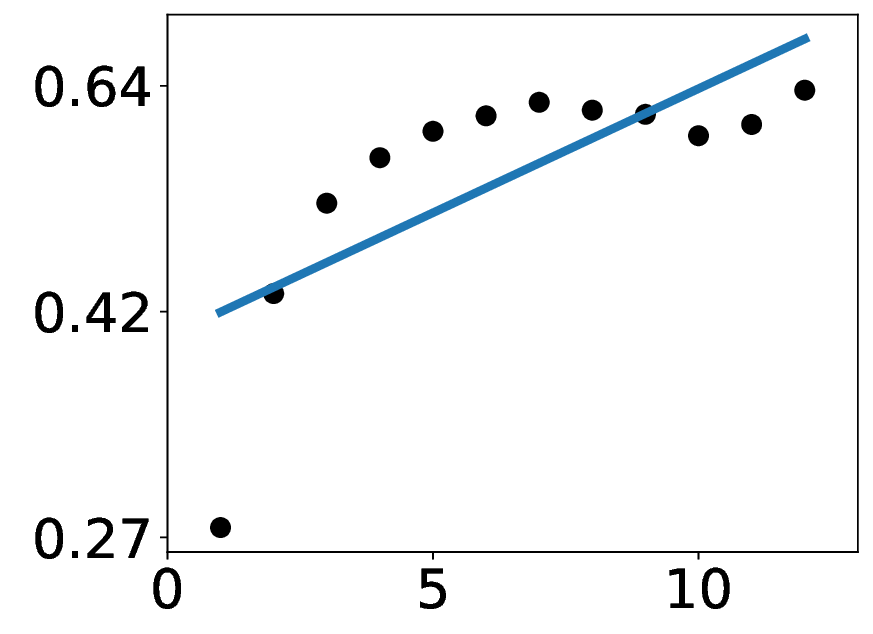}}
            \label{fig:roberta-base-ntp}\hfill
        \subfloat[RoBERTa-large]{
			\centering
		\includegraphics[scale=0.24]{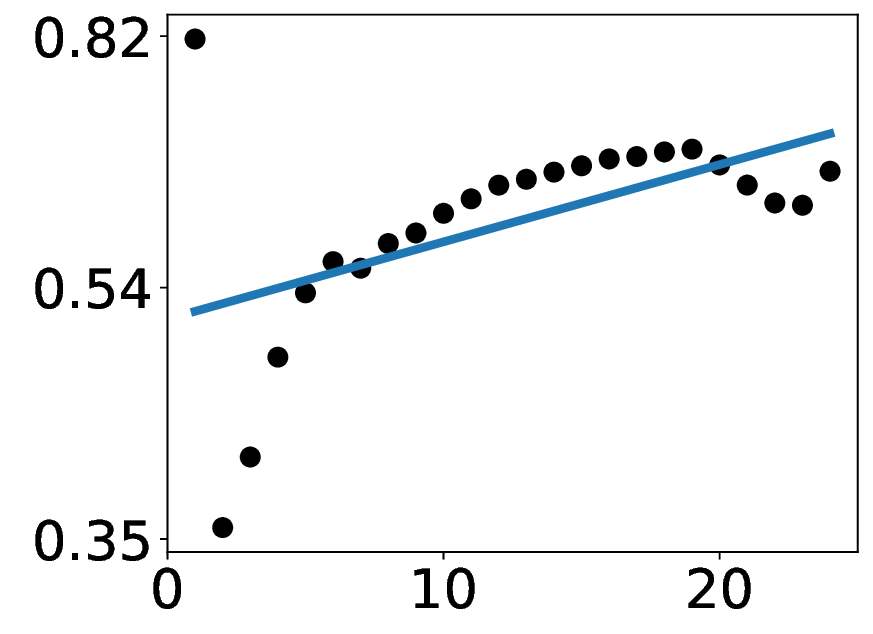}}
            \label{fig:roberta-large-ntp}\hfill

         \rotatebox[y=1.5cm]{90}{\footnotesize MLM}\quad
            \subfloat[BERT-base (uncased)]{
			\centering
		\includegraphics[scale=0.24]{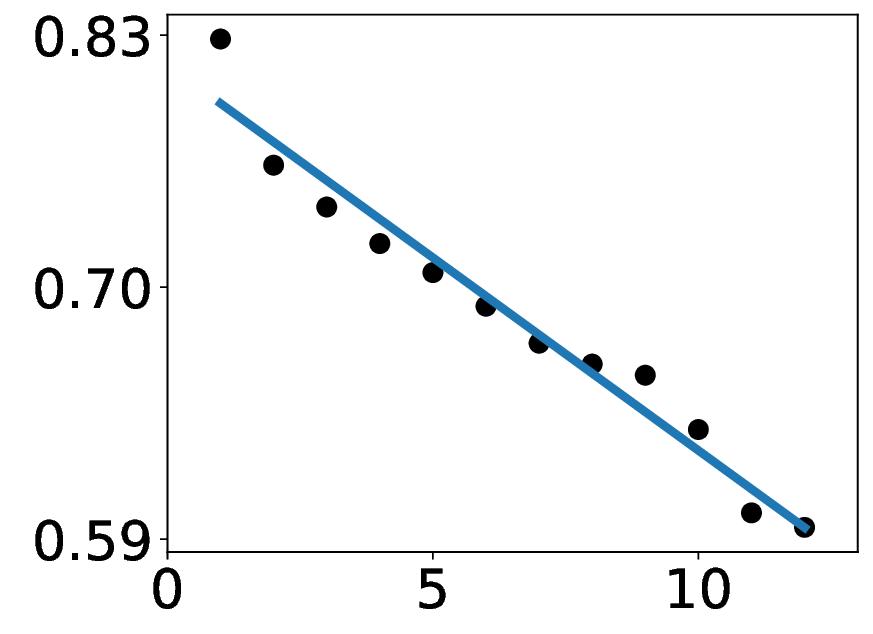}}
            \label{fig:bert-base-mlm}\hfill
            \subfloat[BERT-large (uncased)]{
			\centering
		\includegraphics[scale=0.24]{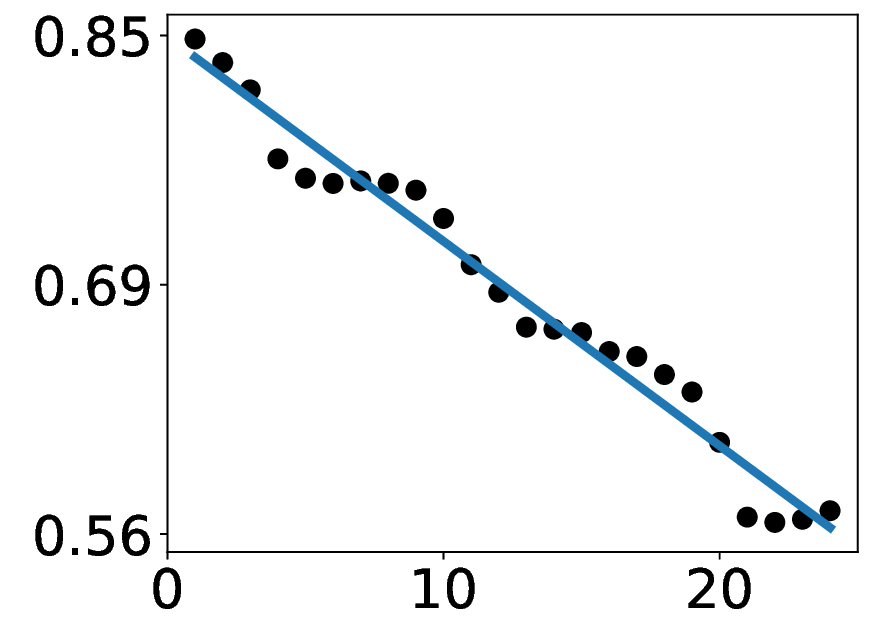}}
            \label{fig:bert-large-mlm}\hfill
            \subfloat[RoBERTa-base]{
			\centering
		\includegraphics[scale=0.24]{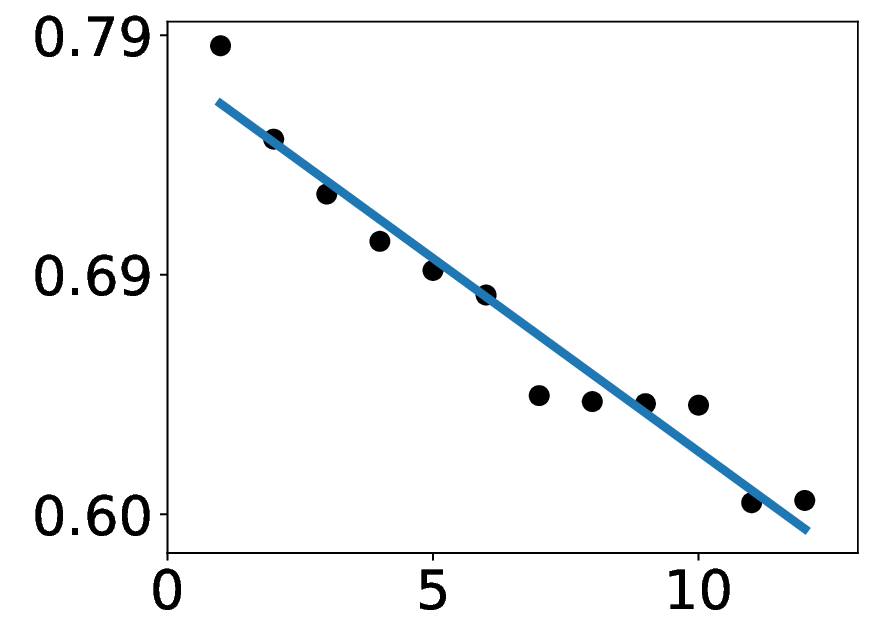}}
            \label{fig:roberta-base-mlm}\hfill
     	\subfloat[RoBERTa-large]{
			\centering
		\includegraphics[scale=0.24]{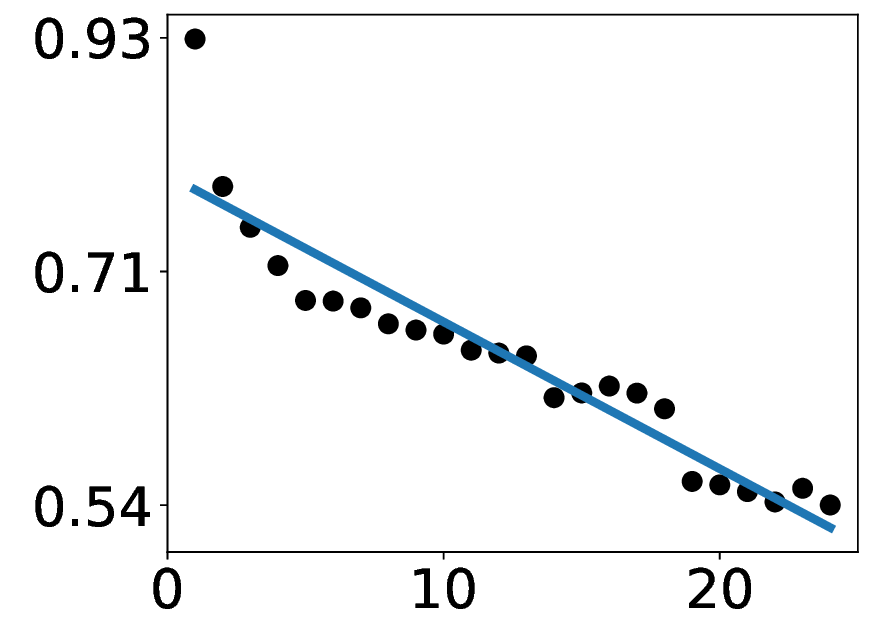}}
            \label{fig:roberta-large-mlm}\hfill

            \rotatebox[y=1.5cm]{90}{\footnotesize NTP}\quad
            \subfloat[T5-base]{
			\centering
		\includegraphics[scale=0.24]{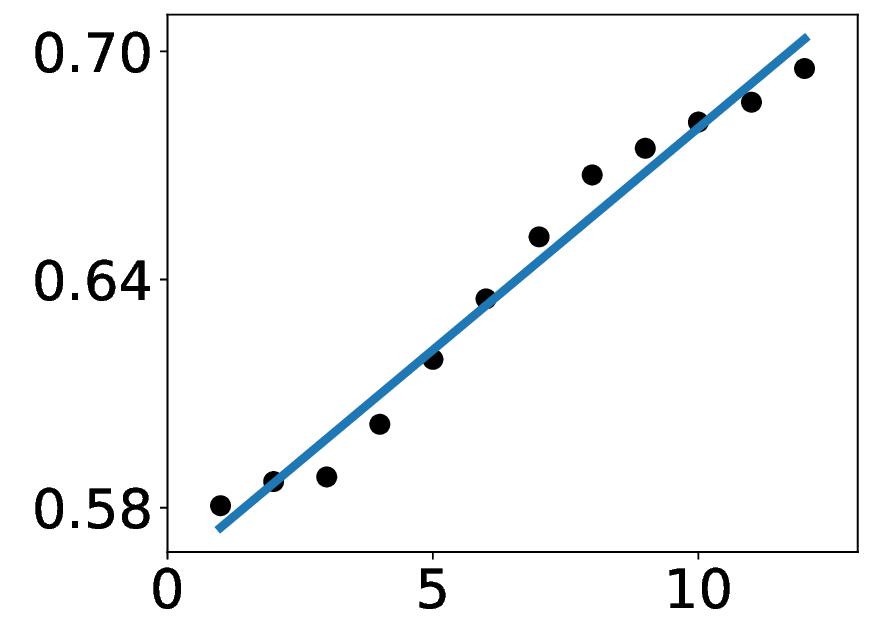}}
            \label{fig:t5-base-ntp}\hfill
            \subfloat[T5-large]{
			\centering
		\includegraphics[scale=0.24]{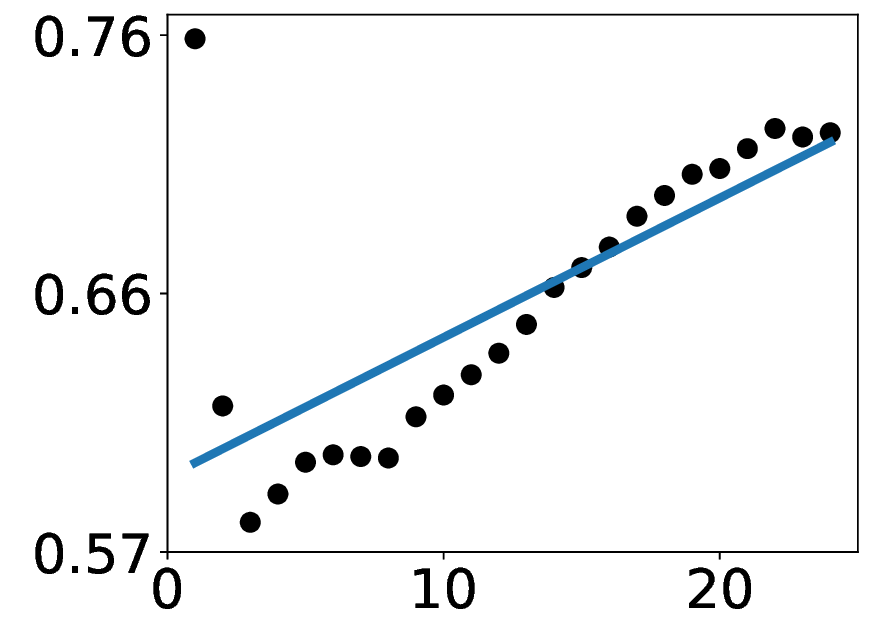}}
            \label{fig:t5-large-ntp}\hfill
            \subfloat[T5-3B]{
			\centering
		\includegraphics[scale=0.24]{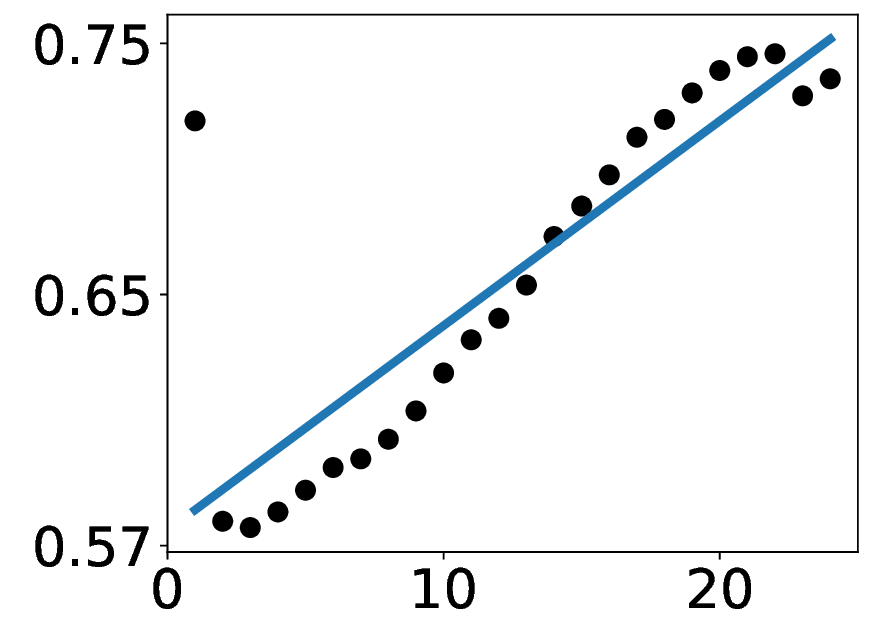}}
            \label{fig:t5-3b-ntp}\hfill
        \subfloat[T5-11B]{
			\centering
		\includegraphics[scale=0.24]{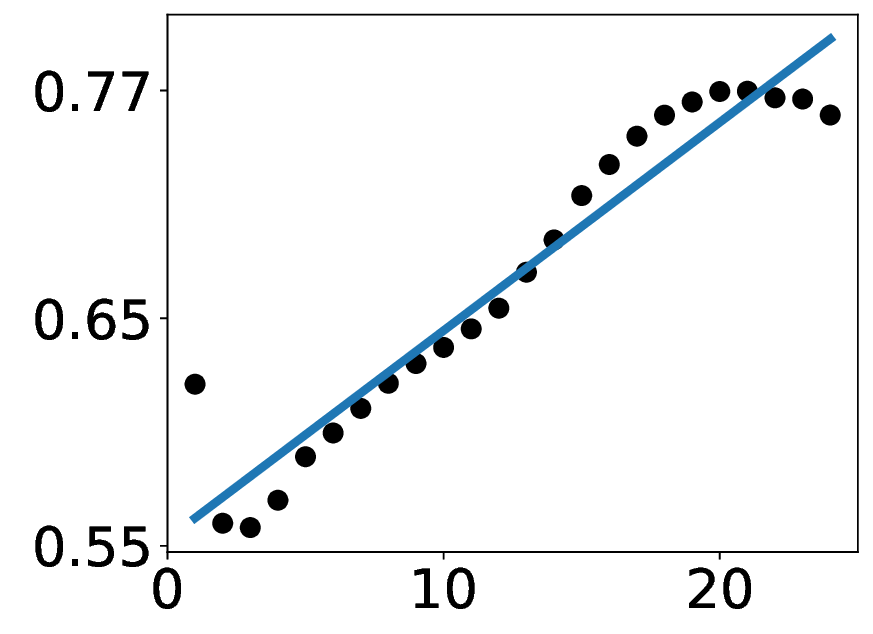}}
            \label{fig:t5-11b-ntp}\hfill

         \rotatebox[y=1.5cm]{90}{\footnotesize SC}\quad
        \subfloat[T5-base]{
			\centering
		\includegraphics[scale=0.24]{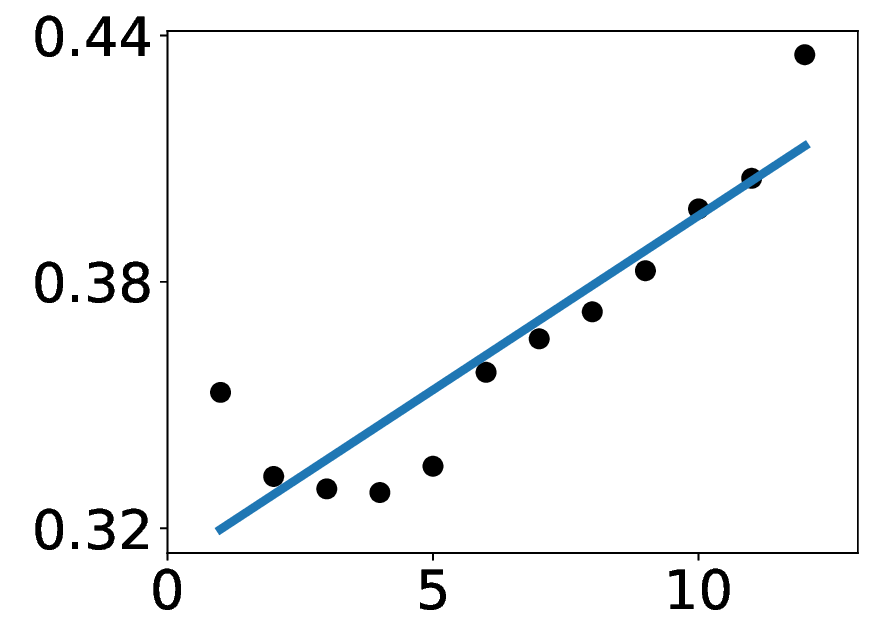}}
            \label{fig:t5-base-sc}\hfill
            \subfloat[T5-large]{
			\centering
		\includegraphics[scale=0.24]{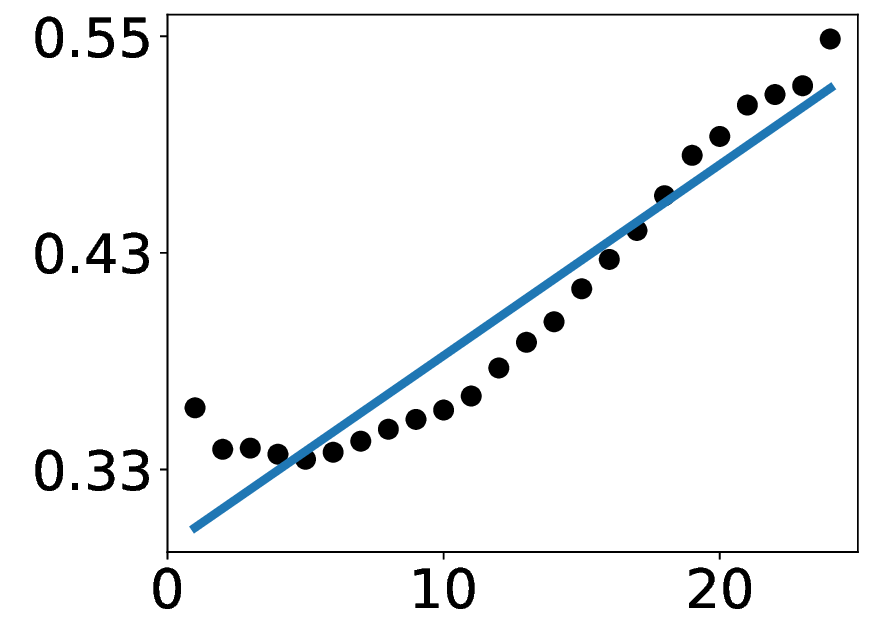}}
            \label{fig:t5-large-sc}\hfill
     	\subfloat[T5-3B]{
			\centering
		\includegraphics[scale=0.24]{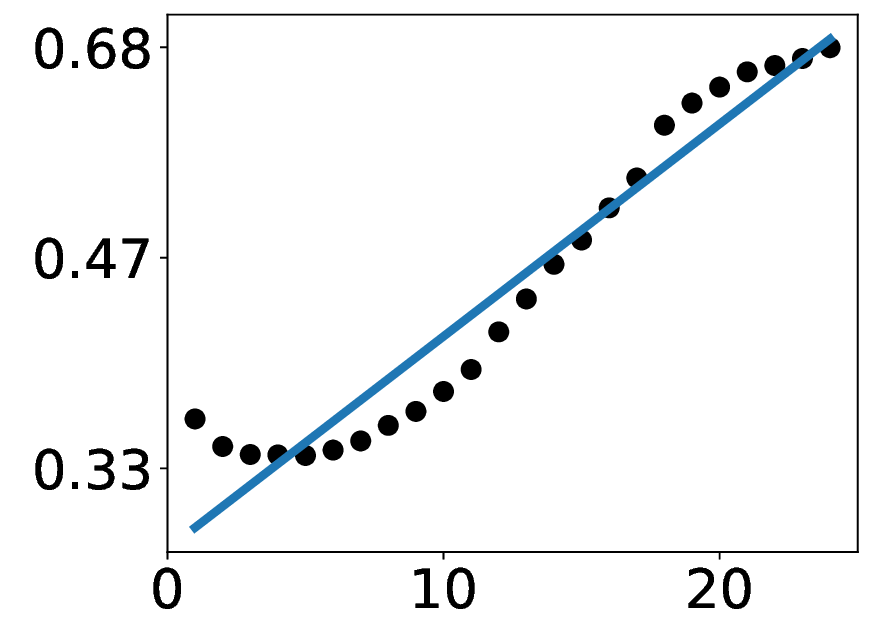}}
            \label{fig:t5-3b-sc}\hfill
             \subfloat[T5-11B]{
			\centering
		\includegraphics[scale=0.24]{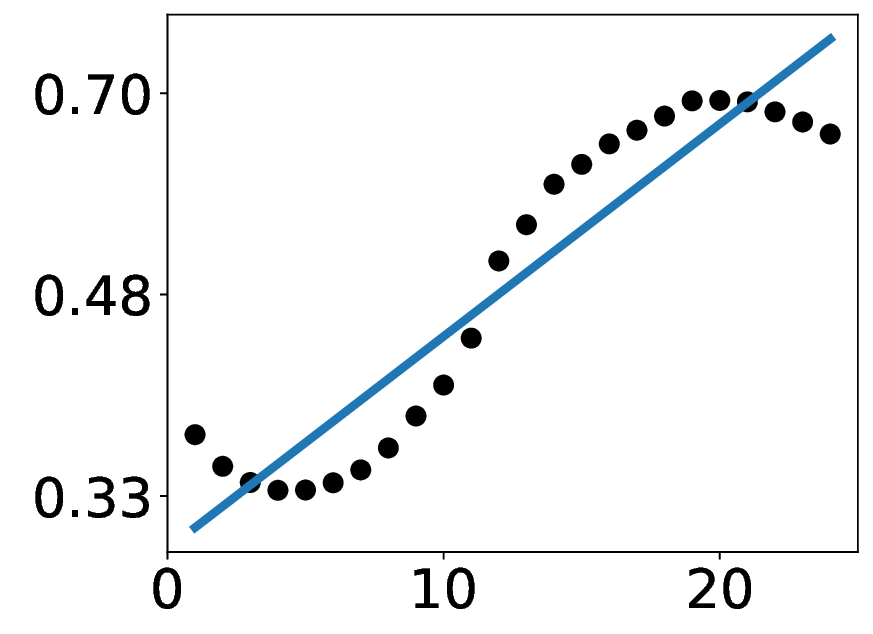}}
            \label{fig:t5-11b-sc}\hfill
		\caption{Different probing tasks are used for BERT (uncased), RoBERTa, and T5. Under the next-token prediction (NTP) task, the law of equi-learning does not appear because these models are pre-trained on different tasks. For BERT and RoBERTa, under their pre-training task, masked language modeling (MLM), the law of equi-learning appears but is somewhat noisy. In contrast, for T5, even with its pre-training task of span corruption (SC), the law of equi-learning does not appear. The x-axis denotes the layer index, while the y-axis (log scale) shows the prediction residual (PR) as defined in Eq.~\ref{eq:PR}. Note that the prediction target in PR is changed from the next token to the masked token for MLM and the corrupted span for SC. 
		}
		\label{fig:task-formulation}
\end{figure*}

\textbf{Pre-training task.} In addition to the prevalent next-token prediction task, various other pre-training tasks have been employed in the development of LLMs, including masked language modeling \cite{devlin2019bert} and span corruption \cite{raffel2020exploring} (for clarity, these three tasks are abbreviated as NTP, MLM, and SC, respectively). As shown in Fig.~\ref{fig:task-formulation}, different probing tasks are used to analyze the contextualized token embeddings of BERT \cite{devlin2019bert}, RoBERTa \cite{liu2019roberta}, and T5 \cite{raffel2020exploring}. Initially, we employ the mainstream NTP task to examine BERT, RoBERTa, and T5. However, the law of equi-learning does not emerge, possibly due to differences in their pre-training tasks compared to NTP. We then apply their respective pre-training tasks---MLM for BERT and RoBERTa, and SC for T5---to probe the models. For BERT and RoBERTa, under MLM, the law of equi-learning appears but is noisy, likely due to the complexity of the token replacement strategy used during pre-training. Specifically, 15\% of tokens were masked and replaced with 80\% [MASK] tokens, 10\% random tokens, and 10\% unchanged tokens for the purpose of predicting the original tokens. For T5, even with SC, the law does not emerge, which may be attributed to its encoder-decoder architecture. In SC, the decoder input often lacks natural coherence and relies heavily on the encoder input, while the decoder's cross-attention layer might further complicate the learning of contextualized token embeddings. These findings suggest that the choice of pre-training task is critical for the emergence of the law of equi-learning, with more naturalistic tasks potentially facilitating its appearance. This may provide supporting evidence for the superiority of NTP, the currently dominant pre-training task.

\begin{figure*}[!t]
    \captionsetup[subfigure]{labelformat=empty}
		\centering
     \rotatebox[y=1.5cm]{90}{\footnotesize GPT-2 XL}\quad
            \subfloat[Previous Token]{
			\centering
		\includegraphics[scale=0.24]{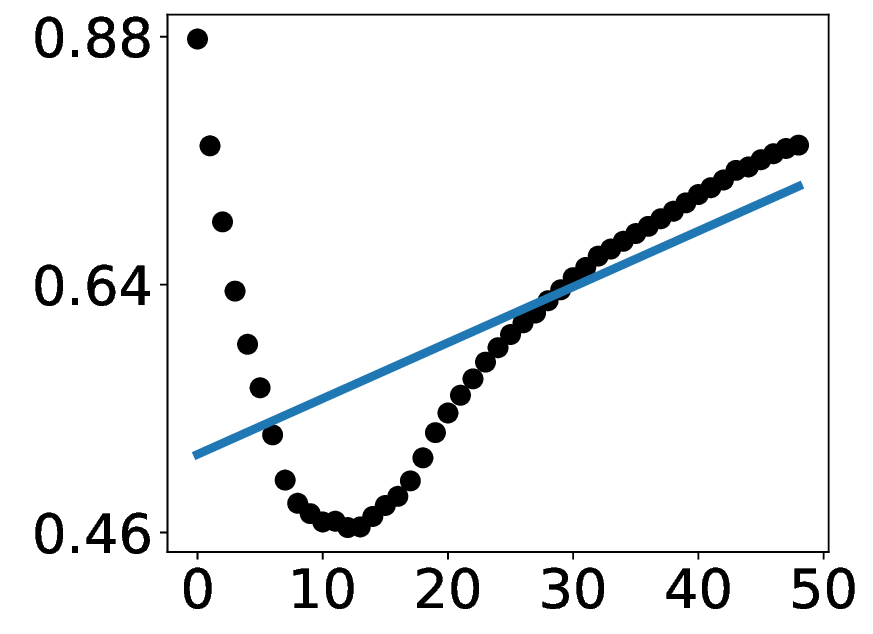}}
            \label{fig:gpt2-xl-w(-1)}\hfill
     	\subfloat[Current Token]{
			\centering
		\includegraphics[scale=0.24]{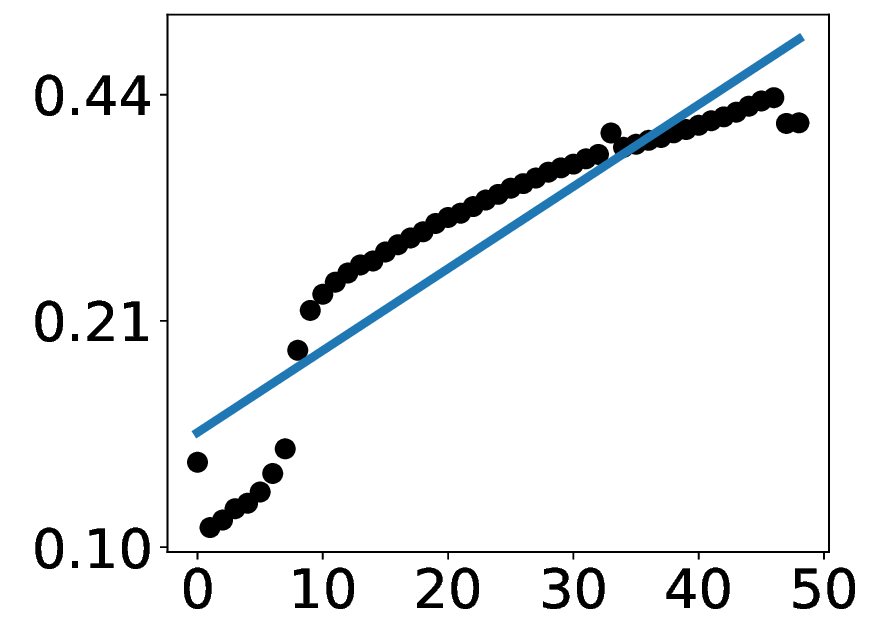}}
            \label{fig:gpt2-xl-w(0)}\hfill
        \subfloat[Next Token]{
			\centering
		\includegraphics[scale=0.24]{figures/gpt2-xl_bookcorpus_pretrained_features_size=3000_seed=666.eps}}
            \label{fig:gpt2-xl-w(+1)}\hfill
            \subfloat[Next Next Token]{
			\centering
		\includegraphics[scale=0.24]{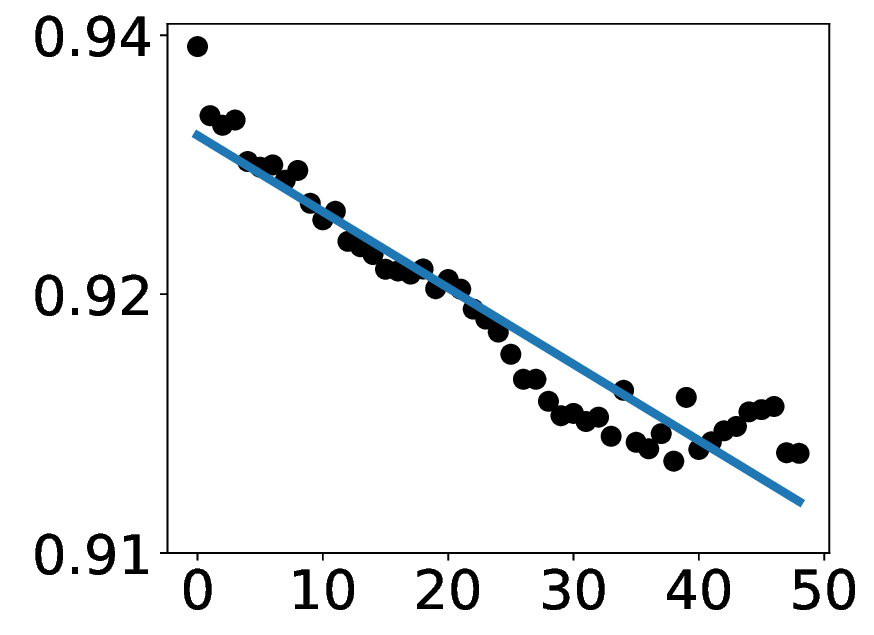}}
            \label{fig:gpt2-xl-w(+2)}\hfill

        \rotatebox[y=1.5cm]{90}{\footnotesize Llama-3-8B-Instruct}\quad
            \subfloat[Previous Token]{
			\centering
		\includegraphics[scale=0.24]{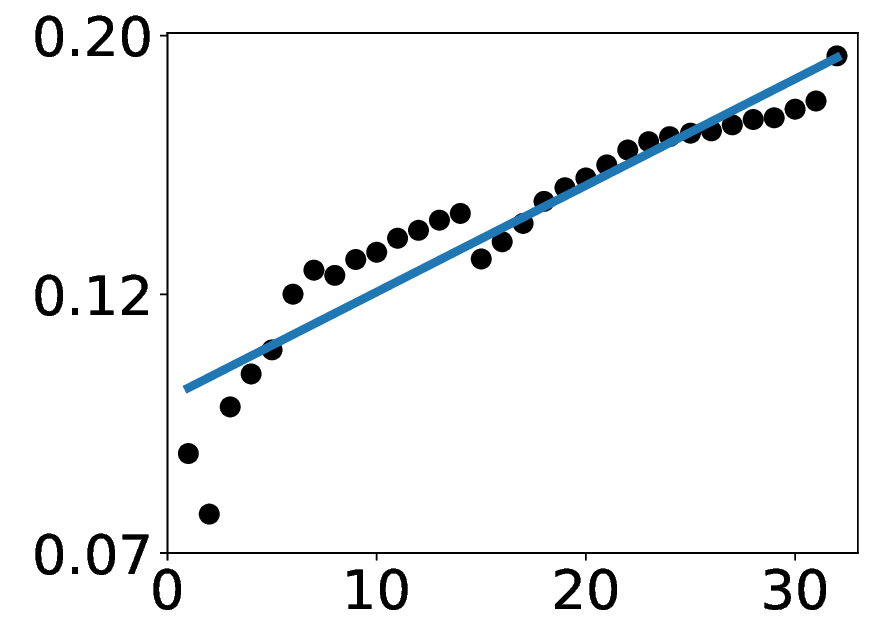}}
            \label{fig:llama-3-8b-it-w(-1)}\hfill
     	\subfloat[Current Token]{
			\centering
		\includegraphics[scale=0.24]{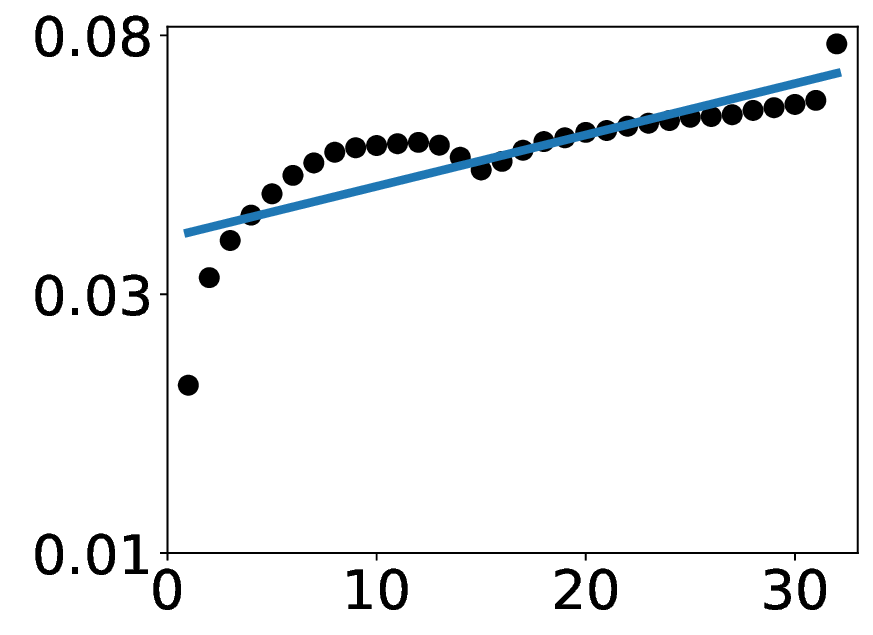}}
            \label{fig:llama-3-8b-it-w(0)}\hfill
        \subfloat[Next Token]{
			\centering
		\includegraphics[scale=0.24]{figures/Meta-Llama-3-8B-Instruct_bookcorpus_pretrained_features_size=3000_seed=666.eps}}
            \label{fig:llama-3-8b-it-w(+1)}\hfill
            \subfloat[Next Next Token]{
			\centering
		\includegraphics[scale=0.24]{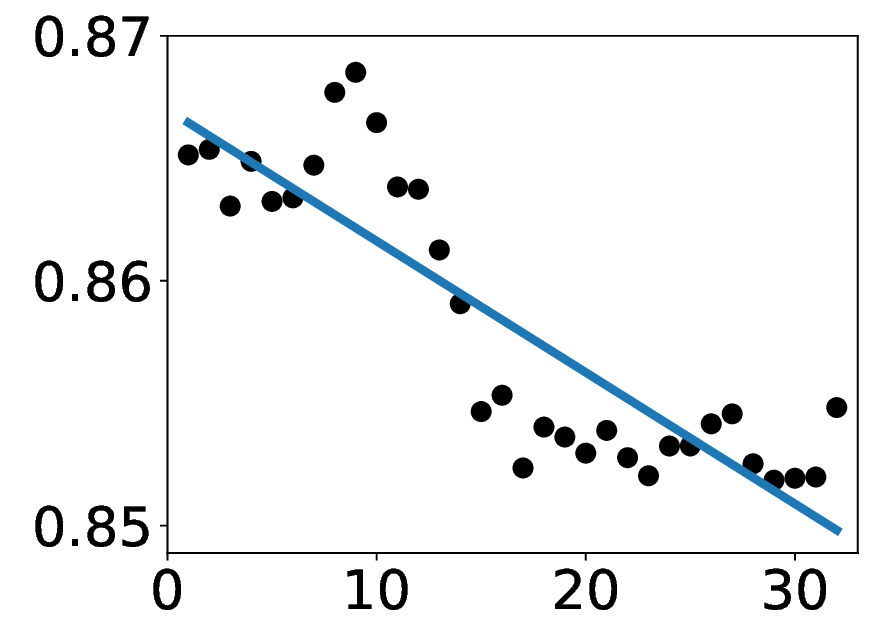}}
            \label{fig:llama-3-8b-it-w(+2)}\hfill

            \rotatebox[y=1.5cm]{90}{\footnotesize RWKV-Raven-14B}\quad
            \subfloat[Previous Token]{
			\centering
		\includegraphics[scale=0.24]{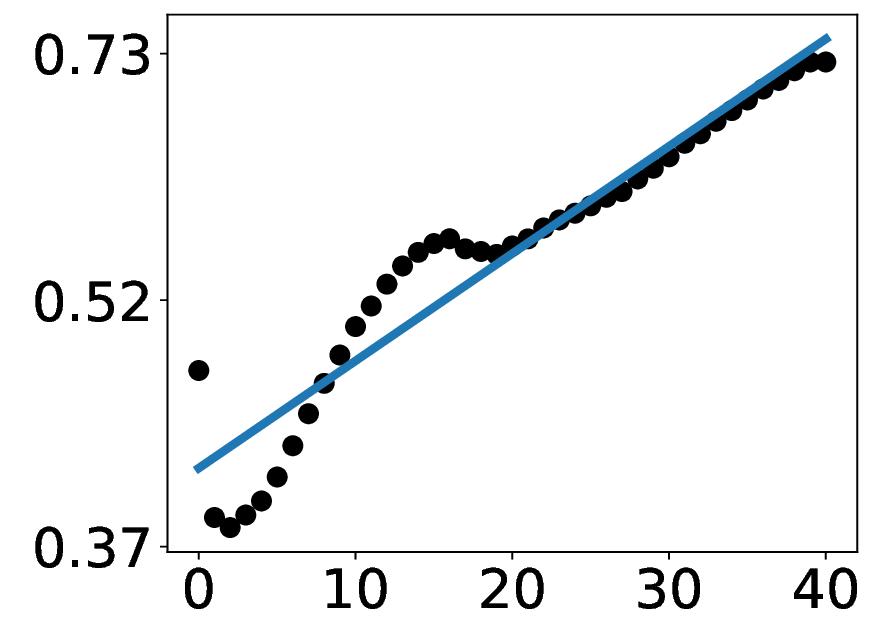}}
            \label{fig:rwkv-raven-14b-w(-1)}\hfill
     	\subfloat[Current Token]{
			\centering
		\includegraphics[scale=0.24]{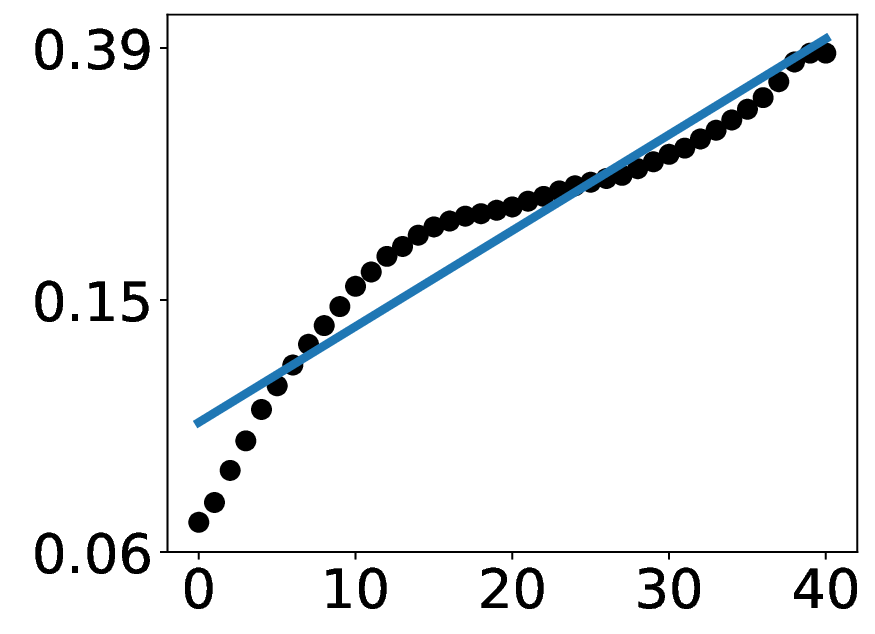}}
            \label{fig:rwkv-raven-14b-w(0)}\hfill
        \subfloat[Next Token]{
			\centering
		\includegraphics[scale=0.24]{figures/rwkv-raven-14b_c4_pretrained_features_size=200_seed=666.eps}}
            \label{fig:rwkv-raven-14b-w(+1)}\hfill
            \subfloat[Next Next Token]{
			\centering
		\includegraphics[scale=0.24]{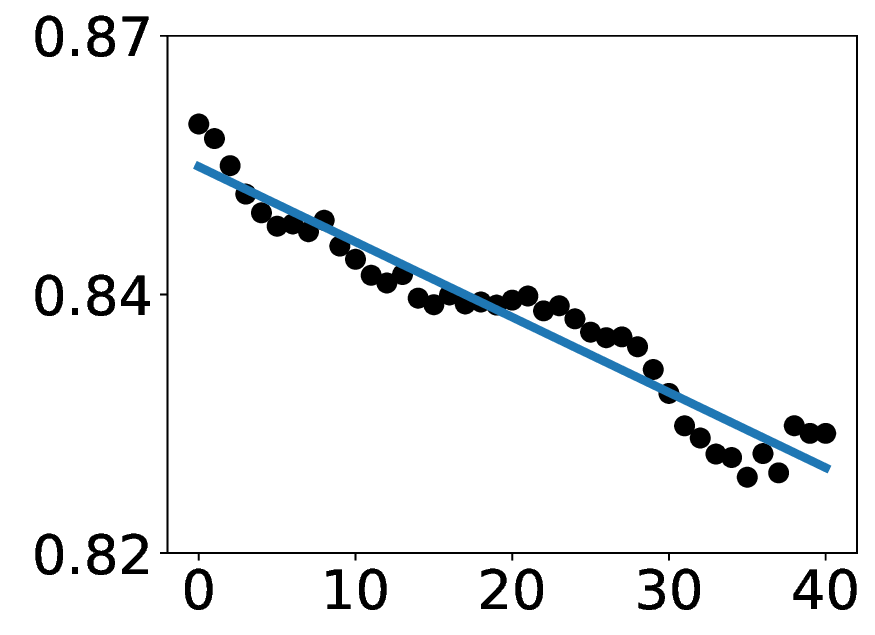}}
            \label{fig:rwkv-raven-14b-w(+2)}\hfill

            \rotatebox[y=1.5cm]{90}{\footnotesize Mamba-2.8B}\quad
            \subfloat[Previous Token]{
			\centering
		\includegraphics[scale=0.24]{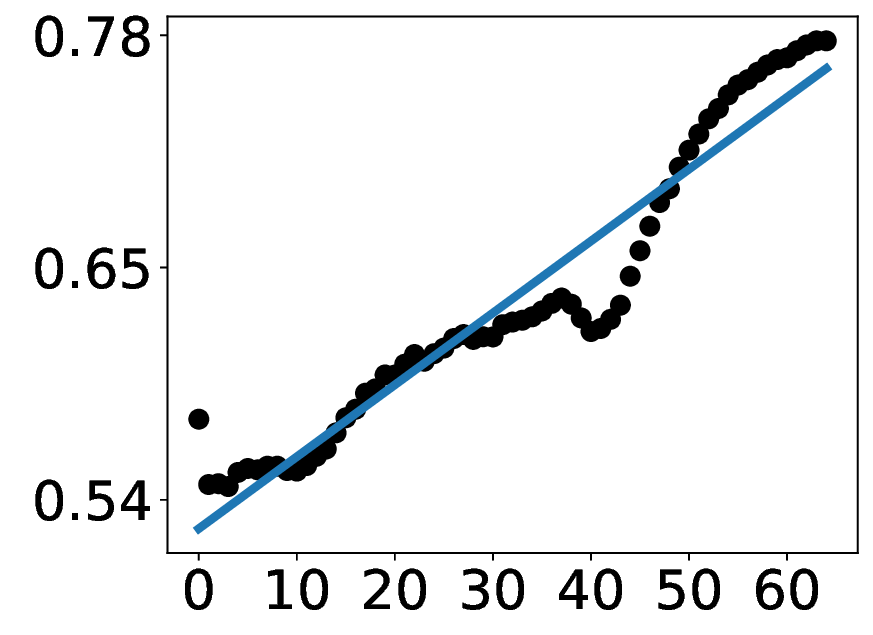}}
            \label{fig:mamba-2.8b-w(-1)}\hfill
     	\subfloat[Current Token]{
			\centering
		\includegraphics[scale=0.24]{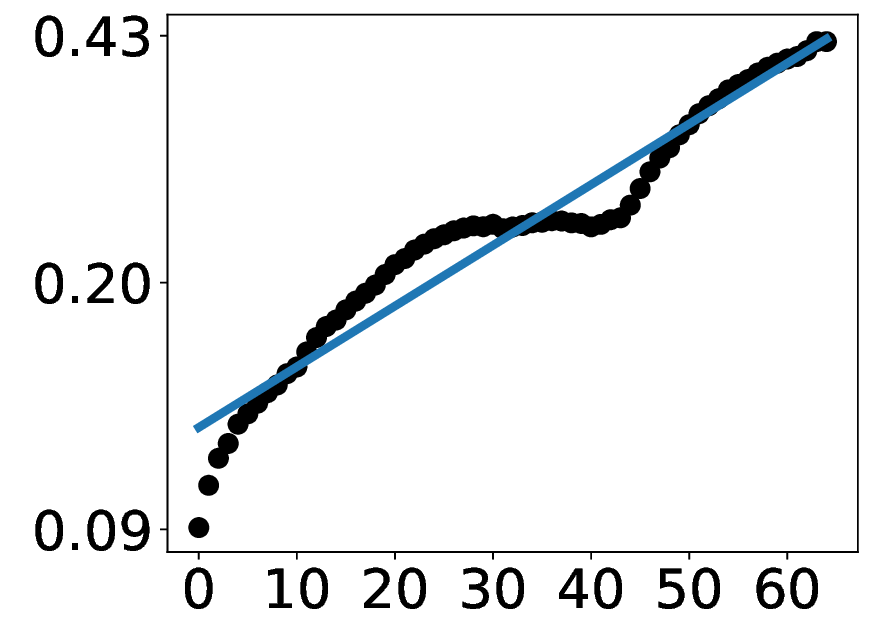}}
            \label{fig:mamba-2.8b-w(0)}\hfill
        \subfloat[Next Token]{
			\centering
		\includegraphics[scale=0.24]{figures/mamba-2.8b-hf_bookcorpus_pretrained_features_size=3000_seed=666.eps}}
            \label{fig:mamba-2.8b-w(+1)}\hfill
            \subfloat[Next Next Token]{
			\centering
		\includegraphics[scale=0.24]{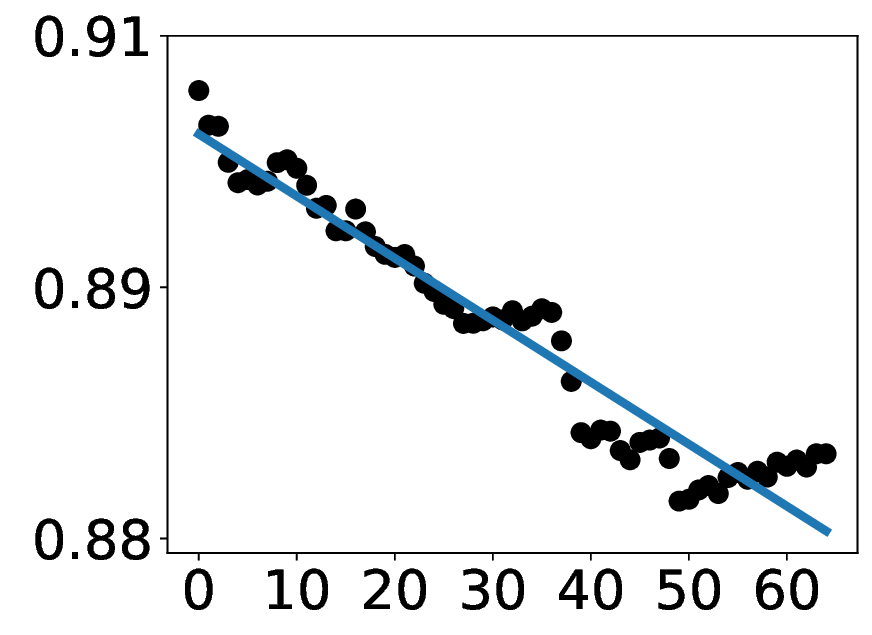}}
            \label{fig:mamba-2.8b-w(+2)}\hfill
		\caption{The contextualized token embeddings of the current token ($x_t$) at each layer are utilized to predict various tokens in the sequence, ranging from previous token ($x_{t-1}$) to next next token ($x_{t+2}$), including the default next token ($x_{t+1}$). It is observed that LLMs tend to forget previous information (positive correlations between the PR and the layer index), including the current token, and improve their prediction of future tokens (negative correlations between the PR and the layer index) after the learning of contextualized token embeddings across layers. The x-axis denotes the layer index, while the y-axis (log scale) shows the prediction residual (PR) as defined in Eq.~\ref{eq:PR}. Note that the prediction target in PR is changed from the next token to the previous token, current token, or next next token in settings other than next-token prediction.
		}
		\label{fig:token-prediction}
\end{figure*}

\textbf{Information flow.} Controlling the flow of information is crucial for effective feature learning in sequential data \cite{hochreiter1997long, chung2014empirical}. To elucidate the dynamics of information flow in LLMs across different architectures, we analyze the contextualized token embeddings of the current token ($x_t$) at each layer and their ability to predict other tokens within the sequence, ranging from the previous token ($x_{t-1}$) to the next-next token ($x_{t+2}$), including the default next token ($x_{t+1}$). Specifically, the embedding of the current token at layer \( \ell \), denoted as \( \mathbf{h}_{t,\ell} \), is used to predict not only the next token (\( x_{t+1} \)), but also the previous token (\( x_{t-1} \)), the current token (\( x_t \)), and the next next token (\( x_{t+2} \)) when computing \( \text{PR}_\ell \). As depicted in Fig.~\ref{fig:token-prediction}, our results reveal a clear pattern: following the learning of contextualized token embeddings across layers, LLMs exhibit a tendency to forget prior information, including the current token itself, as indicated by the positive correlations between PR and layer index. In contrast, the prediction of future tokens improves, evidenced by negative correlations between PR and layer index. These findings suggest that as the learning of contextualized token embeddings progresses from lower to higher layers, LLMs increasingly discard historical information while simultaneously enhancing their predictive capabilities for upcoming tokens.

\section{Discussion}
\label{sec:disc}

Despite the extensive research on structures within pre-trained LLMs, many have found it challenging to identify precise, quantitative laws governing their internal dynamics. In this work, we challenge this view by introducing the law of equi-learning, which describes how contextualized token embeddings evolve from the first to the last layer. This law, which is both quantitative and precise, has been consistently observed across various architectures, including Transformer, Mamba, and RWKV. Its emergence provides crucial insights into the training and interpretation of LLMs, offering new perspectives that deepen our understanding of their internal mechanisms.

The significance of the equi-learning law lies in its potential to refine the development and application of LLMs. An open question is how the decay ratio $\rho$ depends on factors such as model depth and pre-training data. Understanding this dependence could lead to the development of more efficient LLMs by minimizing $\rho^{L-1}$, the overall decay ratio, in the equi-learning law. The emergence of this law specifically under the PR metric defined in \eqref{eq:PR}, rather than alternative metrics (see Fig.\ \ref{fig:separation-fuzziness} in the Supplementary Materials), warrants further investigation. We hypothesize that this specificity may arise from the PR metric's incorporation of token indices derived from byte pair encoding \cite{sennrich2016neural}, which might capture structural information beyond simple classification metrics. Nevertheless, we believe that the law could be extended beyond the PR metric, though this remains an avenue for future research. The law's eventual emergence also suggests the possibility of setting different learning rates across layers to accelerate convergence to the equilibrium described by the law. Moreover, preserving the equi-learning law during model pruning and fine-tuning may yield practical benefits, potentially through the preservation of certain weights or the use of techniques like LoRA~\cite{hu2022lora}.

A central open question is to uncover the mechanism underlying the equi-learning law. Related phenomena have been analytically derived in deep linear networks (DLNs) with linearly separable data \cite{wang2023understanding}, but the strong assumptions---such as the absence of nonlinearity and simplified data---limit their relevance to real-world LLMs. A spring-block analogy has been proposed to illustrate a similar law in deep neural networks (DNNs) \cite{shi2024spring}. However, this framework remains primarily heuristic, as it does not establish a concrete correspondence between fundamental physical elements---such as elastic potential energy and friction---and specific components within DNNs. Furthermore, it does not address the substantial architectural and functional disparities between DNNs and modern LLMs. A formal derivation of the equi-learning law in LLMs will likely require a deeper understanding of model dynamics and the structure of data, aligned with recent geometric approaches to analyzing DNNs and LLMs \cite{papyan2020prevalence, wu2024linguistic, chan2022redunet, yu2024white}, which we leave for future work.

Broadly, the equi-learning law could be leveraged in transfer learning, particularly when the bottom layers are frozen while the top layers are re-trained to adapt models to new domains. Additionally, our preliminary results in the Supplementary Materials suggest that higher-quality pre-training data may require high-quality probing data to facilitate the emergence of the equi-learning law. This finding implies the potential for using this law to improve the evaluation of LLM capabilities across different tasks. For interpretability, a welcome advance would be the development of new methodologies that consider the collective contributions of all layers, rather than just a few, in interpreting the predictions of LLMs.

\section*{Acknowledgments}
We would like to thank Zhehang Du, Hangyu Lin, Cheng Shi, Peng Wang, and Yuan Yao for stimulating discussions. We are grateful to the two anonymous reviewers whose comments helped improve the presentation of this paper. This research was supported in part by NSF grant DMS-2310679 and Wharton AI for Business.


\bibliography{ref}
\bibliographystyle{plain}





\clearpage
\appendix
\setcounter{figure}{0}
\renewcommand{\figurename}{Fig.}
\renewcommand{\thefigure}{S\arabic{figure}}
\renewcommand{\thetable}{S\arabic{table}}

\section*{Supplementary Text}
\label{sec:experiments}

This section outlines the general experimental setup, distinctive configurations for main text figures, and additional results, with comprehensive details available in our code repository\footnote{Our code is publicly available at \url{https://github.com/HornHehhf/LLM-ELL}.}.

\subsection*{General Setup}
\label{subsec:general-setup}

In this subsection, we detail the general experimental setup utilized throughout this study.

\textbf{LLMs.} In this study, we focus on autoregressive LLMs, where the objective is to predict the subsequent token in a sequence, constrained to attend solely to preceding tokens. Formally, the model takes as input a sequence of discrete tokens \( x_1, x_2, \dots, x_t \in \mathcal{V}^t \), where \( \mathcal{V} \) denotes the vocabulary specific to the model. These tokens are obtained via a tokenizer applied to raw text; for instance, the sentence \textit{``We love Physics.''} is tokenized by the Llama 3 tokenizer into four tokens: ``We'', `` love'', `` Physics'', and ``.'' Vocabulary sizes vary across models (e.g., 32K for Llama 2 and 128K for Llama 3). Each token \( x_i \) is mapped to an initial embedding vector \( \mathbf{h}_{i,0} \) via a learned embedding table. These embeddings are then propagated through a stack of model layers---typically transformer layers---resulting in a set of contextualized token embeddings \( \mathbf{h}_{i,\ell} \) at each layer \( 1 \le \ell \le L\). Due to the autoregressive nature of the model, the contextualized token embeddings \( \mathbf{h}_{i,\ell} \) is computed using only information from position \( i \) and all positions preceding it, specifically from the previous layer’s outputs \( \{ \mathbf{h}_{j,\ell-1} \mid 1 \le j \le i \} \). This ensures that the model does not access future tokens. The last-layer embedding of the current token, \( \mathbf{h}_{t,L} \) (also denoted \( \mathbf{h}_{t,\text{last}} \)), is used to predict the next token \( x_{t+1} \) by projecting it into vocabulary space and applying a softmax to produce a probability distribution over the vocabulary. This autoregressive decoding procedure is iteratively applied at each position \( 1 \le t \le T-1 \), where \( T \) denotes the length of the input token sequence.

\textbf{Prediction residual (PR).} To assess the capability of contextualized token embeddings in predicting the next token, we calculate the fraction of variance unexplained (FVU) by a linear regression model predicting the next token. Notably, this metric is similar to the concurrent measure proposed by \cite{shi2024spring}, specifically the Root Mean Squared Error (RMSE) of the optimal linear regressor based on the features in MLPs. In their study, RMSE was shown to reproduce a similar noise–nonlinearity phase diagram in MLP training under regression, extending phenomena originally observed in classification to the regression setting. This finding supports the potential of PR as a meaningful measure of representation quality.

\textbf{Layer normalization.} In pre-layer normalization (pre-LN) models, default initialized layer normalization is applied to all layers except the last-layer token embeddings. This is because layer normalization is moved to the input of each sub-block, with an additional layer normalization added after the final self-attention block \cite{radford2019language}. Given that different models utilize distinct forms of layer normalization—such as LayerNorm \cite{ba2016layer} in GPT-2 and RMSNorm \cite{zhang2019root} in Llama-1—we will apply the specific layer normalization technique used by each model to normalize its contextualized token embeddings. Throughout this paper, nearly all models are pre-LN models, with the exceptions of GPT-1, BERT, and RoBERTa. 

\textbf{Probing datasets.} For our experiments, we consider eight distinct probing datasets: BookCorpus, C4, OpenWebText, Wikipedia, peS2o, Pile, Redpajama, and OSCAR. We sampled sentences based on their average length, extracting $3,000$ sentences from BookCorpus, $200$ from C4, and $100$ from each of the remaining datasets. For consistency, sentences were truncated to a maximum length of $512$ tokens across all datasets, except for BookCorpus. Unless otherwise specified, we will utilize the probing dataset that exhibits the strongest Pearson correlation in next-token prediction for each LLM.

\subsection*{Detailed Experimental Settings}
\label{subsec:experimental-details}

In this subsection, we show detailed experimental settings for the figures in the main text.

\textbf{Large language models.} As depicted in Fig.~\ref{fig:law}, the law of equi-learning is observed in various open-source large language models. Please note that phi-1 was excluded from our analysis, as it is a LLM specifically designed for code. For each model, we evaluate the largest size that can be executed on a local machine equipped with two L40S GPUs, each possessing 48 GB of memory. The corresponding model sizes are as follows: 117M for GPT-1, 1.5B for GPT-2, 13B for Llama-1, 13B for Llama 2 and Llama 2-Chat, 8B for Llama 3 and Llama 3 Instruct, 7B for three versions of Mistral 7B and Mistral 7B-Instruct (v0.1, v0.2, v0.3), 1.3B for phi-1.5, 2.7B for phi-2, 14B for phi-3 (phi-3-medium) with varying context lengths (4K, 128K), 14B for RWKV and RWKV-Raven, and 2.8B for Mamba. The corresponding probing datasets used are as follows: BookCorpus for GPT-1, BookCorpus for GPT-2, peS2o for Llama-1, peS2o for Llama 2 and Llama 2-Chat, BookCorpus for Llama 3 and Llama 3 Instruct, C4 for three versions of Mistral 7B and Mistral 7B-Instruct (v0.1, v0.2, v0.3), BookCorpus for phi-1.5, BookCorpus for phi-2, C4 for phi-3 (phi-3-medium) with varying context lengths (4K, 128K), C4 for RWKV and RWKV-Raven, and BookCorpus for Mamba. Notably, reinforcement learning from human feedback (RLHF) \cite{ouyang2022training} does not significantly impact the law, as demonstrated by Llama-2-13B, Llama-3-8B, Mistral-7B-v0.1, Mistral-7B-v0.2, Mistral-7B-v0.3, and RWKV, along with their fine-tuned versions, as illustrated in Fig.~\ref{fig:law}.

\begin{figure*}[!htp]
    \captionsetup[subfigure]{labelformat=empty}
		\centering
            \subfloat[Layer=1]{
			\centering
	\includegraphics[scale=0.24]{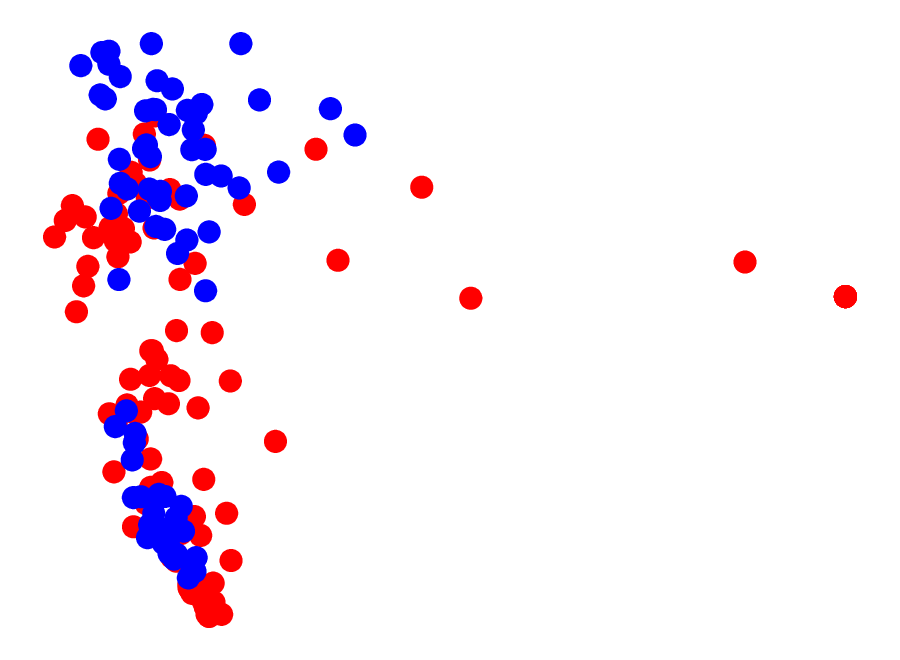}}
            \label{fig:they-them-layer1}\hfill
     	\subfloat[Layer=2]{
			\centering
	\includegraphics[scale=0.24]{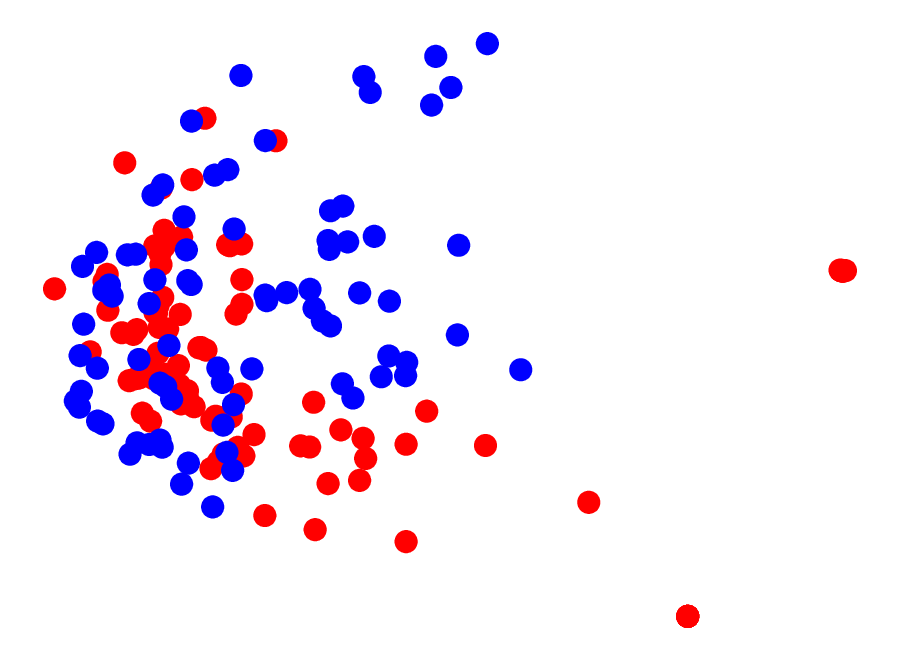}}
            \label{fig:they-them-layer2}\hfill
        \subfloat[Layer=3]{
			\centering
	\includegraphics[scale=0.24]{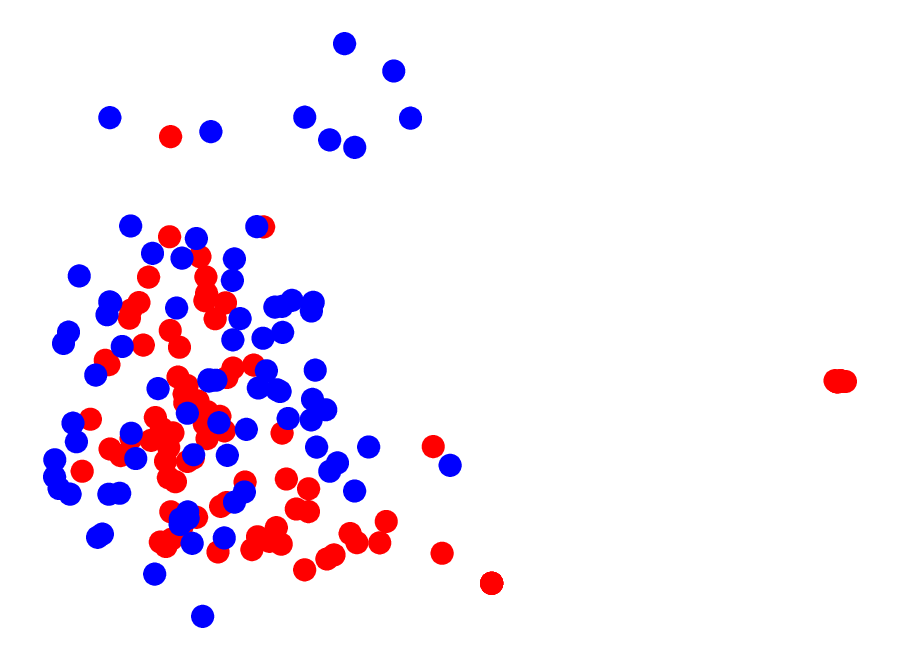}}
            \label{fig:they-them-layer3}\hfill
            \subfloat[Layer=4]{
			\centering
	\includegraphics[scale=0.24]{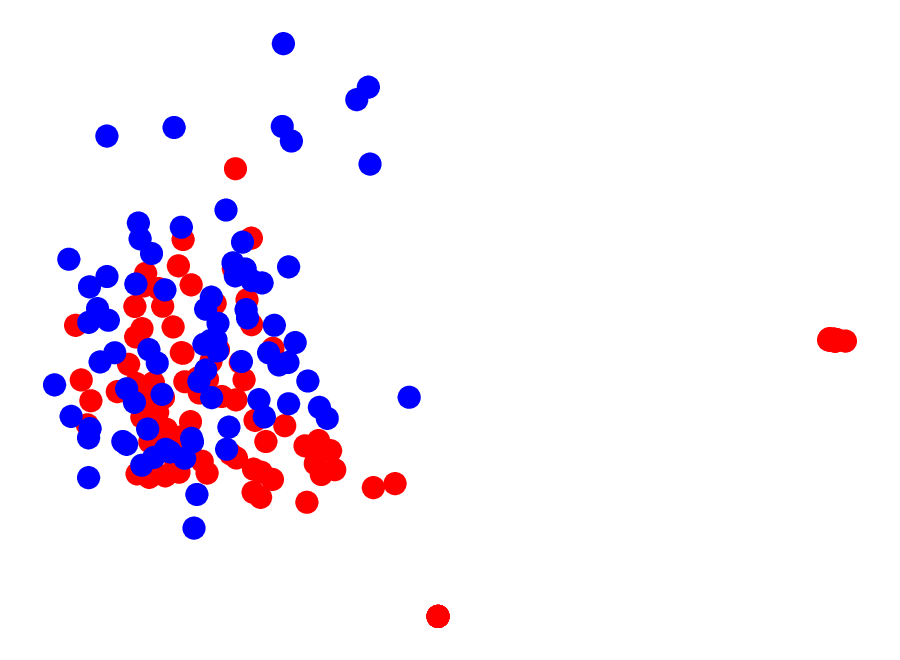}}
            \label{fig:they-them-layer4}
            
     	\subfloat[Layer=5]{
			\centering
	\includegraphics[scale=0.24]{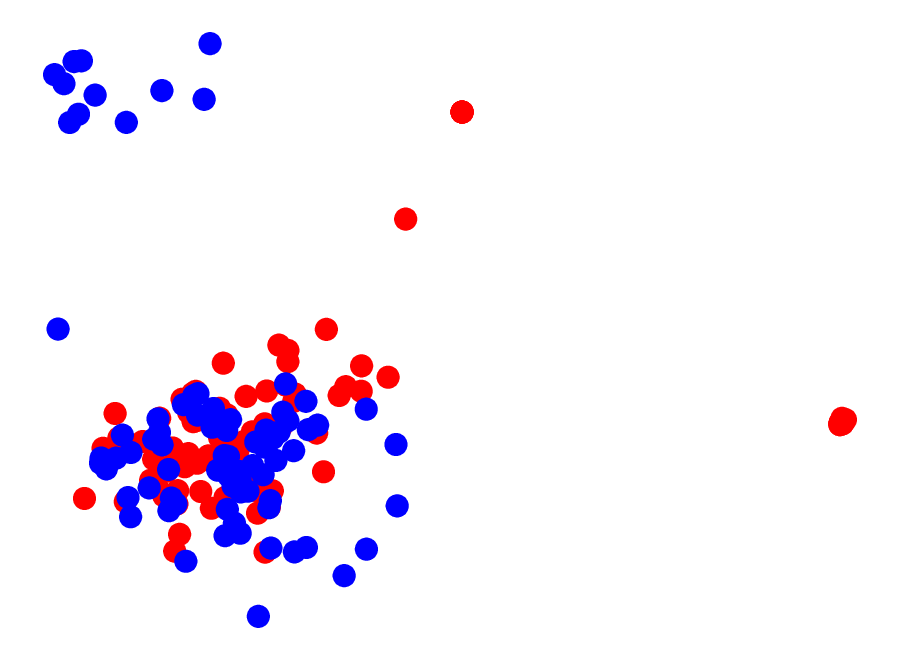}}
            \label{fig:they-them-layer5}\hfill
            \subfloat[Layer=6]{
			\centering
	\includegraphics[scale=0.24]{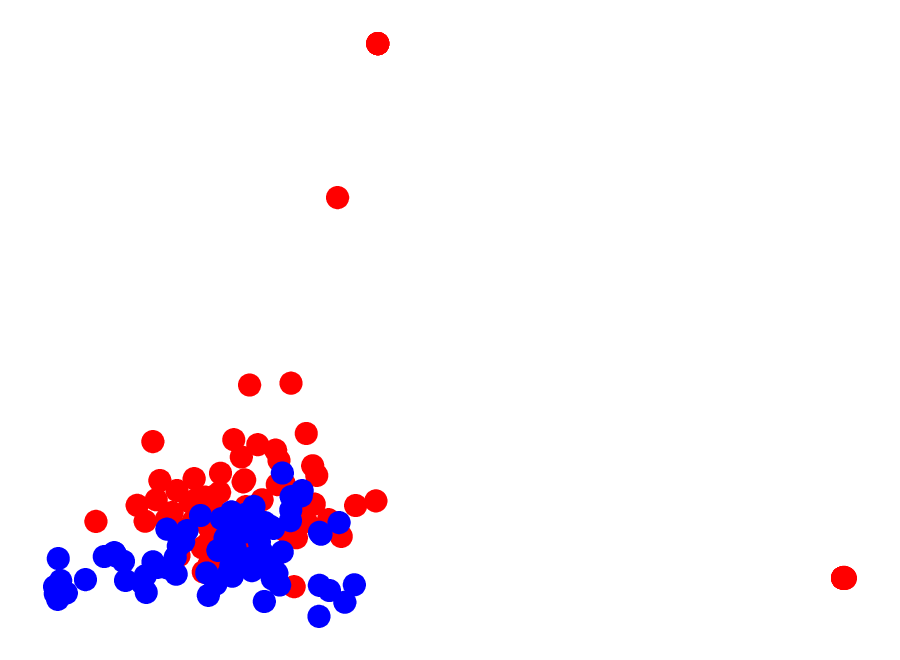}}
            \label{fig:they-them-layer6}\hfill
        \subfloat[Layer=7]{
			\centering
	\includegraphics[scale=0.24]{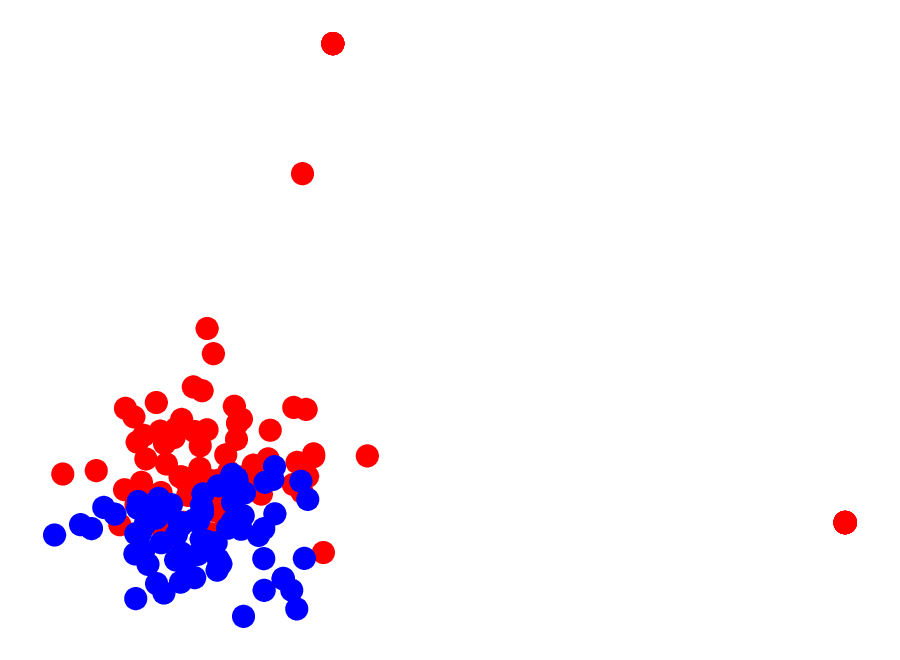}}
            \label{fig:they-them-layer7}\hfill   
        \subfloat[Layer=8]{
			\centering
	\includegraphics[scale=0.24]{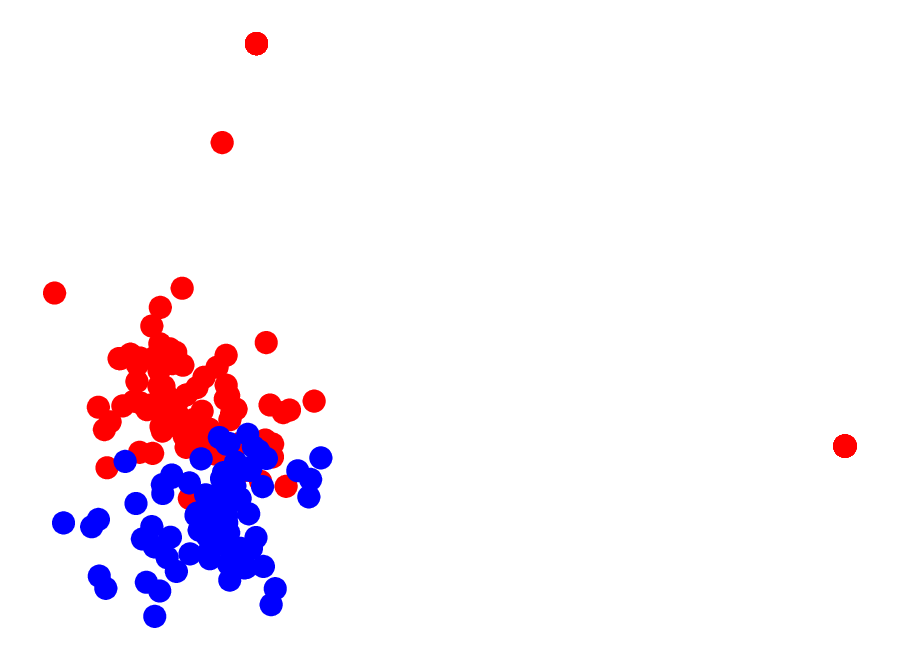}}
            \label{fig:they-them-layer8}
            
        \subfloat[Layer=9]{
			\centering
	\includegraphics[scale=0.24]{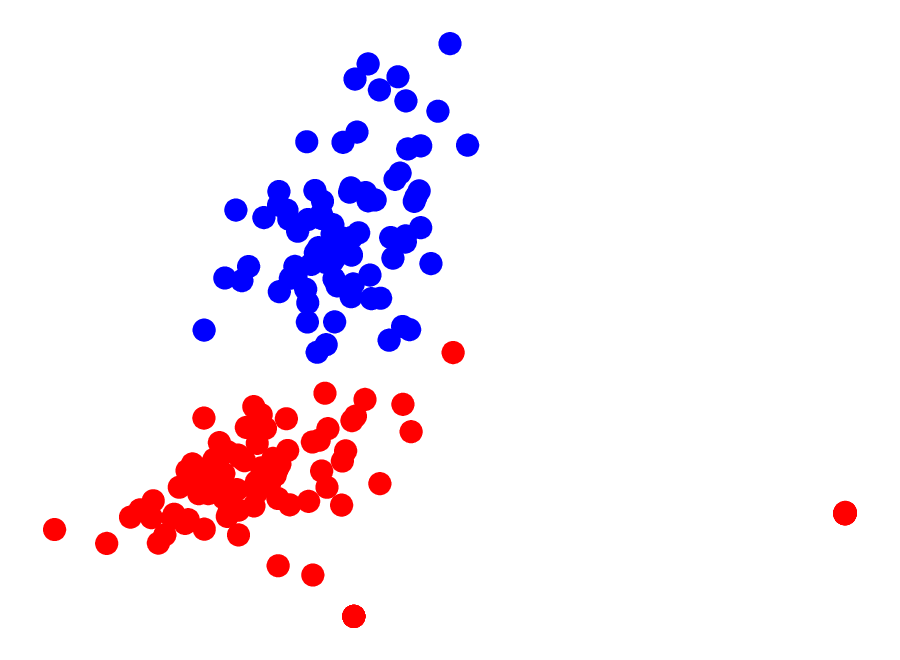}}
            \label{fig:they-them-layer9}\hfill
        \subfloat[Layer=10]{
			\centering
	\includegraphics[scale=0.24]{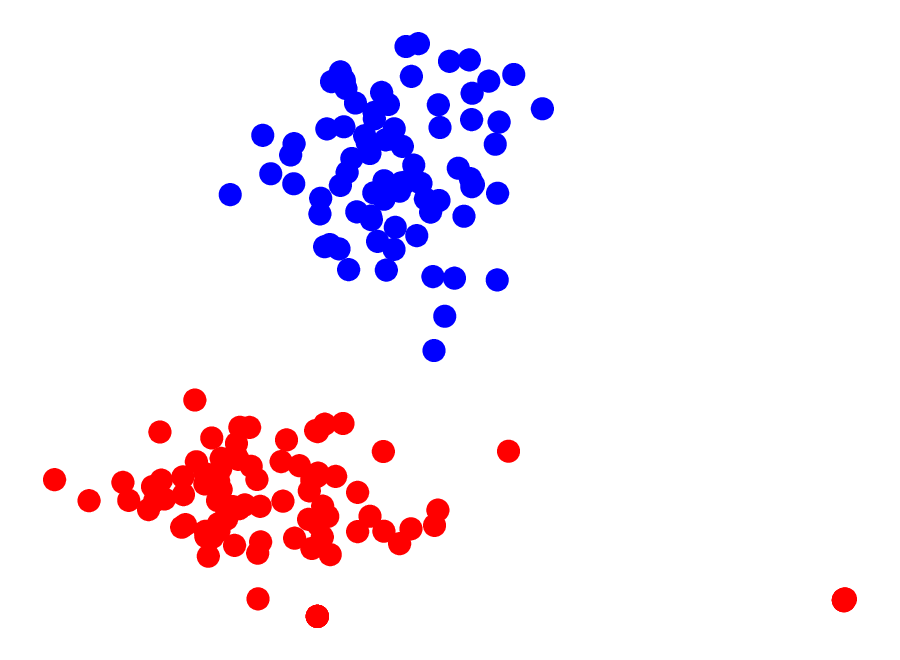}}
            \label{fig:they-them-layer10}\hfill
        \subfloat[Layer=11]{
			\centering
	\includegraphics[scale=0.24]{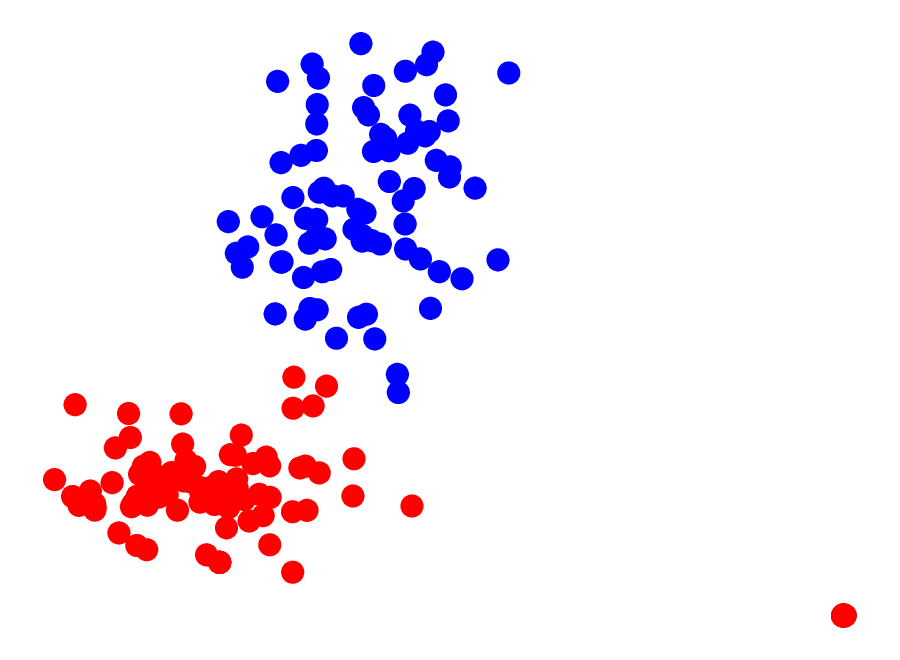}}
            \label{fig:they-them-layer11}\hfill 
        \subfloat[Layer=12]{
			\centering
	\includegraphics[scale=0.24]{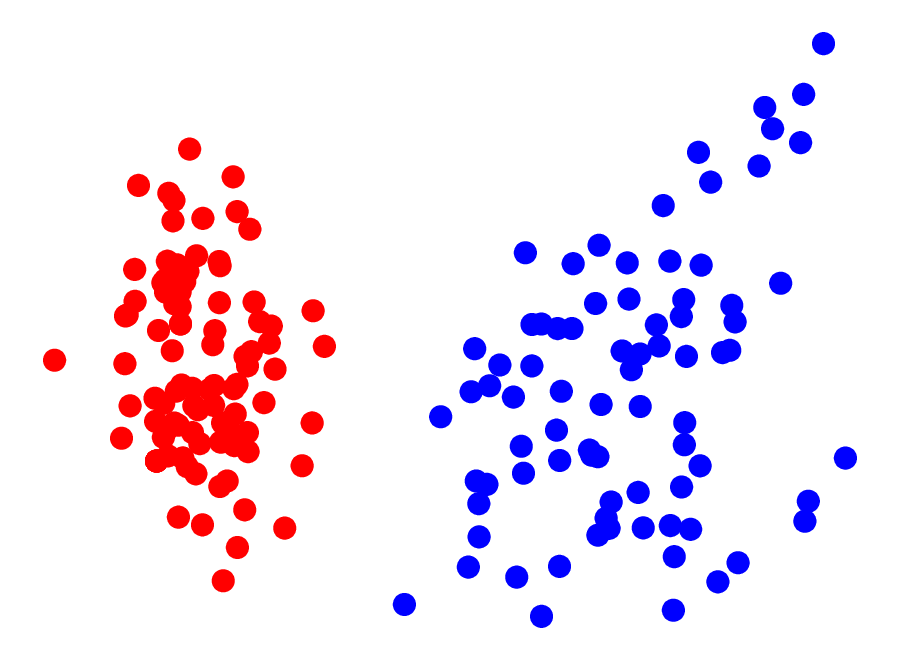}}
            \label{fig:they-them-layer12}\hfill
		\caption{Intermediate-layer contextualized token embeddings for \texttt{they</w>} (red) and \texttt{them</w>} (blue) plotted on the plane of the first two principal components. The x-axis and y-axis represent the first and second principal components, respectively.
		}
		\label{fig:pca-visualization-they-them}
\end{figure*}

\begin{figure*}[!htp]
    \captionsetup[subfigure]{labelformat=empty}
		\centering
            \subfloat[Layer=1]{
			\centering
	\includegraphics[scale=0.24]{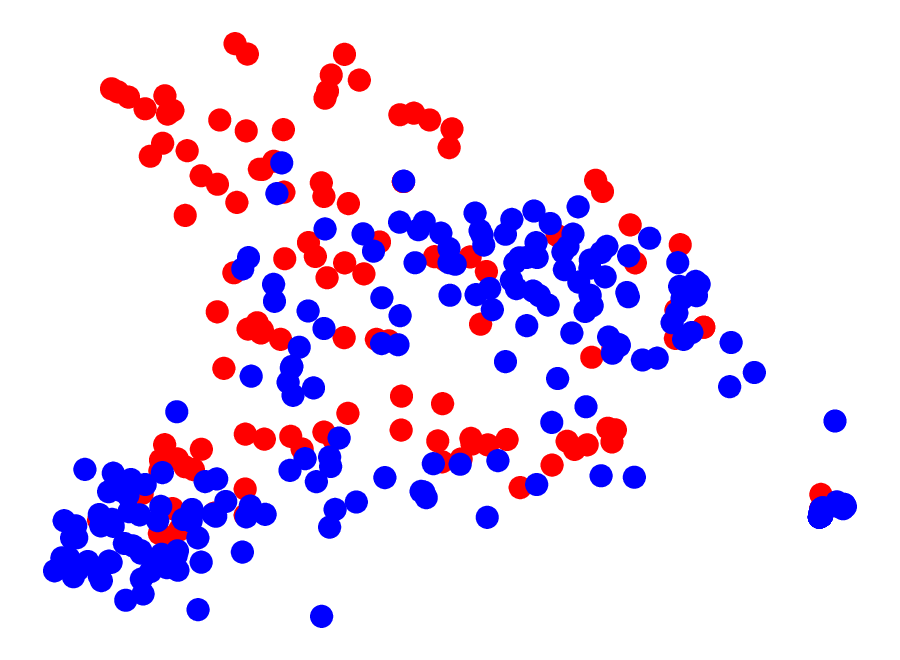}}
            \label{fig:have-had-layer1}\hfill
     	\subfloat[Layer=2]{
			\centering
	\includegraphics[scale=0.24]{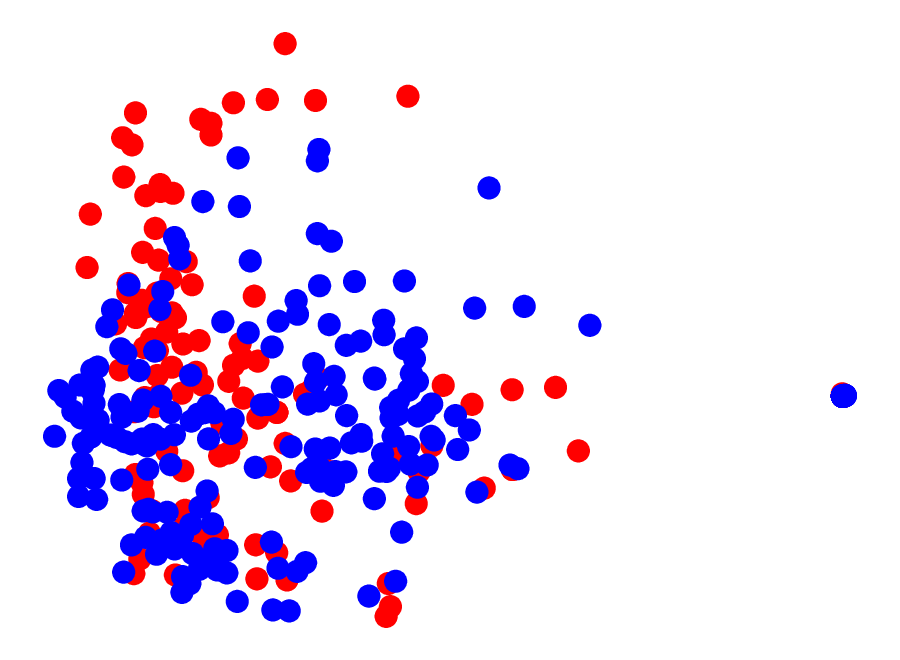}}
            \label{fig:have-had-layer2}\hfill
        \subfloat[Layer=3]{
			\centering
	\includegraphics[scale=0.24]{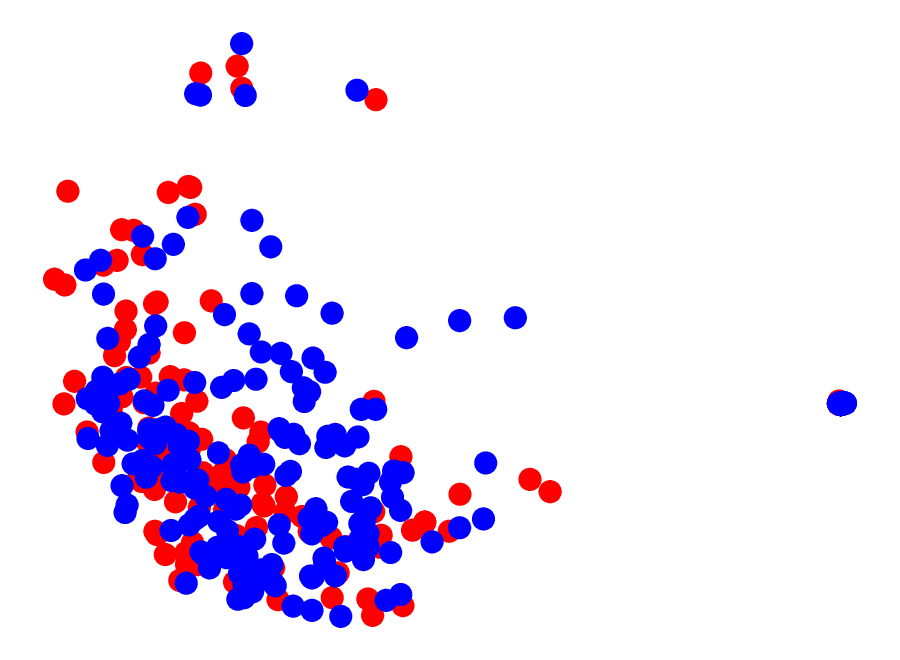}}
            \label{fig:have-had-layer3}\hfill
            \subfloat[Layer=4]{
			\centering
	\includegraphics[scale=0.24]{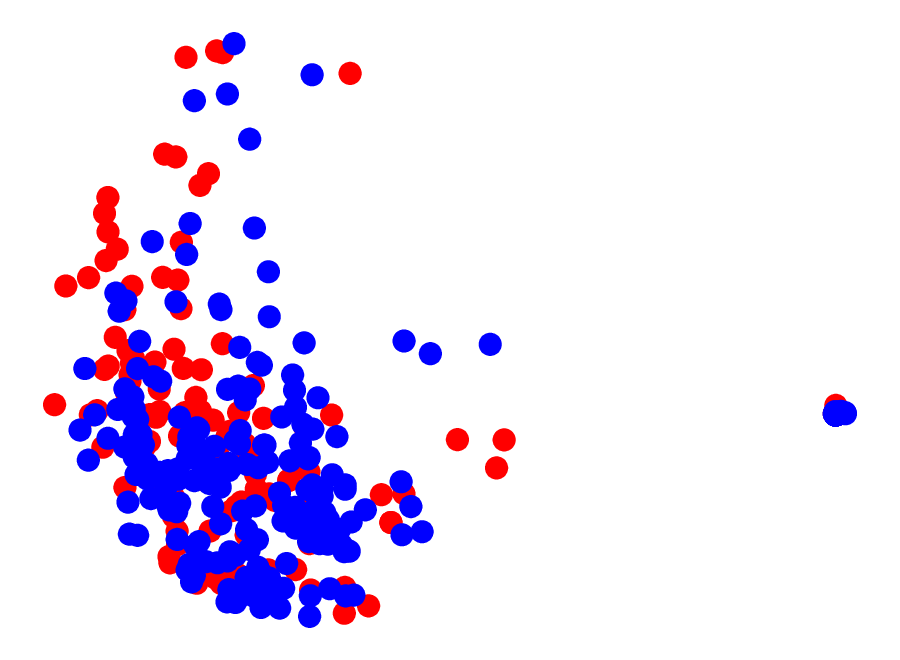}}
            \label{fig:have-had-layer4}
            
     	\subfloat[Layer=5]{
			\centering
	\includegraphics[scale=0.24]{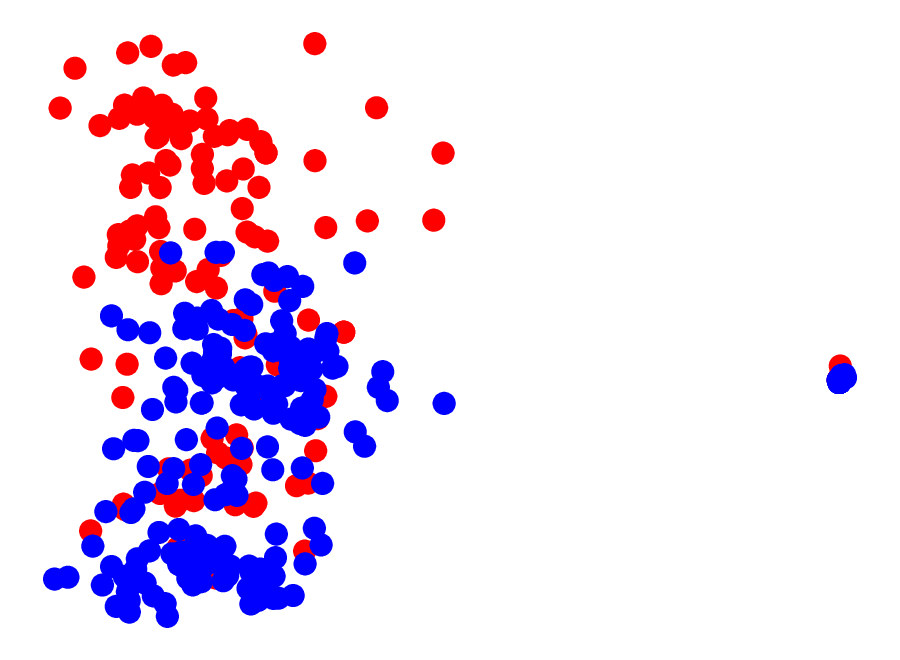}}
            \label{fig:have-had-layer5}\hfill
            \subfloat[Layer=6]{
			\centering
	\includegraphics[scale=0.24]{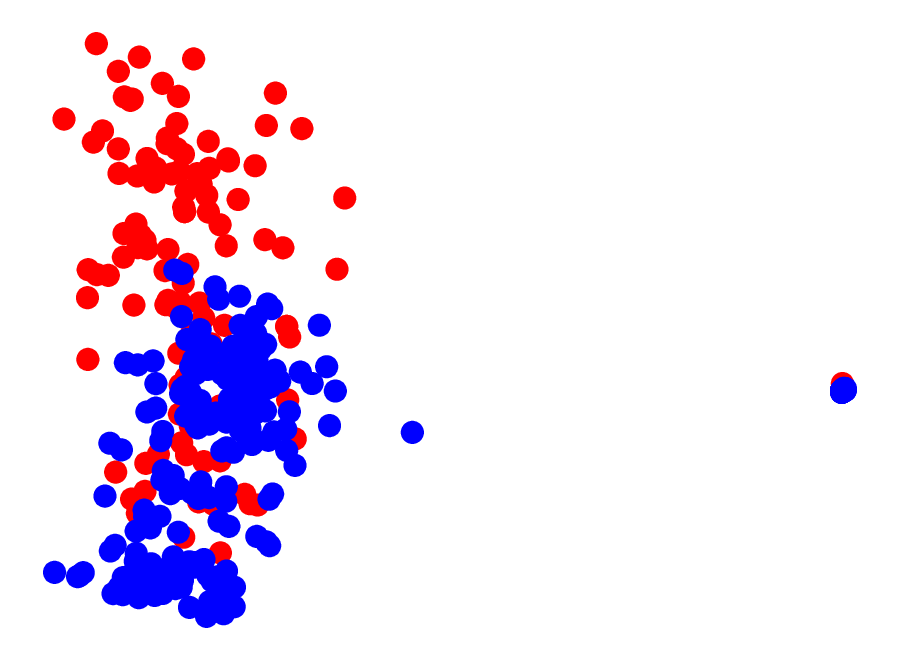}}
            \label{fig:have-had-layer6}\hfill
        \subfloat[Layer=7]{
			\centering
	\includegraphics[scale=0.24]{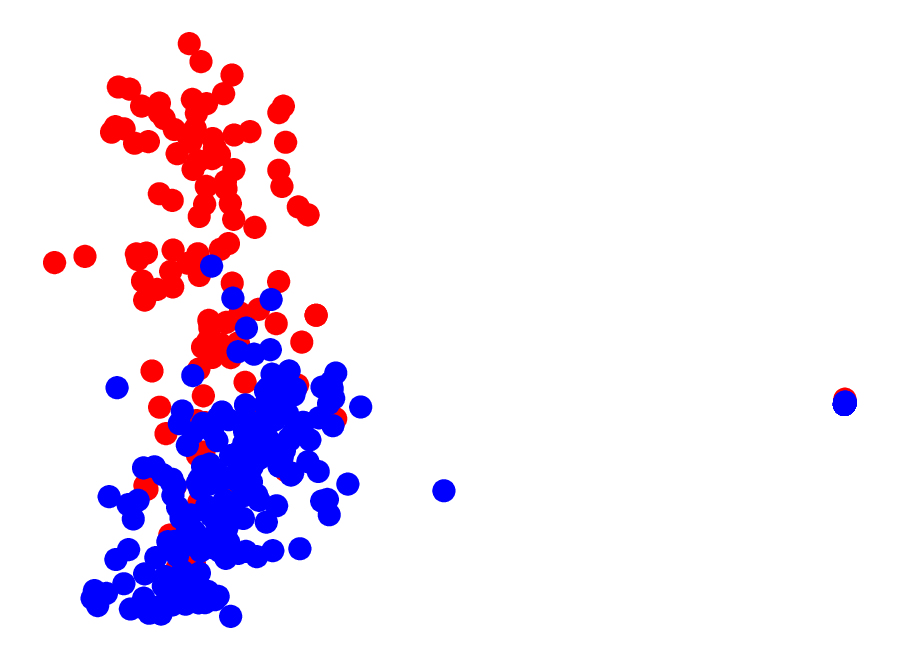}}
            \label{fig:have-had-layer7}\hfill   
        \subfloat[Layer=8]{
			\centering
	\includegraphics[scale=0.24]{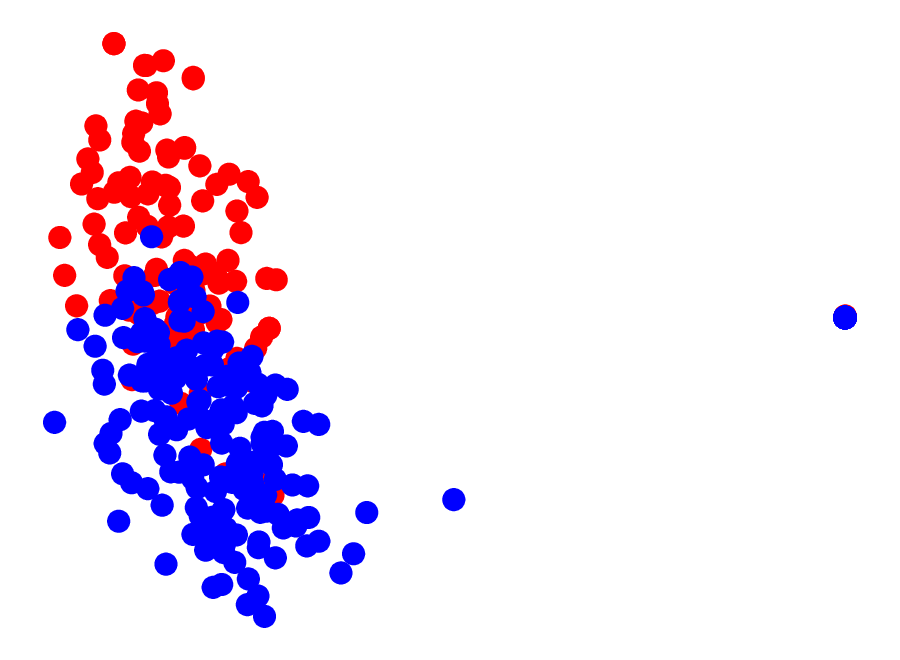}}
            \label{fig:have-had-layer8}
            
        \subfloat[Layer=9]{
			\centering
	\includegraphics[scale=0.24]{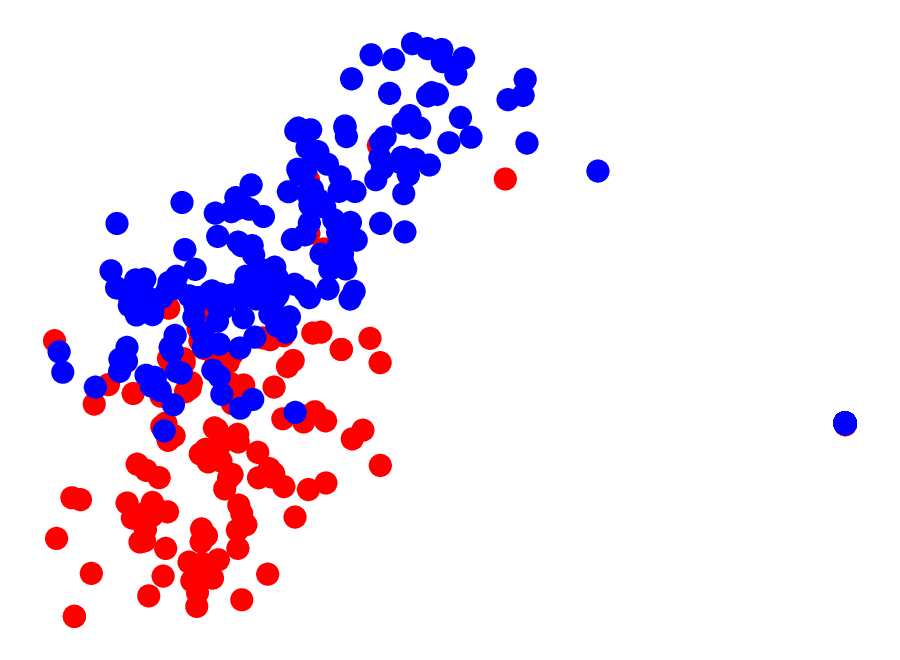}}
            \label{fig:have-had-layer9}\hfill
        \subfloat[Layer=10]{
			\centering
	\includegraphics[scale=0.24]{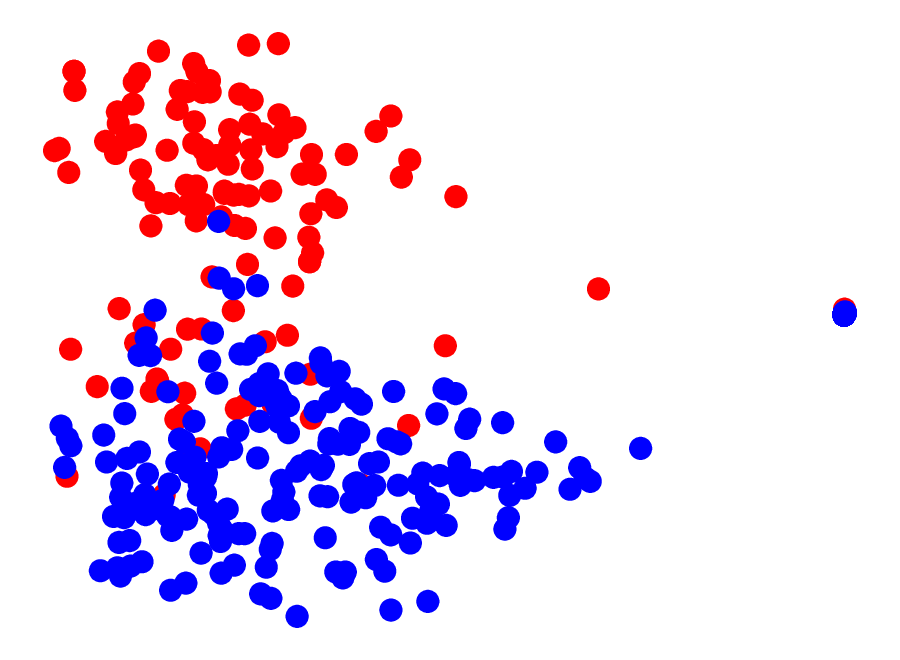}}
            \label{fig:have-had-layer10}\hfill
        \subfloat[Layer=11]{
			\centering
	\includegraphics[scale=0.24]{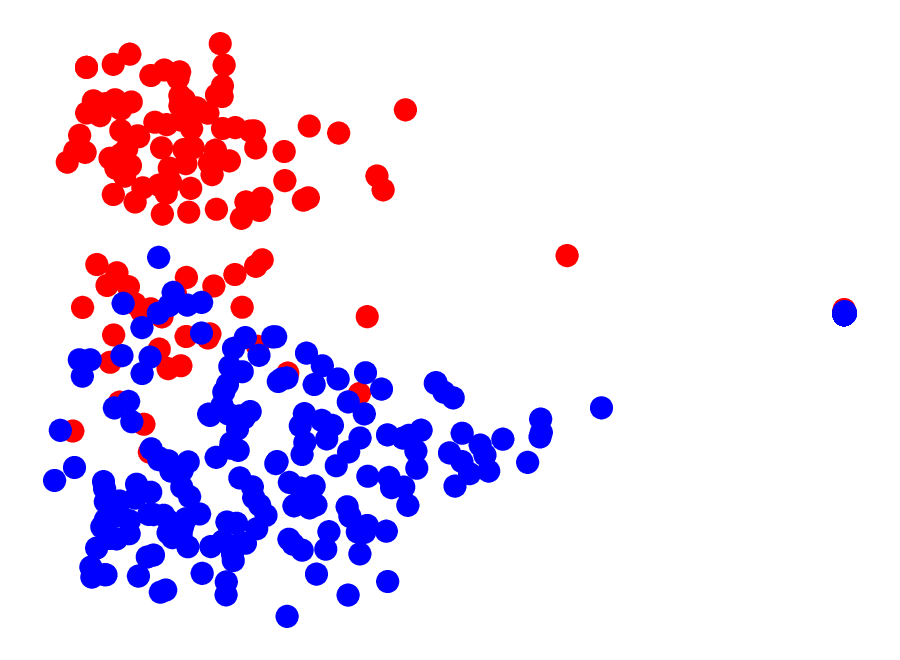}}
            \label{fig:have-had-layer11}\hfill 
        \subfloat[Layer=12]{
			\centering
	\includegraphics[scale=0.24]{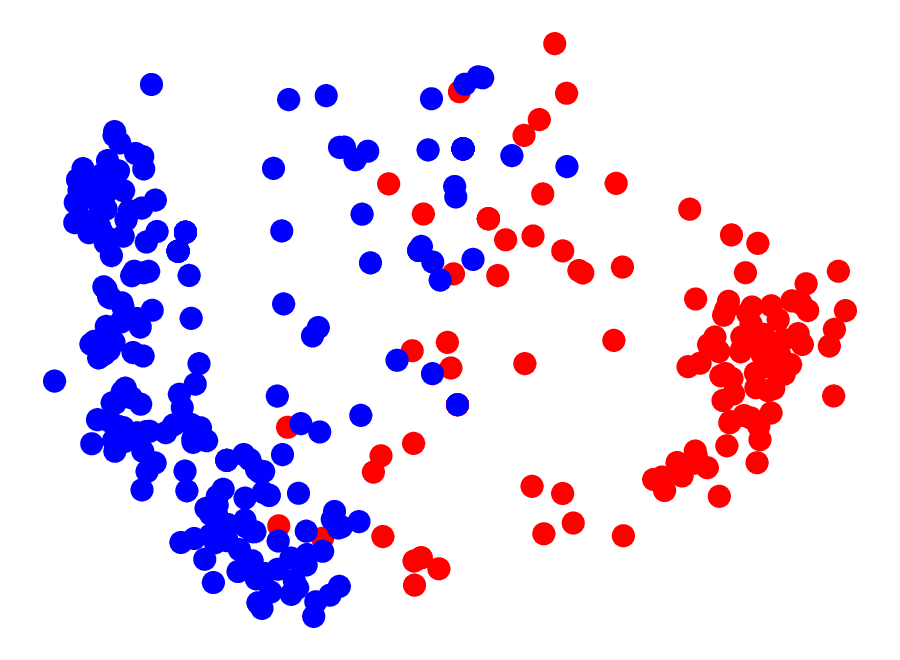}}
            \label{fig:have-had-layer12}\hfill
		\caption{Intermediate-layer contextualized token embeddings for \texttt{have</w>} (red) and \texttt{had</w>} (blue) plotted on the plane of the first two principal components. The x-axis and y-axis represent the first and second principal components, respectively.
		}
		\label{fig:pca-visualization-have-had}
\end{figure*}

\begin{figure*}[!htp]
    \captionsetup[subfigure]{labelformat=empty}
		\centering
            \subfloat[Layer=1]{
			\centering
	\includegraphics[scale=0.24]{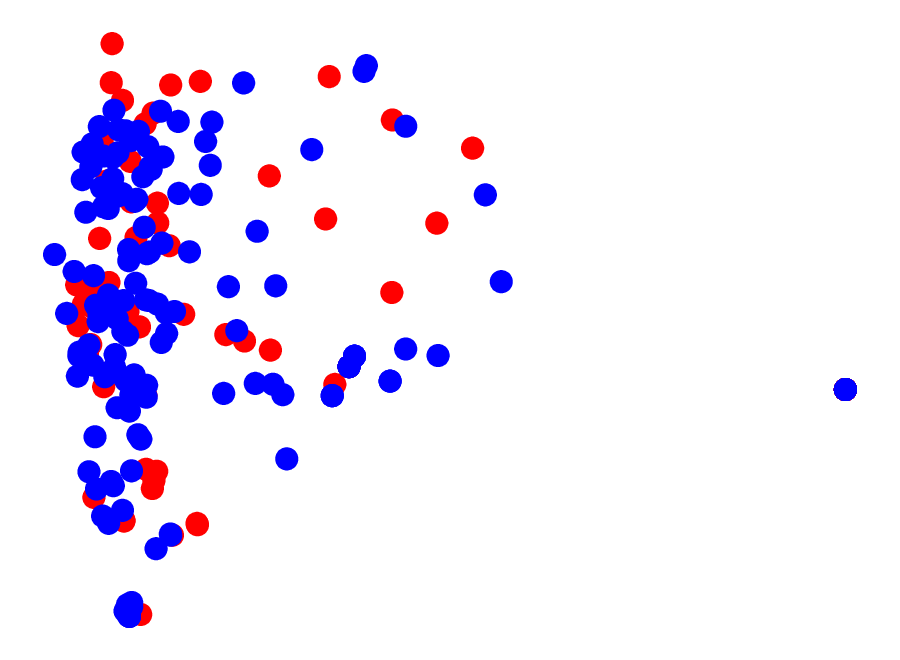}}
            \label{fig:are-is-layer1}\hfill
     	\subfloat[Layer=2]{
			\centering
	\includegraphics[scale=0.24]{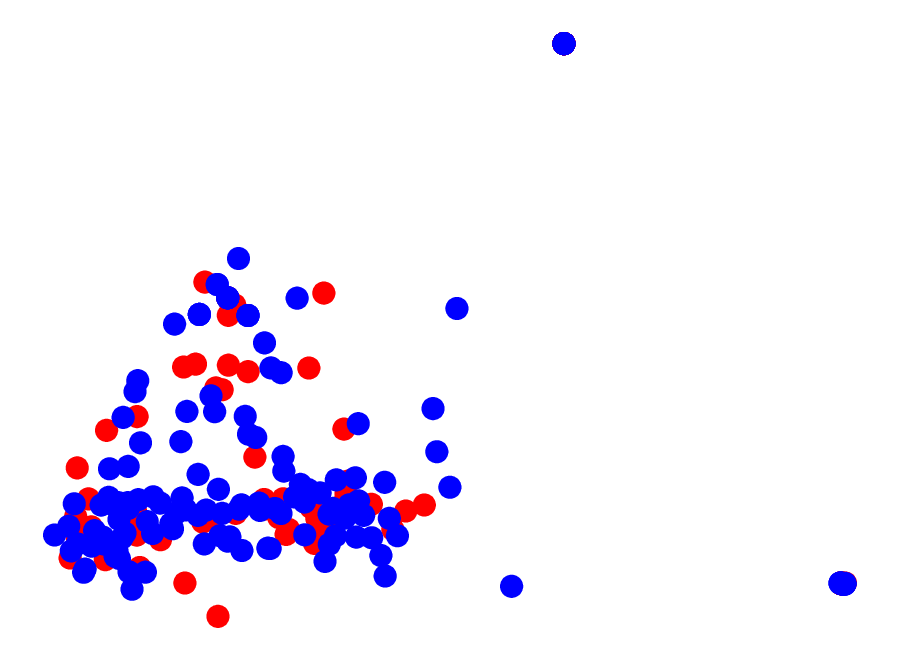}}
            \label{fig:are-is-layer2}\hfill
        \subfloat[Layer=3]{
			\centering
	\includegraphics[scale=0.24]{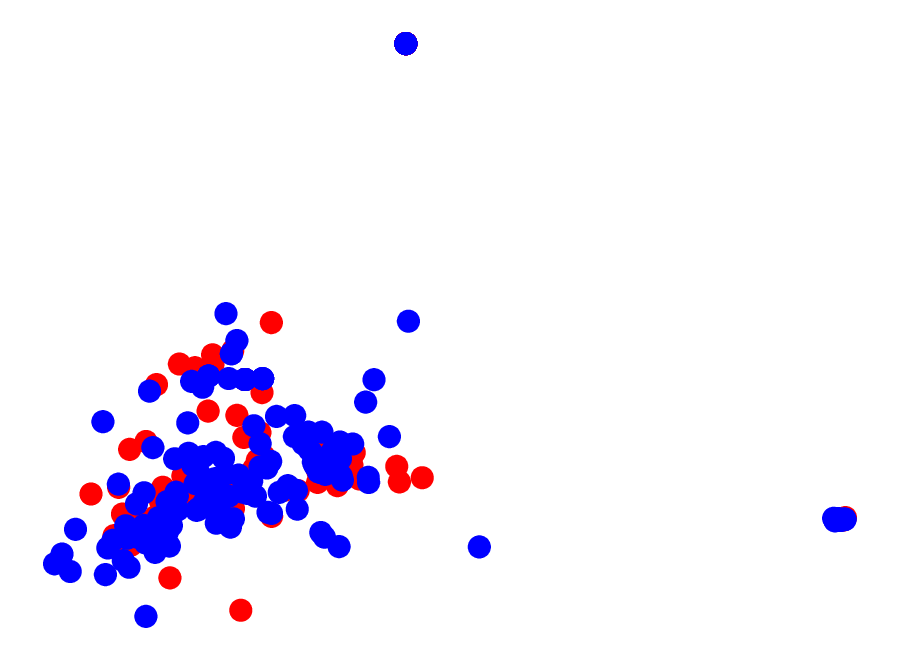}}
            \label{fig:are-is-layer3}\hfill
            \subfloat[Layer=4]{
			\centering
	\includegraphics[scale=0.24]{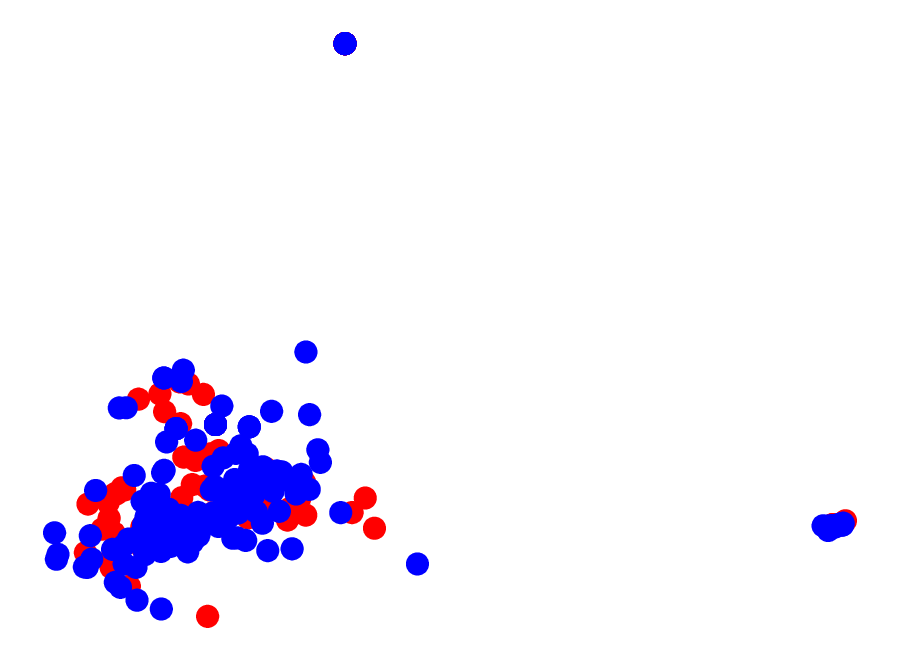}}
            \label{fig:are-is-layer4}
            
     	\subfloat[Layer=5]{
			\centering
	\includegraphics[scale=0.24]{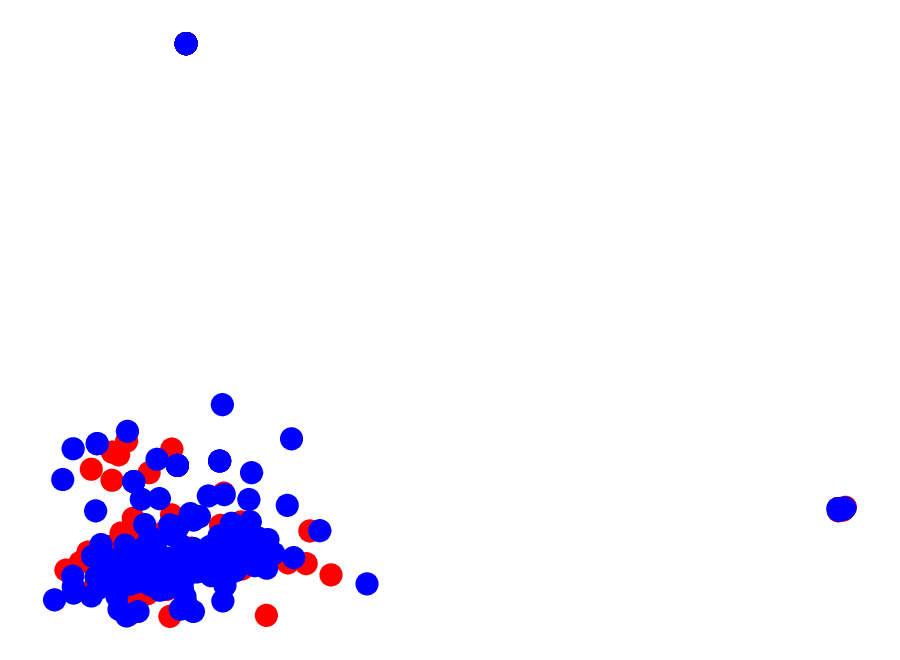}}
            \label{fig:are-is-layer5}\hfill
            \subfloat[Layer=6]{
			\centering
	\includegraphics[scale=0.24]{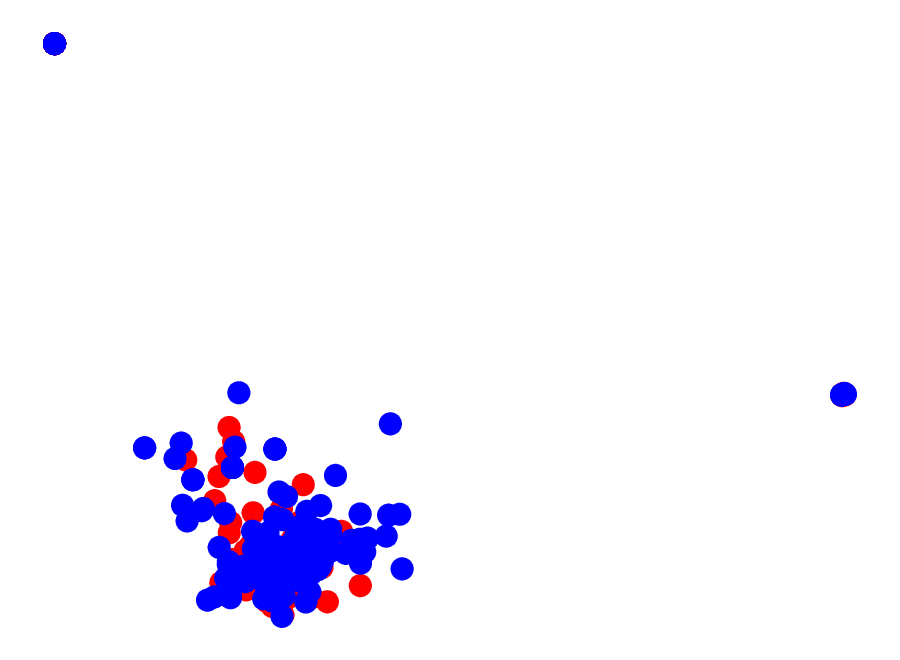}}
            \label{fig:are-is-layer6}\hfill
        \subfloat[Layer=7]{
			\centering
	\includegraphics[scale=0.24]{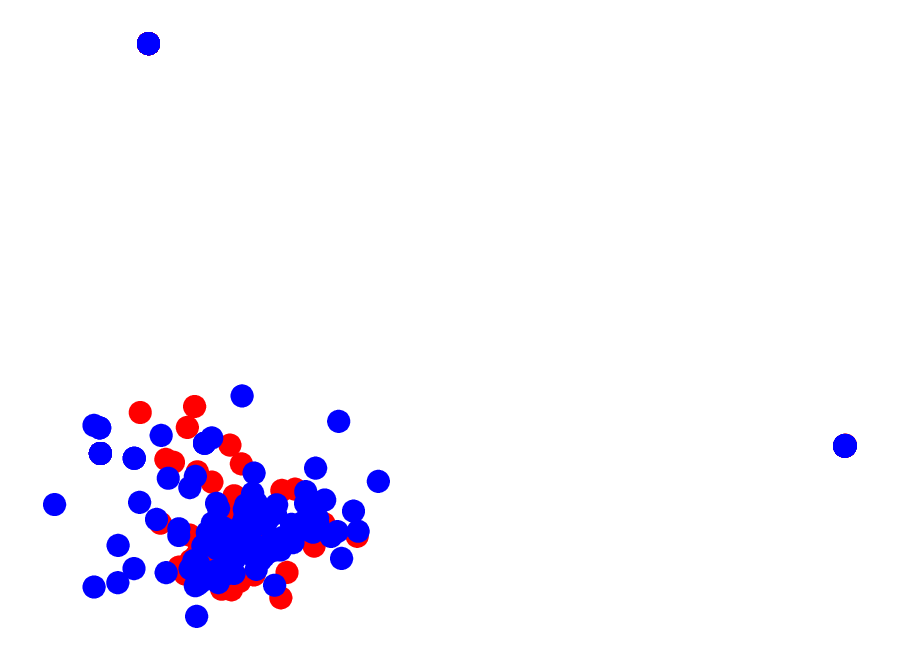}}
            \label{fig:are-is-layer7}\hfill   
        \subfloat[Layer=8]{
			\centering
	\includegraphics[scale=0.24]{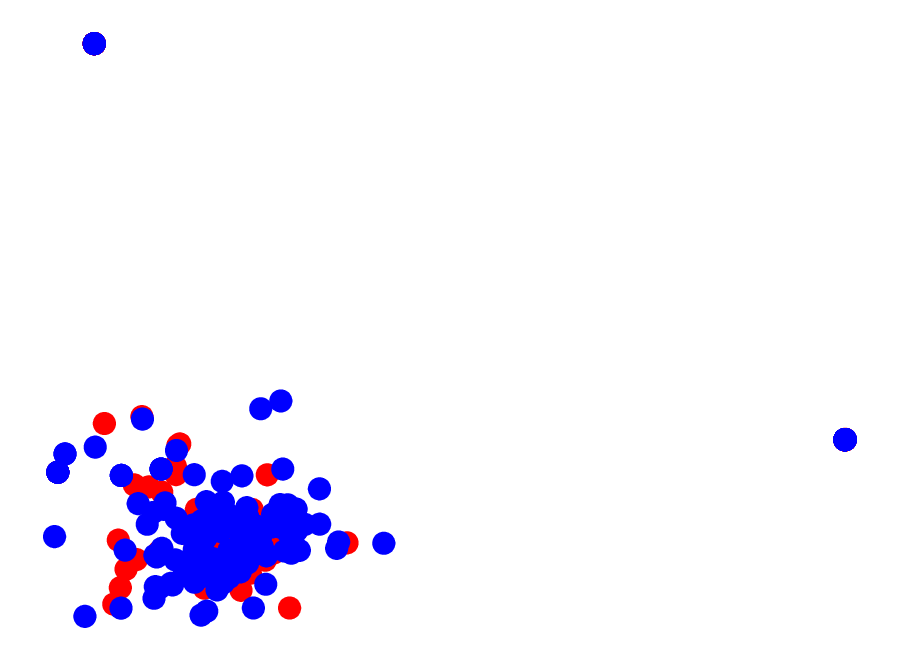}}
            \label{fig:are-is-layer8}
            
        \subfloat[Layer=9]{
			\centering
	\includegraphics[scale=0.24]{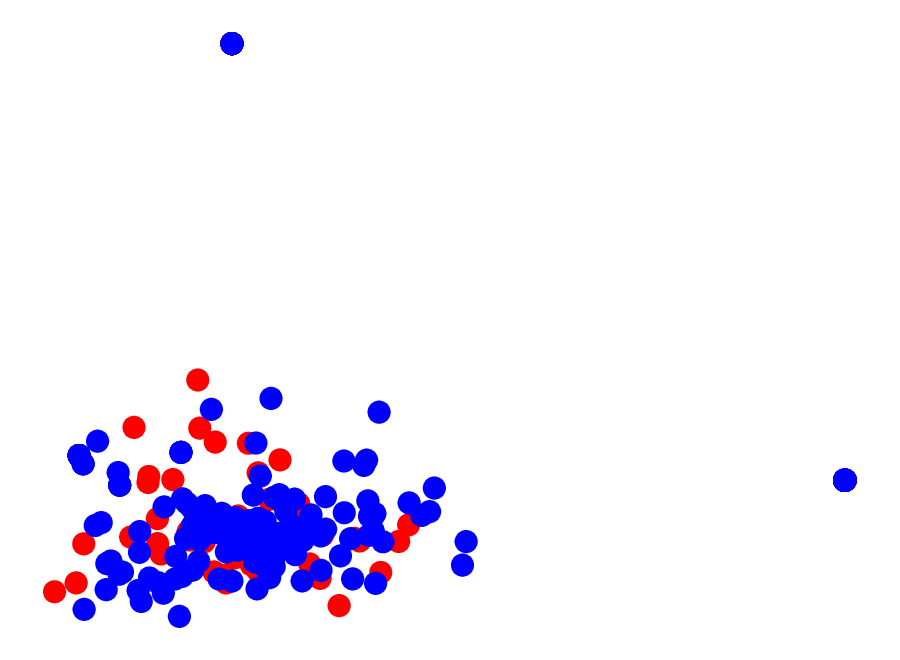}}
            \label{fig:are-is-layer9}\hfill
        \subfloat[Layer=10]{
			\centering
	\includegraphics[scale=0.24]{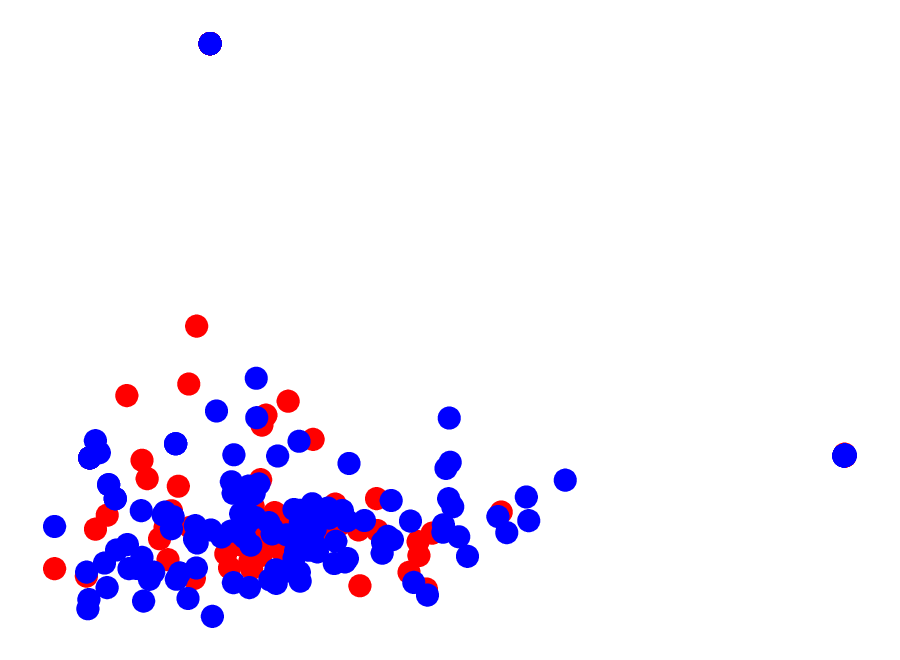}}
            \label{fig:are-is-layer10}\hfill
        \subfloat[Layer=11]{
			\centering
	\includegraphics[scale=0.24]{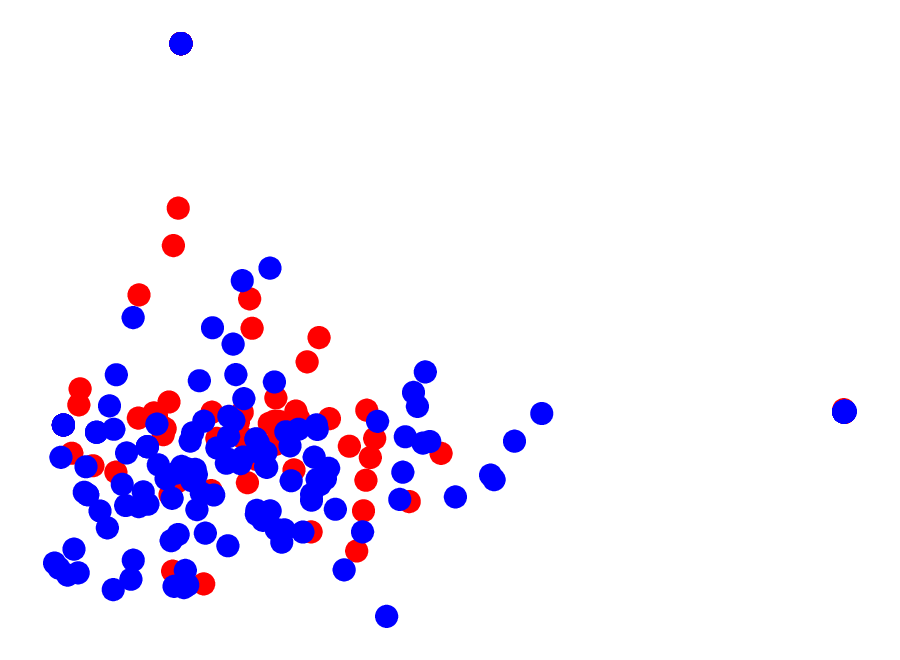}}
            \label{fig:are-is-layer11}\hfill 
        \subfloat[Layer=12]{
			\centering
	\includegraphics[scale=0.24]{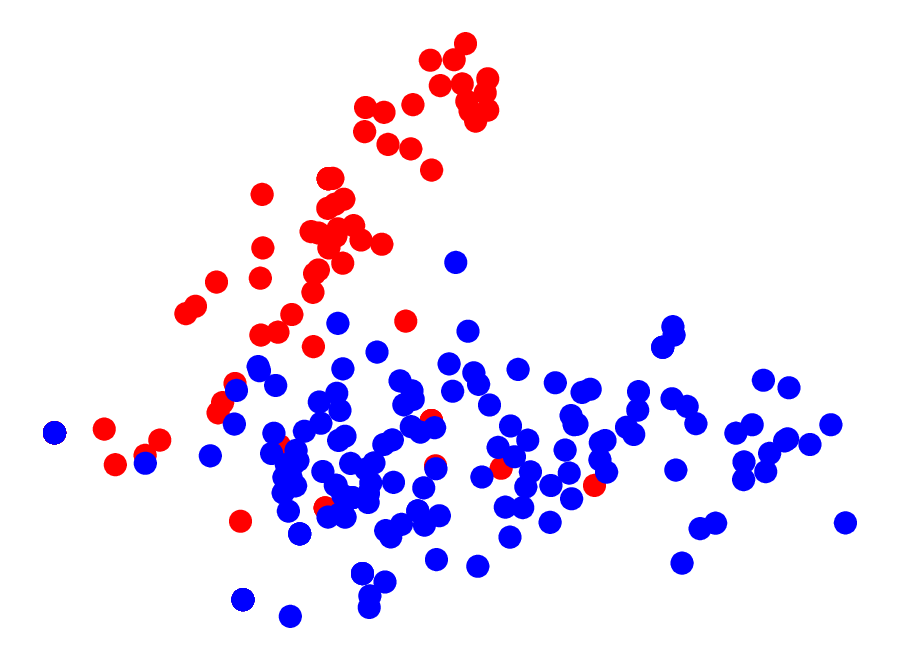}}
            \label{fig:are-is-layer12}\hfill
		\caption{Intermediate-layer contextualized token embeddings for \texttt{are</w>} (red) and \texttt{is</w>} (blue) plotted on the plane of the first two principal components. The x-axis and y-axis represent the first and second principal components, respectively.
		}
		\label{fig:pca-visualization-are-is}
\end{figure*}

\begin{figure*}[!htp]
    \captionsetup[subfigure]{labelformat=empty}
		\centering
            \subfloat[MedRAG-Textbooks]{
			\centering
		\includegraphics[scale=0.32]{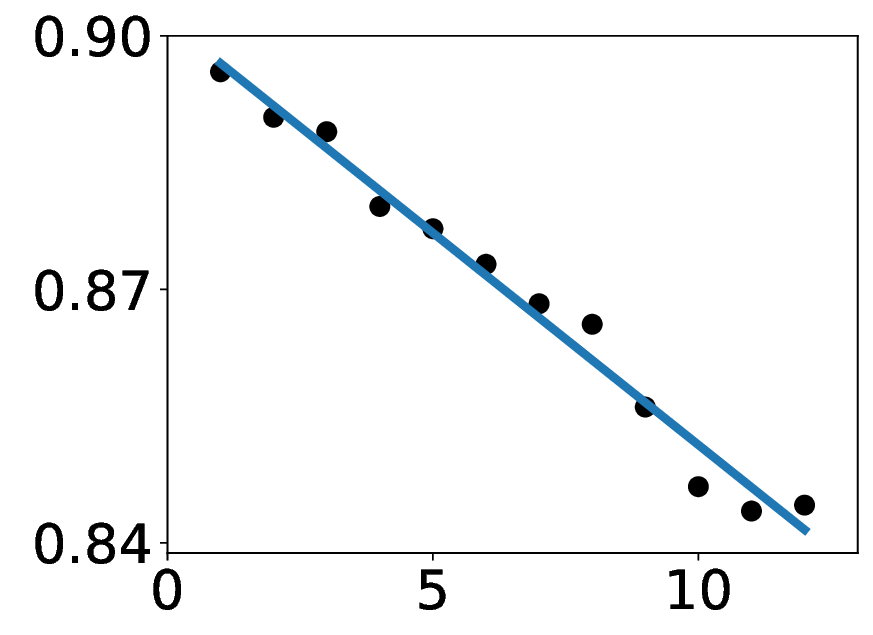}}
            \label{fig:MedRAG-textbooks}\hfill
            \subfloat[LegalBench]{
			\centering
		\includegraphics[scale=0.32]{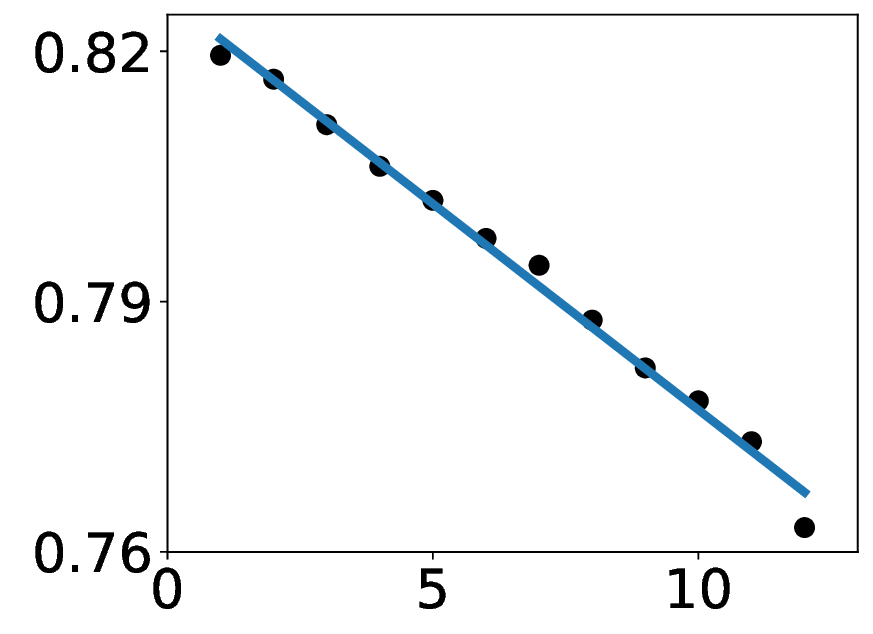}}
            \label{fig:legalbencht}\hfill
     	\subfloat[US-Congressional-Speeches]{
			\centering
		\includegraphics[scale=0.32]{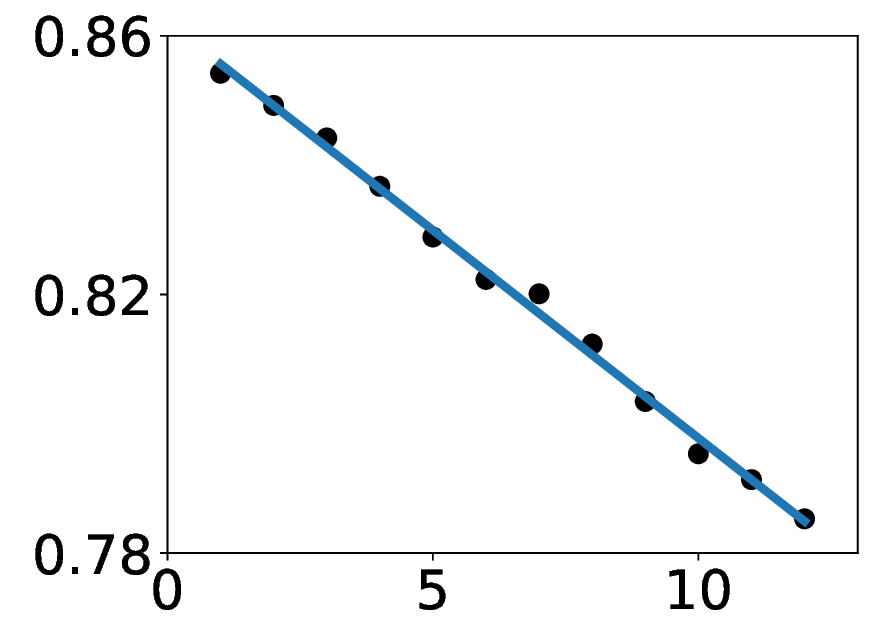}}
            \label{fig:us-congressional-speeches}\hfill

		\caption{The law of equi-learning emerges when GPT-1 is evaluated across different domains, including medicine (MedRAG-Textbooks), law (LegalBench), and politics (US-Congressional-Speeches), with corresponding Pearson correlation coefficients of $-0.990$, $-0.995$, and $-0.998$, respectively. The x-axis denotes the layer index, while the y-axis (log scale) shows the prediction residual (PR) as defined in Eq.~\ref{eq:PR}.
		}
		\label{fig:science-analysis}
\end{figure*}

\begin{figure*}[!htp]
    \captionsetup[subfigure]{labelformat=empty}
		\centering
            \subfloat[Layer=1]{
			\centering
	\includegraphics[scale=0.24]{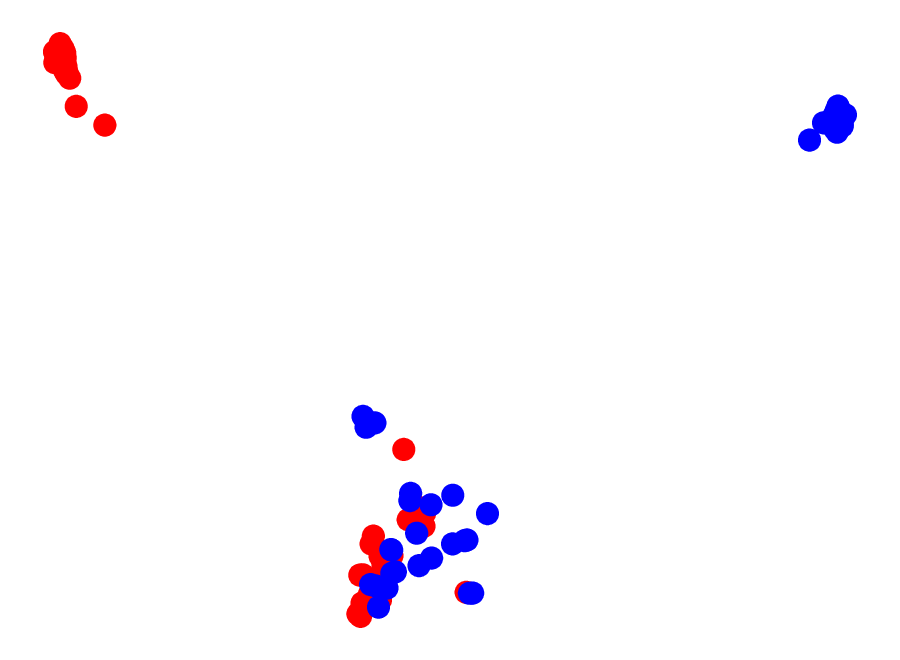}}
            \label{fig:law-layer1}\hfill
     	\subfloat[Layer=2]{
			\centering
	\includegraphics[scale=0.24]{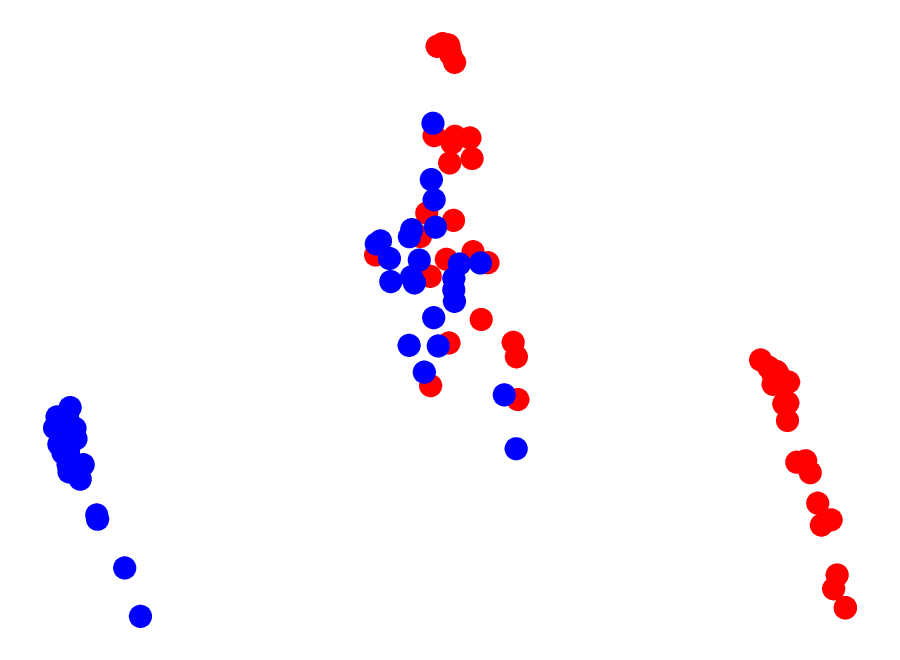}}
            \label{fig:law-layer2}\hfill
        \subfloat[Layer=3]{
			\centering
	\includegraphics[scale=0.24]{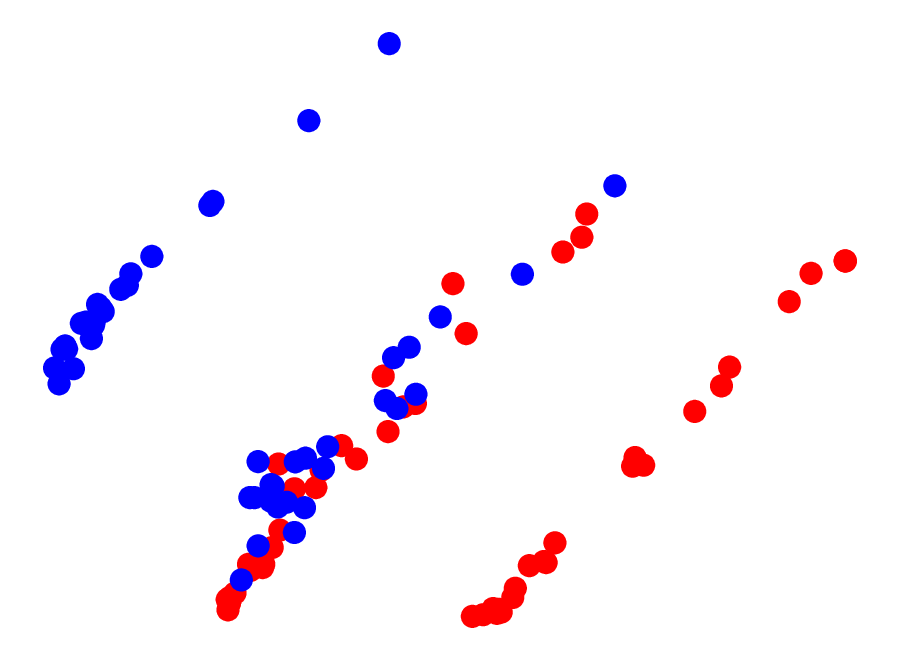}}
            \label{fig:law-layer3}\hfill
            \subfloat[Layer=4]{
			\centering
	\includegraphics[scale=0.24]{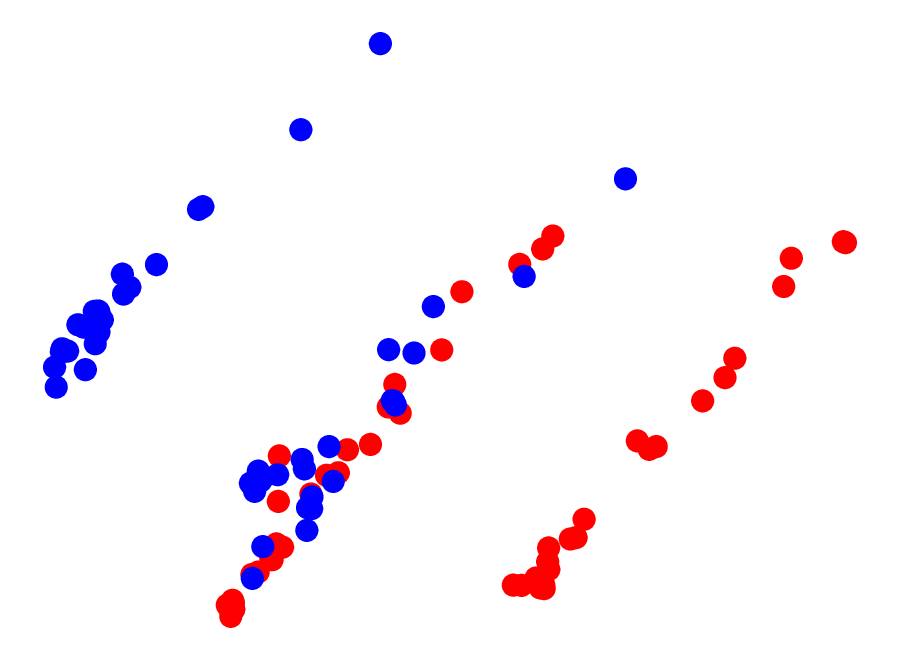}}
            \label{fig:law-layer4}
            
     	\subfloat[Layer=5]{
			\centering
	\includegraphics[scale=0.24]{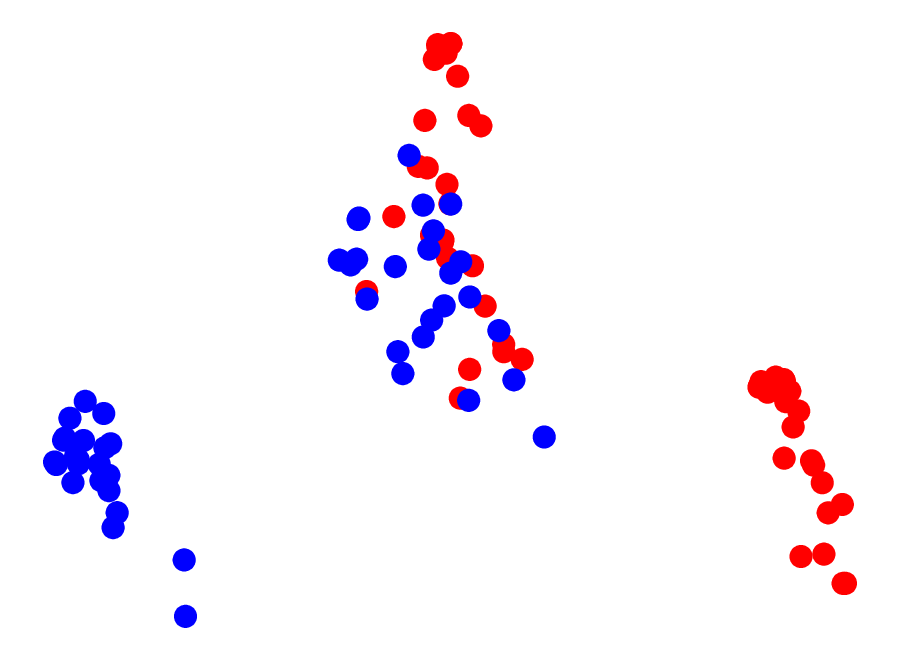}}
            \label{fig:law-layer5}\hfill
            \subfloat[Layer=6]{
			\centering
	\includegraphics[scale=0.24]{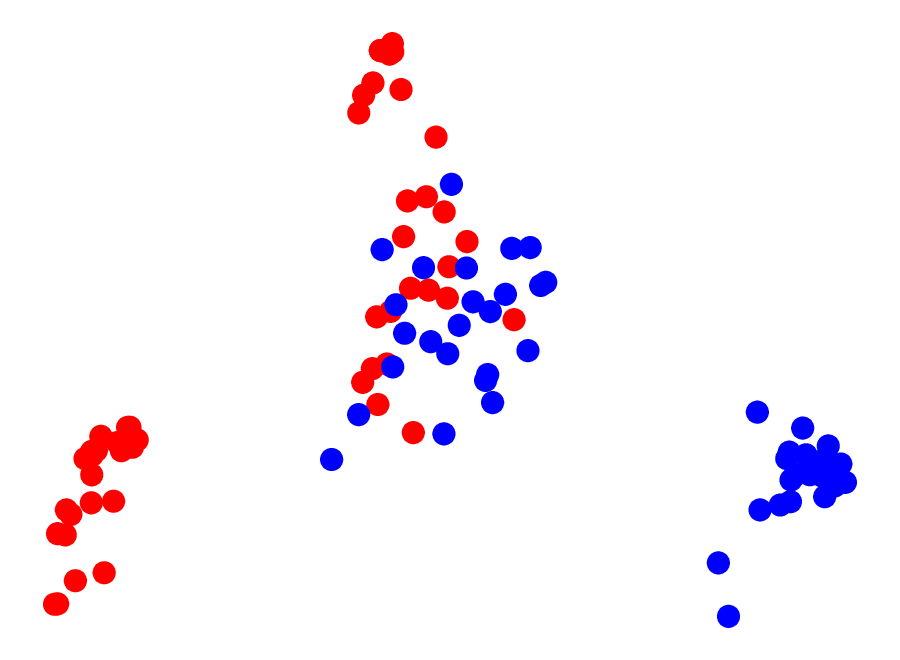}}
            \label{fig:law-layer6}\hfill
        \subfloat[Layer=7]{
			\centering
	\includegraphics[scale=0.24]{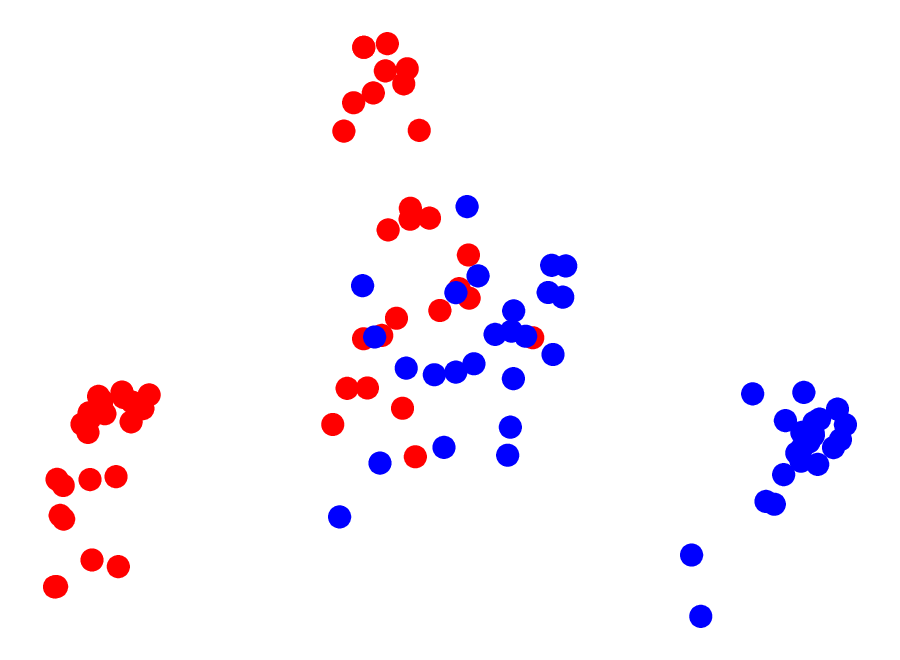}}
            \label{fig:law-layer7}\hfill   
        \subfloat[Layer=8]{
			\centering
	\includegraphics[scale=0.24]{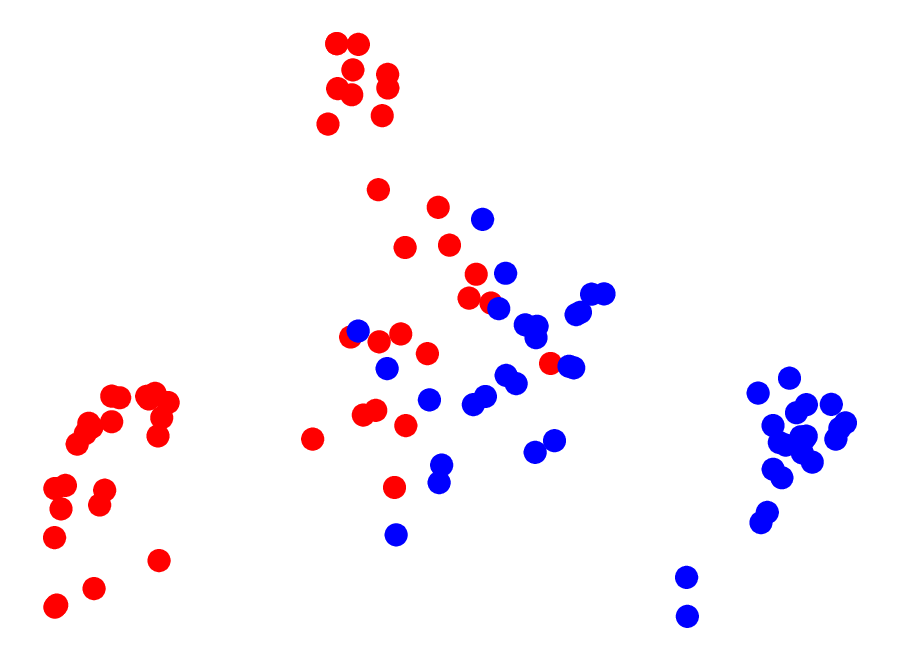}}
            \label{fig:law-layer8}
            
        \subfloat[Layer=9]{
			\centering
	\includegraphics[scale=0.24]{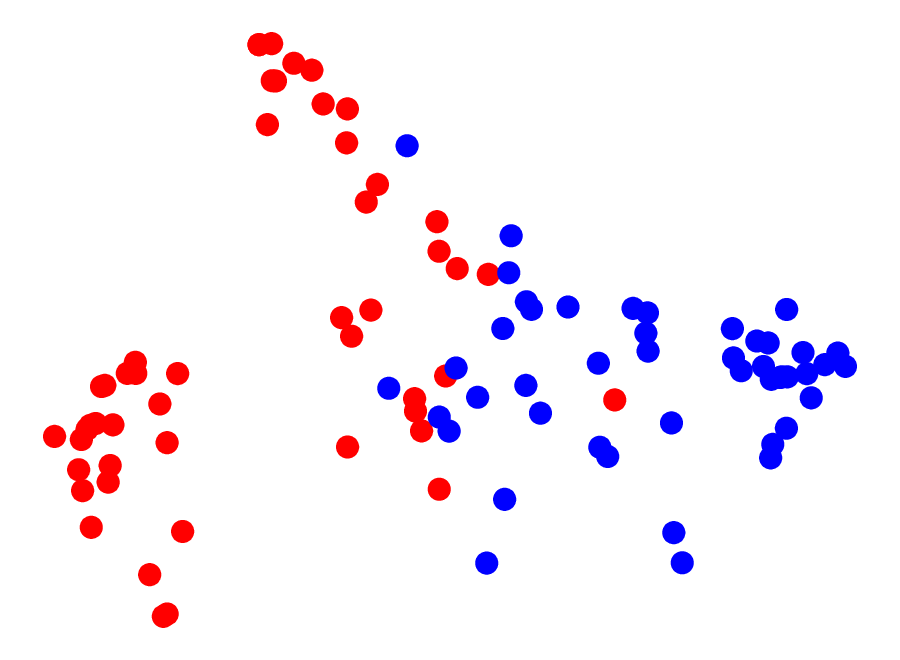}}
            \label{fig:law-layer9}\hfill
        \subfloat[Layer=10]{
			\centering
	\includegraphics[scale=0.24]{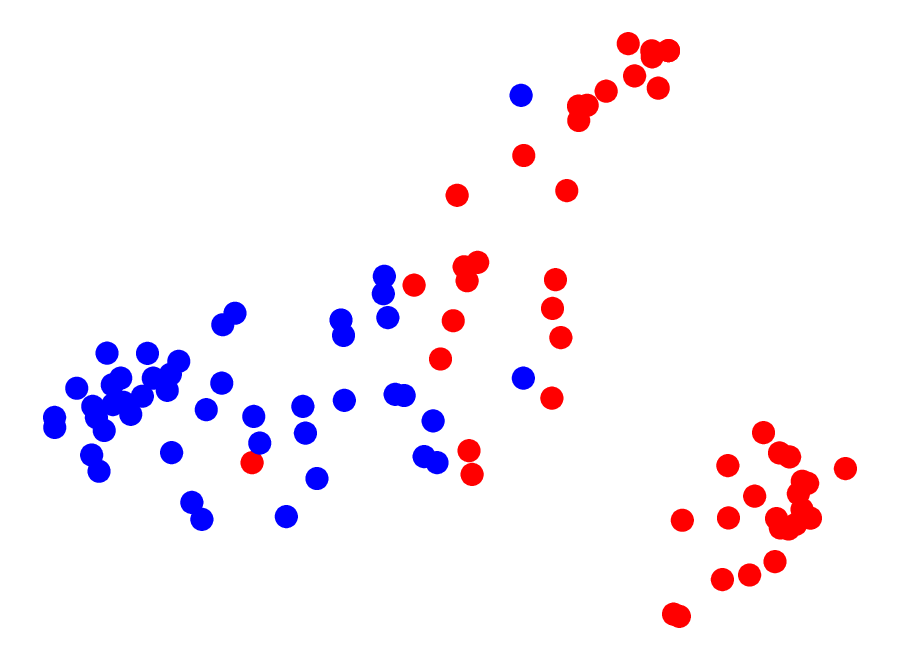}}
            \label{fig:law-layer10}\hfill
        \subfloat[Layer=11]{
			\centering
	\includegraphics[scale=0.24]{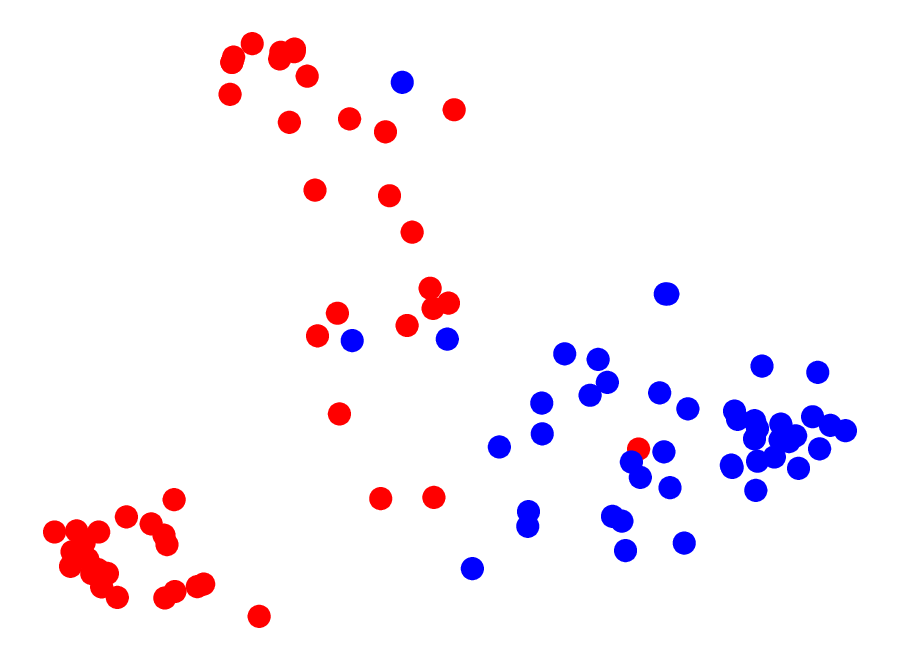}}
            \label{fig:law-layer11}\hfill 
        \subfloat[Layer=12]{
			\centering
	\includegraphics[scale=0.24]{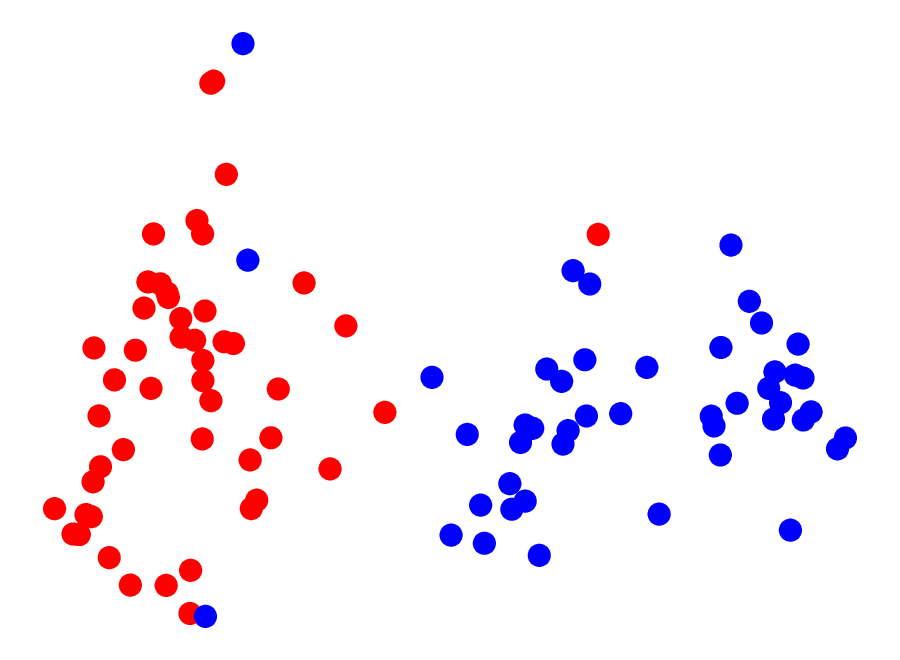}}
            \label{fig:law-layer12}\hfill
		\caption{Intermediate-layer contextualized token embeddings for \texttt{law</w>} (red) and \texttt{policy</w>} (blue) plotted on the plane of the first two principal components. The x-axis and y-axis represent the first and second principal components, respectively.
		}
		\label{fig:pca-visualization-law}
\end{figure*}

\begin{figure*}[!htp]
    \captionsetup[subfigure]{labelformat=empty}
		\centering
            \subfloat[Layer=1]{
			\centering
	\includegraphics[scale=0.24]{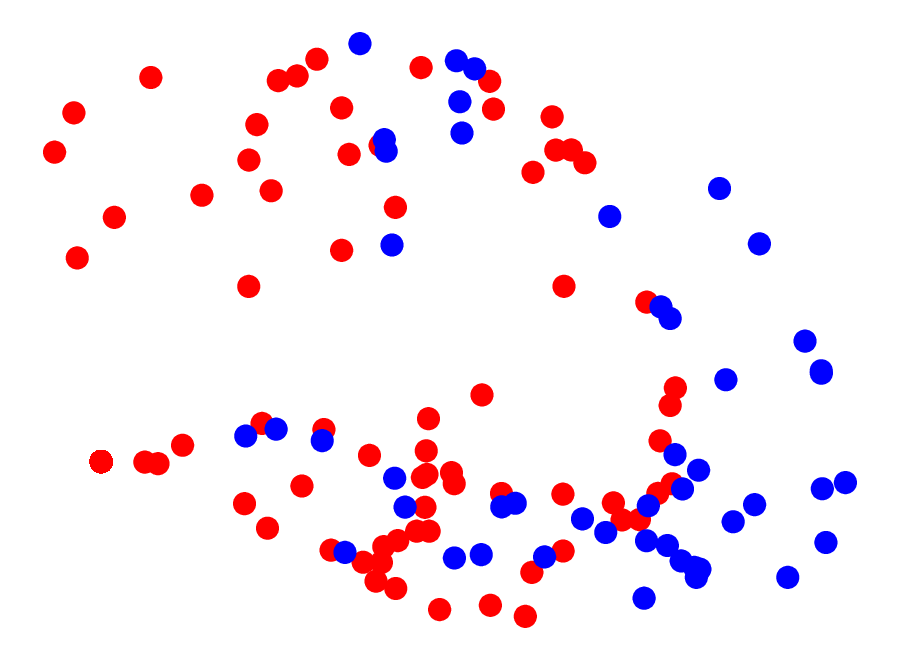}}
            \label{fig:politics-layer1}\hfill
     	\subfloat[Layer=2]{
			\centering
	\includegraphics[scale=0.24]{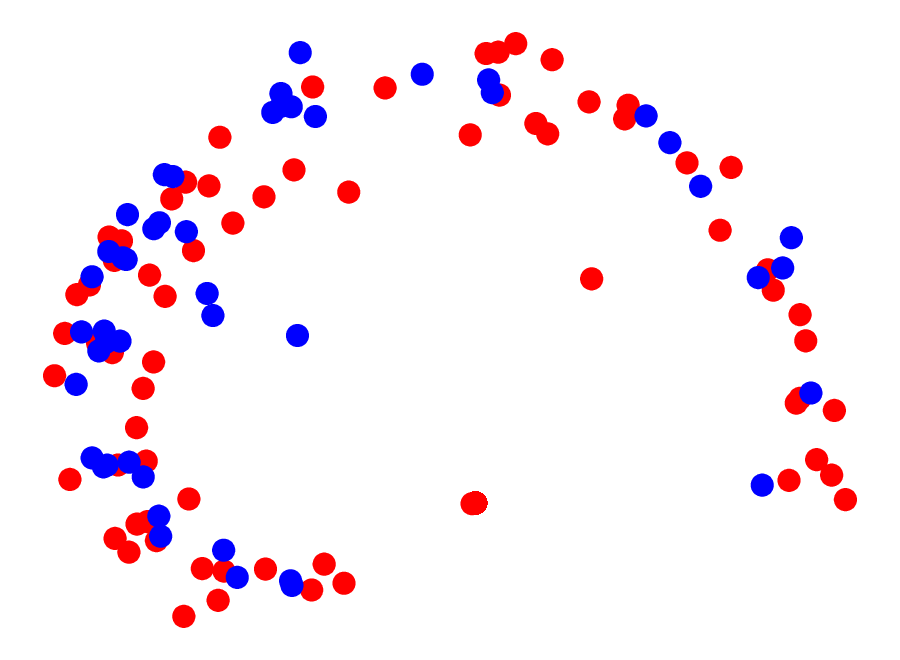}}
            \label{fig:politics-layer2}\hfill
        \subfloat[Layer=3]{
			\centering
	\includegraphics[scale=0.24]{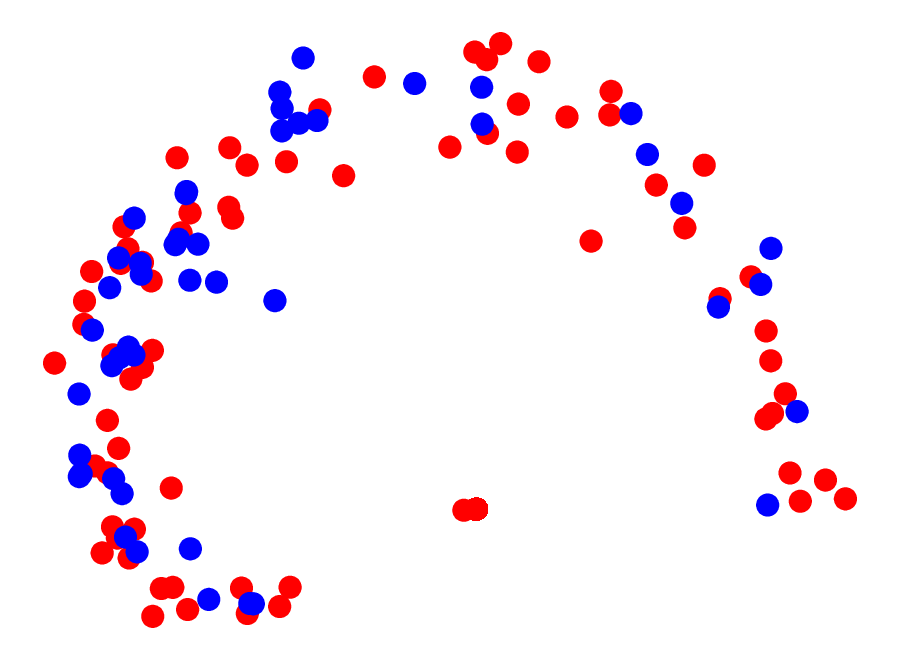}}
            \label{fig:politics-layer3}\hfill
            \subfloat[Layer=4]{
			\centering
	\includegraphics[scale=0.24]{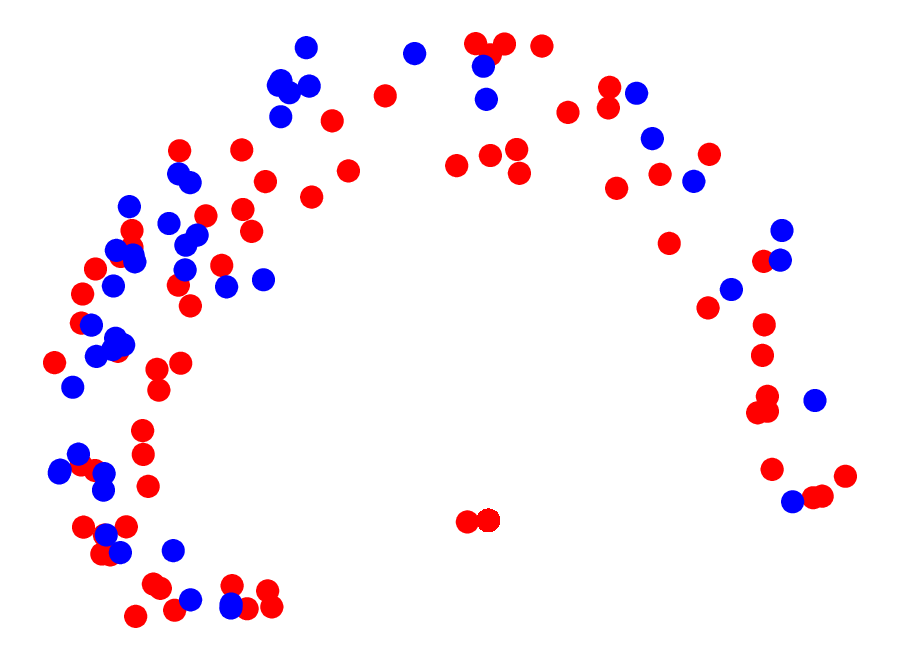}}
            \label{fig:politics-layer4}
            
     	\subfloat[Layer=5]{
			\centering
	\includegraphics[scale=0.24]{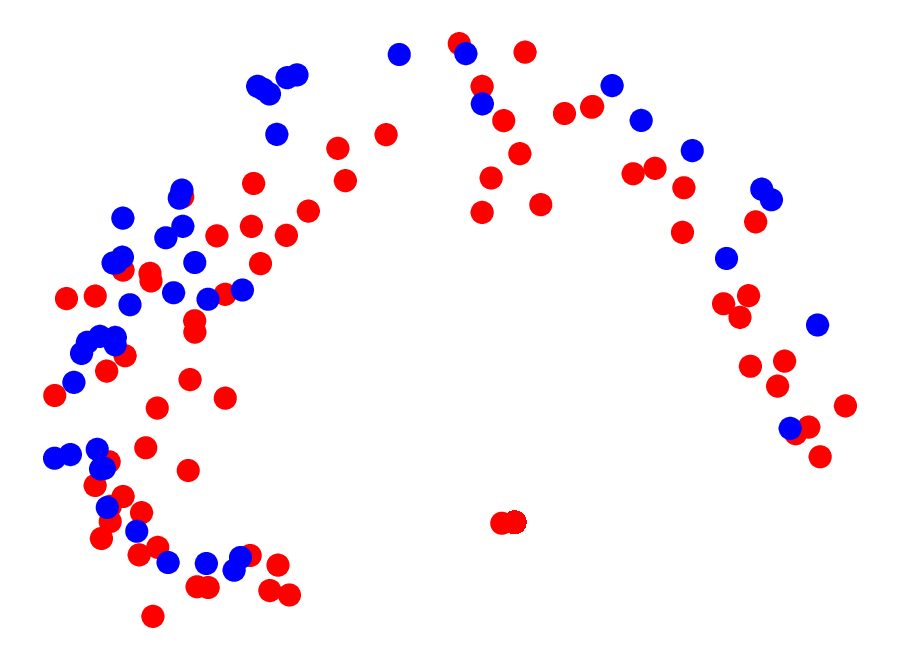}}
            \label{fig:politics-layer5}\hfill
            \subfloat[Layer=6]{
			\centering
	\includegraphics[scale=0.24]{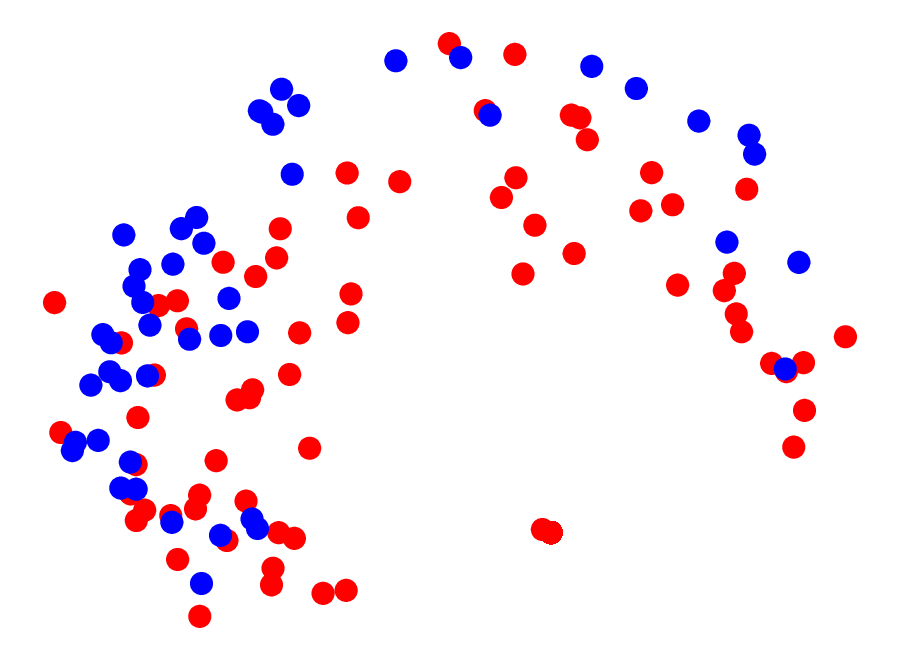}}
            \label{fig:politics-layer6}\hfill
        \subfloat[Layer=7]{
			\centering
	\includegraphics[scale=0.24]{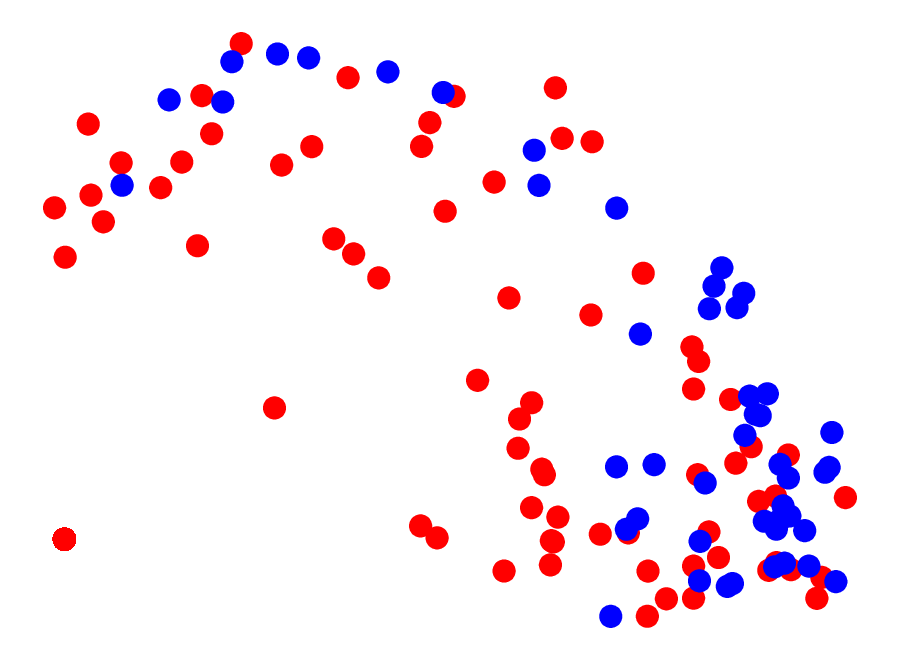}}
            \label{fig:politics-layer7}\hfill   
        \subfloat[Layer=8]{
			\centering
	\includegraphics[scale=0.24]{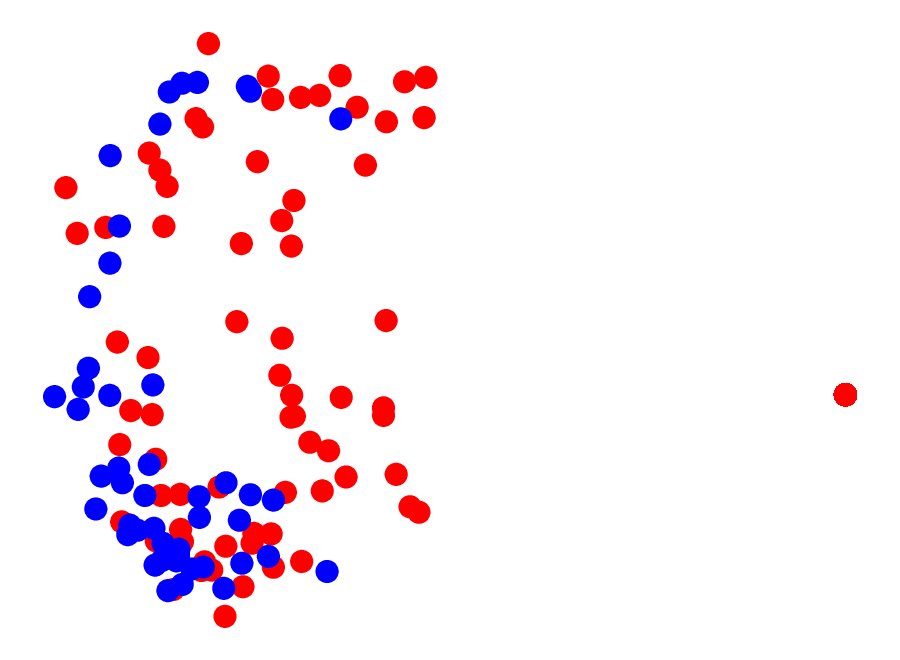}}
            \label{fig:politics-layer8}
            
        \subfloat[Layer=9]{
			\centering
	\includegraphics[scale=0.24]{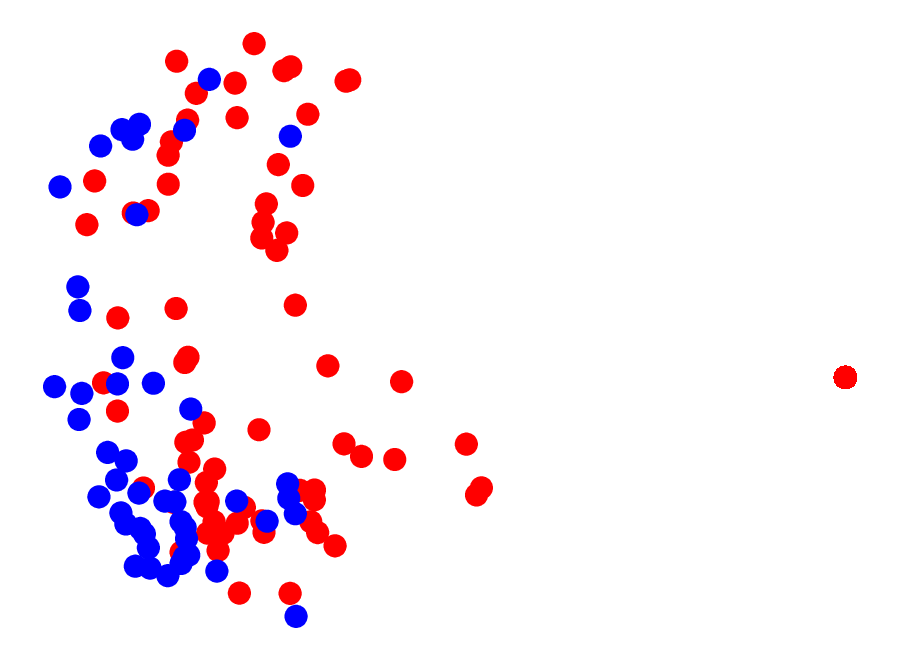}}
            \label{fig:politics-layer9}\hfill
        \subfloat[Layer=10]{
			\centering
	\includegraphics[scale=0.24]{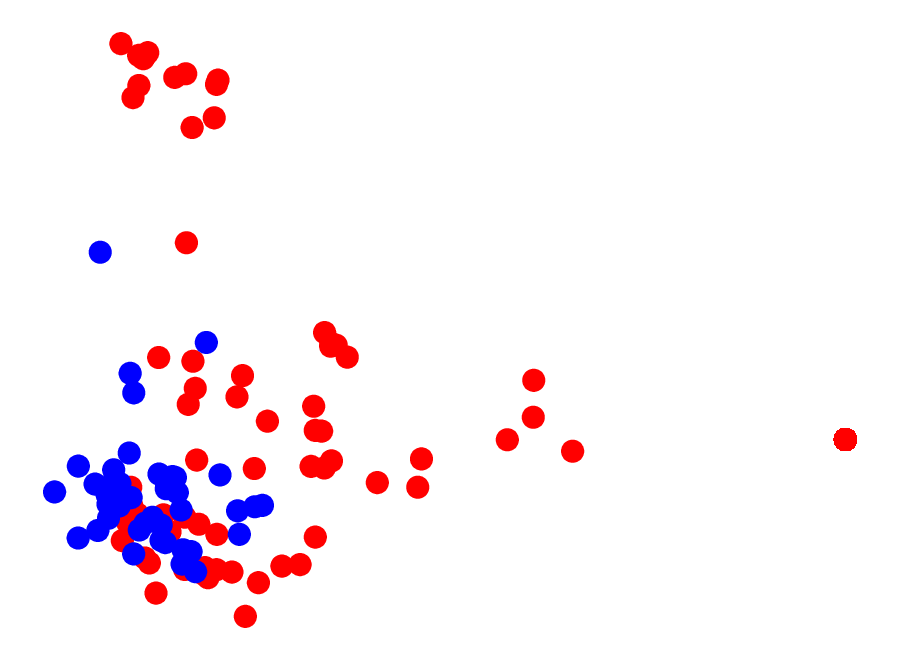}}
            \label{fig:politics-layer10}\hfill
        \subfloat[Layer=11]{
			\centering
	\includegraphics[scale=0.24]{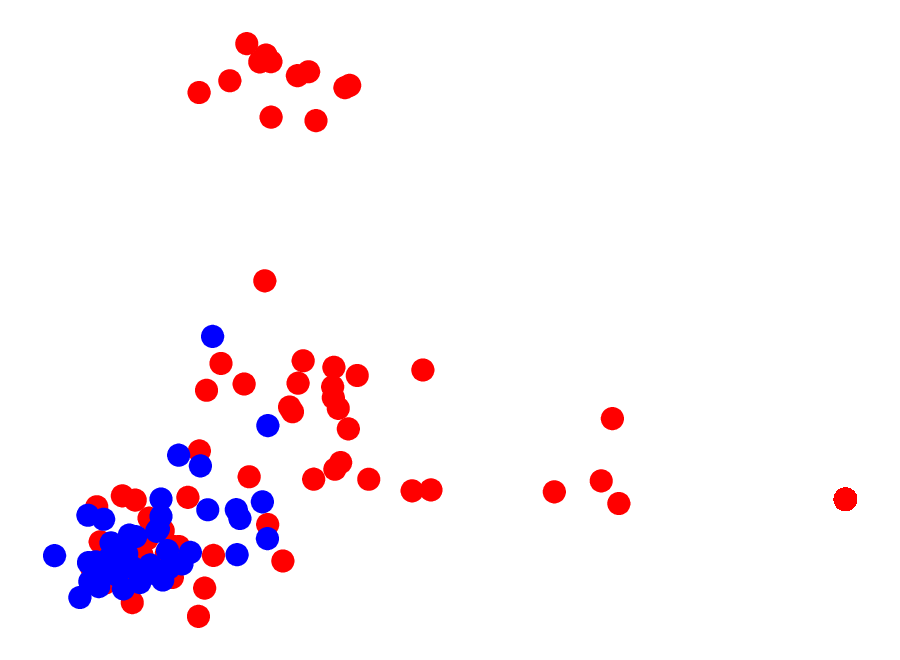}}
            \label{fig:politics-layer11}\hfill 
        \subfloat[Layer=12]{
			\centering
	\includegraphics[scale=0.24]{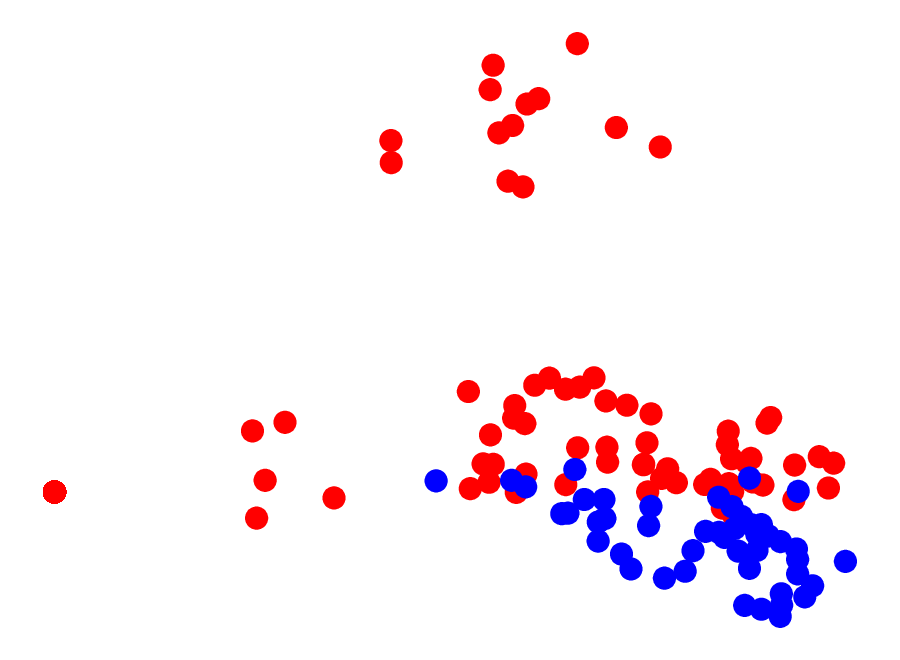}}
            \label{fig:politics-layer12}\hfill
		\caption{Intermediate-layer contextualized token embeddings for \texttt{president</w>} (red) and \texttt{country</w>} (blue) plotted on the plane of the first two principal components. The x-axis and y-axis represent the first and second principal components, respectively.
		}
		\label{fig:pca-visualization-politics}
\end{figure*}

\textbf{Visualization.} Figs.~\ref{fig:pca-visualization-they-them}, \ref{fig:pca-visualization-have-had}, and \ref{fig:pca-visualization-are-is} depict the visualization of contextualized token embeddings across various layers of the GPT-1 model, using $3,000$ sentences from the BookCorpus (consistent with the setting in Fig.\ref{fig:law}). We examine three token pairs: \texttt{they</w>} vs. \texttt{them</w>}, \texttt{have</w>} vs. \texttt{had</w>}, and \texttt{are</w>} vs. \texttt{is</w>}. Principal component analysis (PCA) is applied to project these contextualized token embeddings onto a two-dimensional plane. The results show a distinct and progressive separation of contextualized token embeddings within each pair as one moves from the lower to the upper layers of the model. Moreover, as illustrated in Fig. \ref{fig:science-analysis}, the law of equi-learning manifests when evaluating the GPT-1 model across distinct domains. Specifically, we analyzed $200$ sentences from MedRAG-Textbooks \cite{xiong2024benchmarking} for the medicine domain, $1000$ sentences from LegalBench \cite{guha2024legalbench} for the law domain, and $500$ sentences from US-Congressional-Speeches\footnote{See more in \url{https://huggingface.co/datasets/Eugleo/us-congressional-speeches}.} for the politics domain. Consistent with other probing datasets, the number of sampled sentences was determined by their average length, and sentences were truncated to a maximum length of $512$ tokens. For each domain, we visualized the embeddings of token sets across various layers of the GPT-1 model—\texttt{patients</w>}, \texttt{cells</w>}, and \texttt{disorder</w>} for medicine; \texttt{law</w>} and \texttt{policy</w>} for law; and \texttt{president</w>} and \texttt{country</w>} for politics. Similarly, PCA was employed to project these contextualized token embeddings onto a two-dimensional plane. As depicted in Fig.~\ref{fig:pca-visualization-medicine}, Fig.~\ref{fig:pca-visualization-law}, and Fig.~\ref{fig:pca-visualization-politics}, the results reveal a distinct and progressively increasing separation of contextualized token embeddings within each token set as the model transitions from lower to upper layers.

\begin{figure}[!htp]
    \centering
     \includegraphics[scale=0.5]{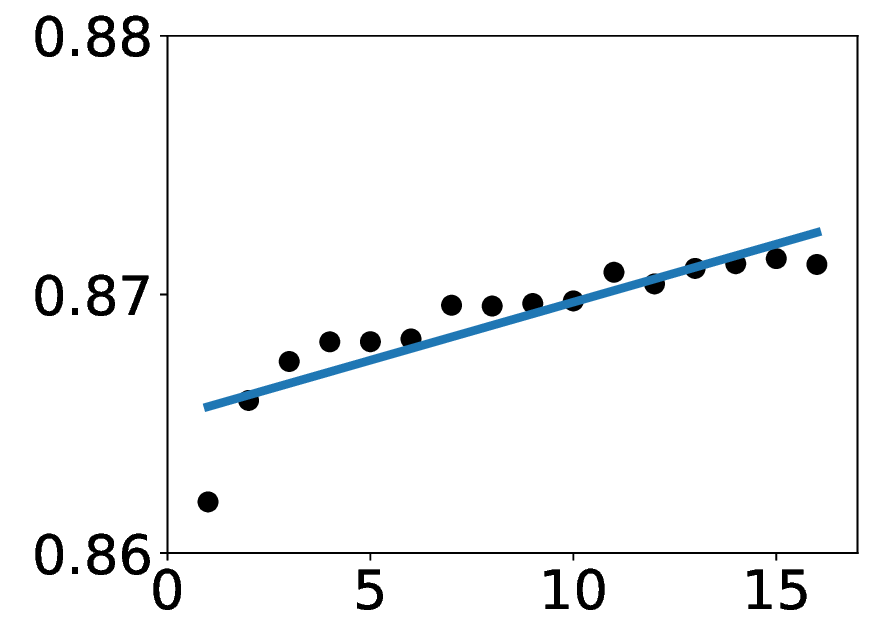}
    \caption{An enlarged version of Pythia-1B at initialization (Step=0). The x-axis denotes the layer index, while the y-axis (log scale) shows the prediction residual (PR) as defined in Eq.~\ref{eq:PR}.}
		\label{fig:pythia-1b-step0-original}
\end{figure}

\textbf{Training dynamics.} As illustrated in Figs.~\ref{fig:training-step}, \ref{fig:training-epoch}, and \ref{fig:repeat-number}, we examine the evolution of the observed law throughout the training process. Specifically, Fig.~\ref{fig:training-step} depicts the PR of contextualiezed token embeddings at each layer of Pythia-1B at various training steps (an enlarged version at initialization (Step=0) is in Fig.~\ref{fig:pythia-1b-step0-original}), using BookCorpus as the probing dataset. Additionally, Figs. \ref{fig:training-epoch} and \ref{fig:repeat-number} present the PR of contextualized token embeddings at each layer of pre-trained 2.8B GPT-2 models, as released by \cite{muennighoff2024scaling}, across different training epochs and data repetitions, with OpenWebText serving as the probing dataset. 

\textbf{Model scaling.} As illustrated in Fig.~\ref{fig:model-size}, the contextualized token embeddings are compared across different model sizes within the same model series. We analyze three distinct model series—GPT-2, RWKV-Raven, and Mamba—each with at least four different model sizes. The largest version of each series is depicted in Fig.~\ref{fig:law}. 

\textbf{Pre-training task.} As illustrated in Fig.~\ref{fig:task-formulation}, different probing tasks are employed to analyze the contextualized token embeddings of BERT, RoBERTa, and T5. 
Notably, the probing datasets used are peS2o for BERT and RoBERTa, and C4 for T5. Since masked language modeling (MLM) and span corruption (SC) mask or corrupt only 15\% of tokens, leading to a corresponding reduction to 15\% of the effective examples compared to NTP, we multiply the size of the probing datasets by 7 for MLM and SC. 


\textbf{Information flow.} As illustrated in Fig.~\ref{fig:token-prediction}, the contextualized token embeddings of the current token ($x_t$) at each layer are leveraged to predict various tokens in the sequence, spanning
previous token ($x_{t-1}$) to 
next next token ($x_{t+2}$), including the default next token ($x_{t+1}$). For clarity, we present four models selected from those shown in Fig.~\ref{fig:law}: GPT2-XL, Llama-3-8B-Instruct, RWKV-Raven-14B, and Mamba-2.8B. For simplicity, we present an explicit formulation for predicting the previous token, $x_{t-1}$; analogous formulas for predicting other tokens can be obtained by appropriately adjusting the prediction target in $\text{PR}^{\text{prev}}$ and \( \text{PR}^{\text{prev}}_\ell \). Specifically, the model uses the last-layer embedding of the current token, denoted as $\mathbf{h}_{t, \text{last}}$, to predict the \textit{previous} token in the sequence. Let $x^{s}_{t-1}$ represent the immediately preceding token that the LLM aims to predict based on the first $t$ tokens, $x^{s}_1, x^{s}_2, \ldots, x^{s}_t$, through the last-layer embedding of the current token $\mathbf{h}_{t, \text{last}}^{s}$, for an index $1 \le s \le S$ over all training corpus. This process forms a dataset $\mathcal{D}_{\text{prev}} := \{(\mathbf{h}_{t, \text{last}}^{s}, x^{s}_{t-1}) \mid 1 \leq s \leq S\}$. To assess the capability of the LLM's current token embeddings in predicting the previous token, we evaluate how well a linear regression model fits on the dataset $\mathcal{D}_{\text{prev}}$. For this purpose, we identify $x$ with its index in the token vocabulary. Let $\hat x_{\text{prev}} = \mathbf{w} \cdot \mathbf{h} + b$ denote the least-squares fit on $\mathcal{D}_{\text{prev}}$. This suggests using the following metric to quantify the LLM's previous-token prediction capability:
\begin{equation}\label{eq:PR}
\text{PR}^{\text{prev}} :=  \frac{\sum (x_{\text{prev}}- \hat x_{\text{prev}})^2}{\sum (x_{\text{prev}} - \bar x_{\text{prev}})^2},
\end{equation}
where the sum is over all $x_{\text{prev}} = x^{s}_{t-1}$, $\hat x_{\text{prev}} = \mathbf{w} \cdot \mathbf{h}_{t, \text{last}}^{s} + b$, and $\bar x_{\text{prev}}$ represents the mean of all $x^{s}_{t-1}$. To investigate how the predictive power of an LLM with depth $L$ evolves across its layers, we calculate the $\text{PR}^{\text{prev}}$ for the previous-token prediction task at each intermediate layer. Let $\text{PR}^{\text{prev}}_l$ denote this value for the $l$-th layer, where $1 \le l \le L$. Specifically, instead of using the last-layer embedding \( \mathbf{h}_{t,\text{last}} \equiv \mathbf{h}_{t,L} \), we use the embedding of the current token at layer \( \ell \), denoted \( \mathbf{h}_{t,\ell} \), to predict the previous token when computing \( \text{PR}^{\text{prev}}_\ell \).

\subsection*{Additional Results}

\begin{figure*}[!htp]
    \captionsetup[subfigure]{labelformat=empty}
		\centering
   \rotatebox[y=1.0cm]{90}{\footnotesize Separation Fuzziness}\quad
            \subfloat[GPT-2 XL]{
			\centering
		\includegraphics[scale=0.22]{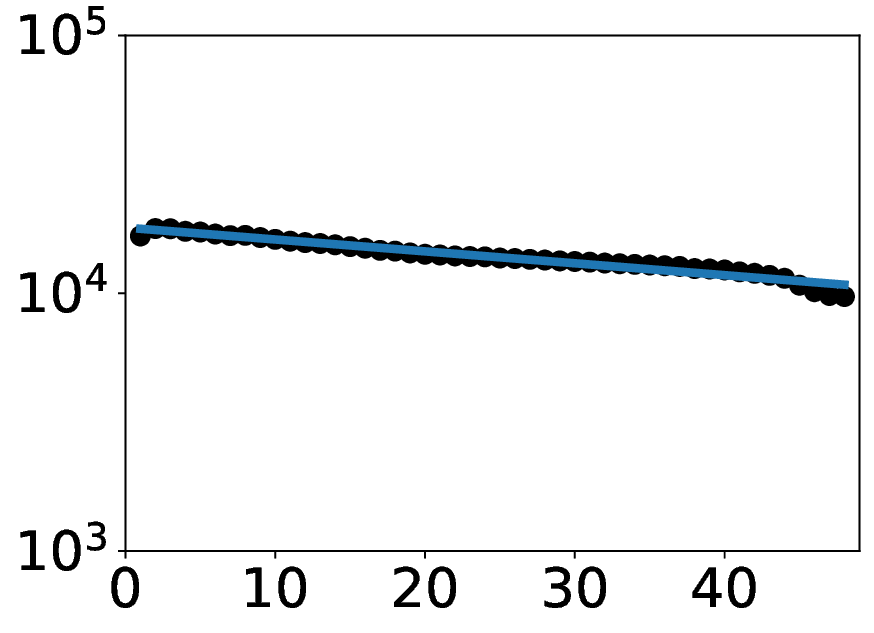}}
     \label{fig:gpt2-xl-separation}\hfill
     	\subfloat[Llama-3-8B-Instruct]{
			\centering
		\includegraphics[scale=0.22]{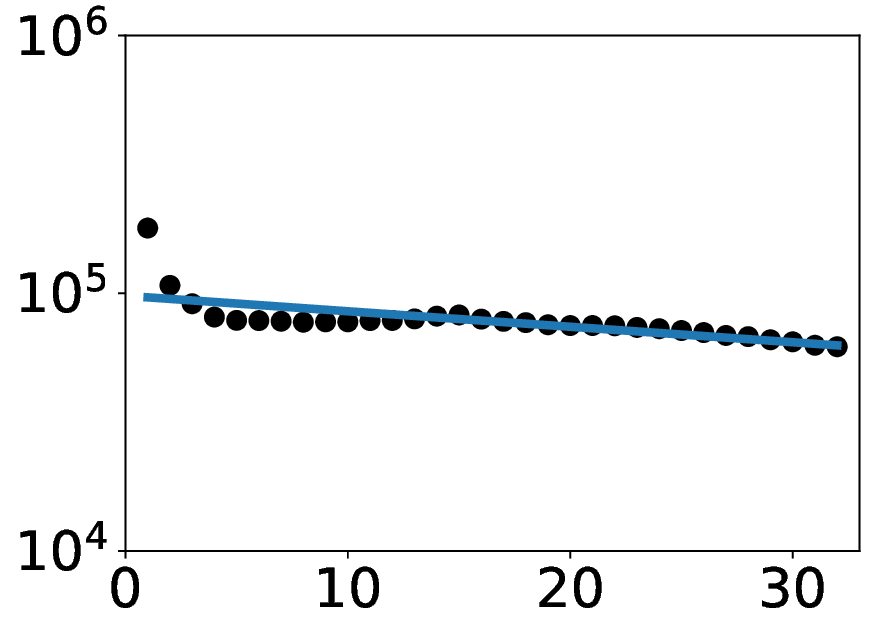}}
            \label{fig:llama-3-8b-it-separation}\hfill
        \subfloat[RWKV-Raven-14B]{
			\centering
		\includegraphics[scale=0.22]{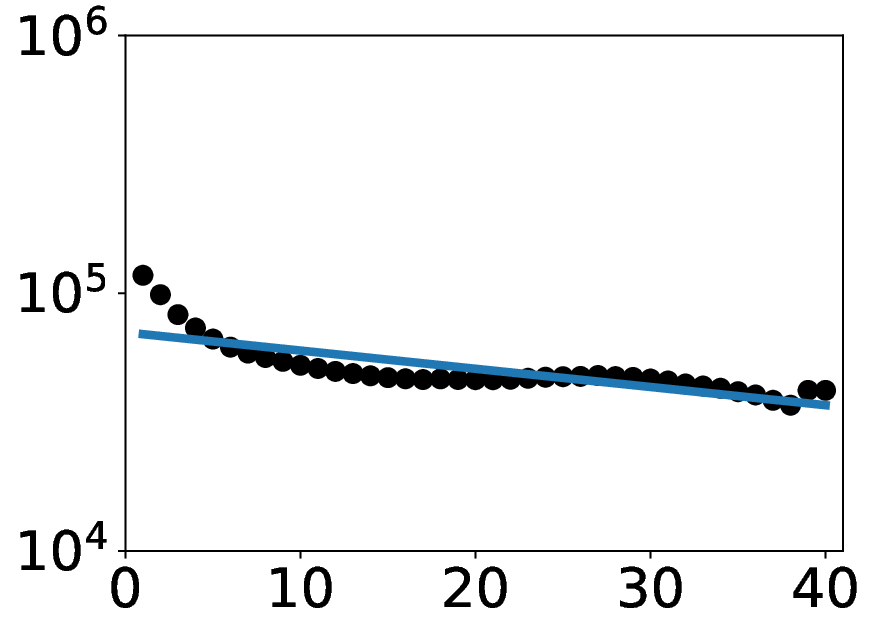}}
            \label{fig:rwkv-raven-14b-separation}\hfill
            \subfloat[Mamba-2.8B]{
			\centering
		\includegraphics[scale=0.22]{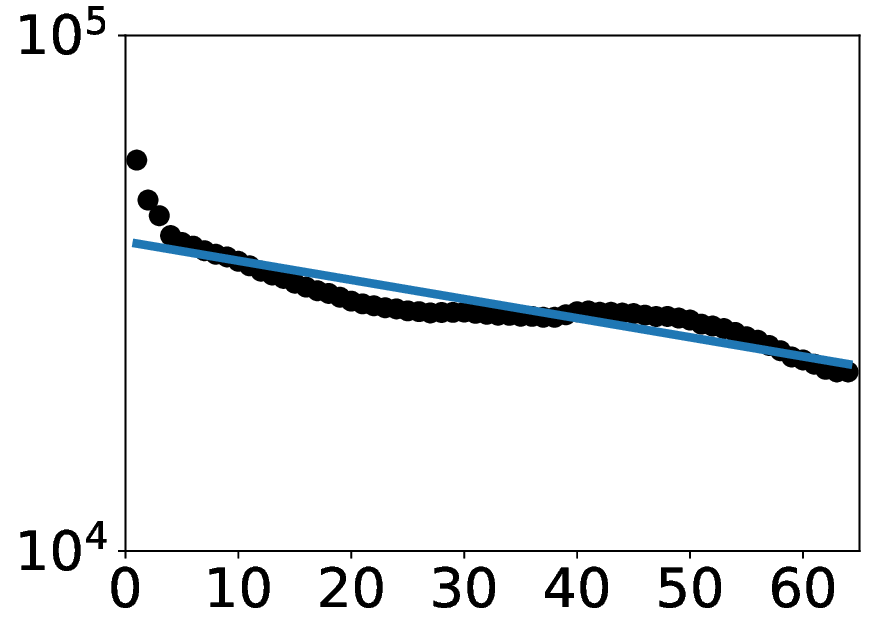}}
            \label{fig:mamba-2.8b-separation}\hfill
		\caption{With the measure of separation fuzziness, the law of equi-learning is not clear though the decreasing trend is still observed. The x-axis denotes the layer index, while the y-axis (log scale) shows the separation fuzziness.
		}
		\label{fig:separation-fuzziness}
\end{figure*}

\begin{figure*}[!htp]
    \captionsetup[subfigure]{labelformat=empty}
		\centering
          \rotatebox[y=1.0cm]{90}{\footnotesize Shuffled Vocabulary}\quad
            \subfloat[GPT-2 XL]{
			\centering
		\includegraphics[scale=0.22]{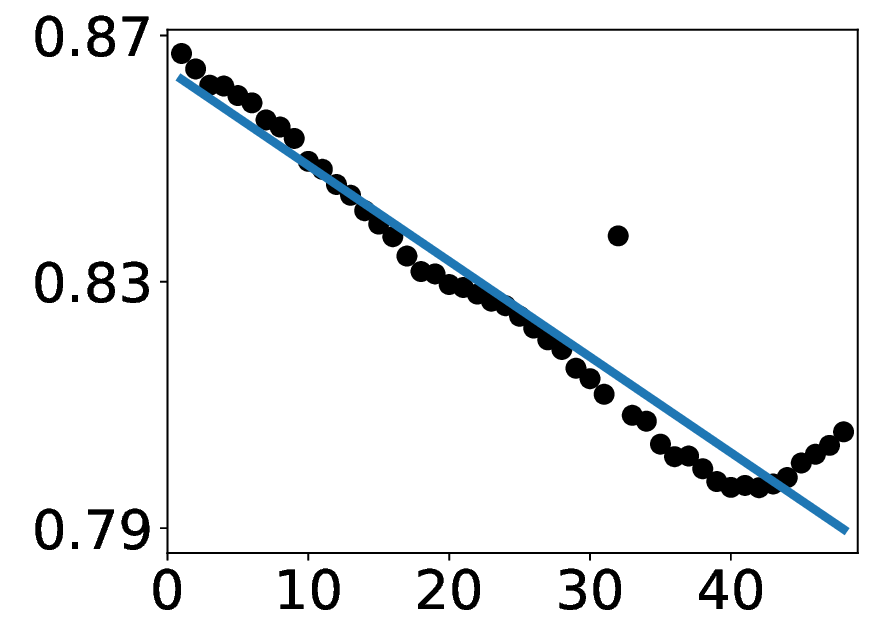}}
     \label{fig:gpt2-xl-shuffled}\hfill
     	\subfloat[Llama-3-8B-Instruct]{
			\centering
		\includegraphics[scale=0.22]{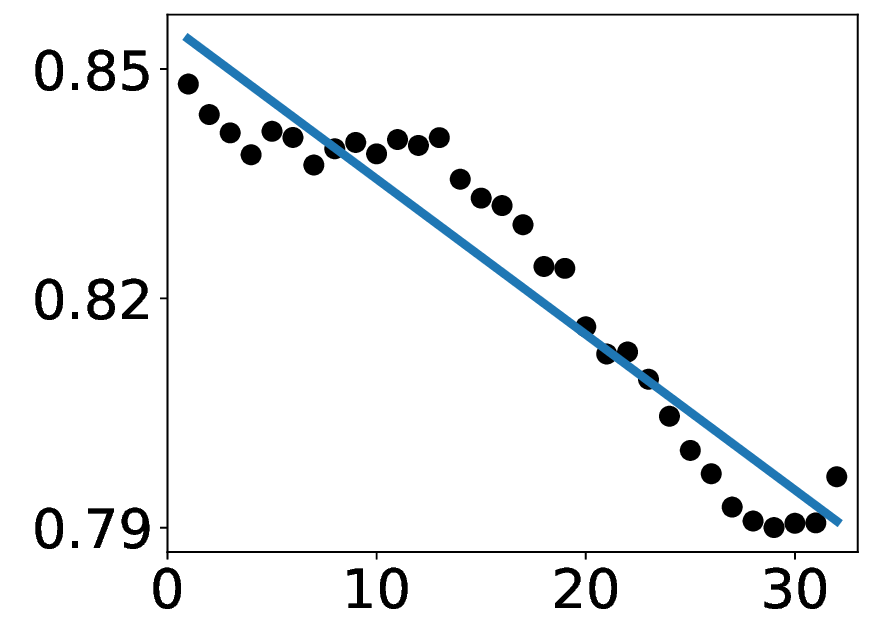}}
            \label{fig:llama-3-8b-it-shuffled}\hfill
        \subfloat[RWKV-Raven-14B]{
			\centering
		\includegraphics[scale=0.22]{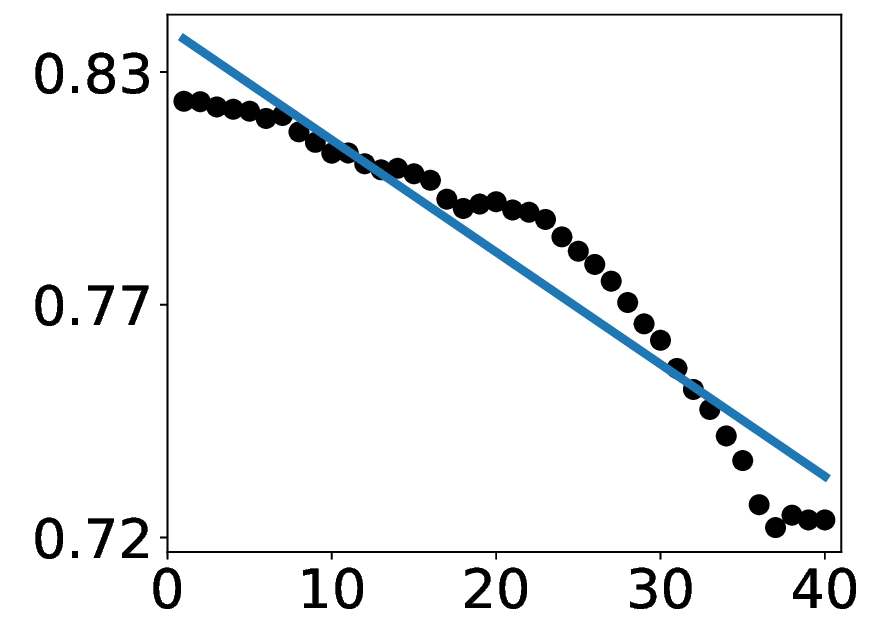}}
            \label{fig:rwkv-raven-14b-shuffled}\hfill
            \subfloat[Mamba-2.8B]{
			\centering
		\includegraphics[scale=0.22]{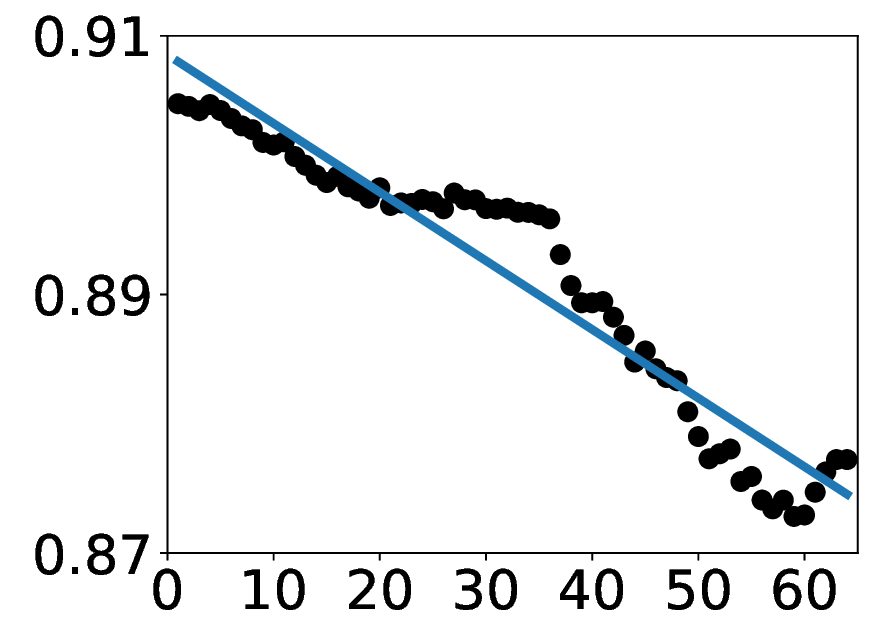}}
            \label{fig:mamba-2.8b-shuffled}\hfill
		\caption{When the vocabulary is shuffled, the law of equi-learning is not very clear though the decreasing trend is still observed. The x-axis denotes the layer index, while the y-axis (log scale) shows the prediction residual (PR) as defined in Eq.~\ref{eq:PR}.
		}
		\label{fig:shuffled-vocabulary}
\end{figure*}

\textbf{Measure analysis.} Fig.\ \ref{fig:separation-fuzziness} illustrates that the widely used metric of separation fuzziness, commonly applied to assess deep learning features in classification tasks \cite{papyan2020prevalence, fang2021exploring, he2023law}, is inadequate for the emergence of the law of equi-learning in LLMs. This inadequacy may stem from the larger token vocabulary size compared to the embedding dimension and the presence of very similar or even identical contexts followed by different tokens in natural language data \cite{wu2024linguistic}. For simplicity, we selected four models from Fig.~\ref{fig:law}: GPT-2 XL, Llama-3-8B-Instruct, RWKV-Raven-14B, and Mamba-2.8B, and utilized separation fuzziness to evaluate the quality of contextualized token embeddings. Furthermore, as demonstrated in Fig.~\ref{fig:shuffled-vocabulary}, the law of equi-learning becomes obscured when the vocabulary is shuffled, resulting in differing token index orders. This observation suggests that the widely adopted byte pair encoding algorithm \cite{sennrich2016neural} in tokenizers can produce a meaningful token index order, which is critical to the emergence of the law of equi-learning. Further investigation is needed and is deferred to future work.

\begin{figure*}[!htp]
    \captionsetup[subfigure]{labelformat=empty}
		\centering
   \rotatebox[y=1.3cm]{90}{\footnotesize No LN}\quad
            \subfloat[GPT-2 XL]{
			\centering
		\includegraphics[scale=0.22]{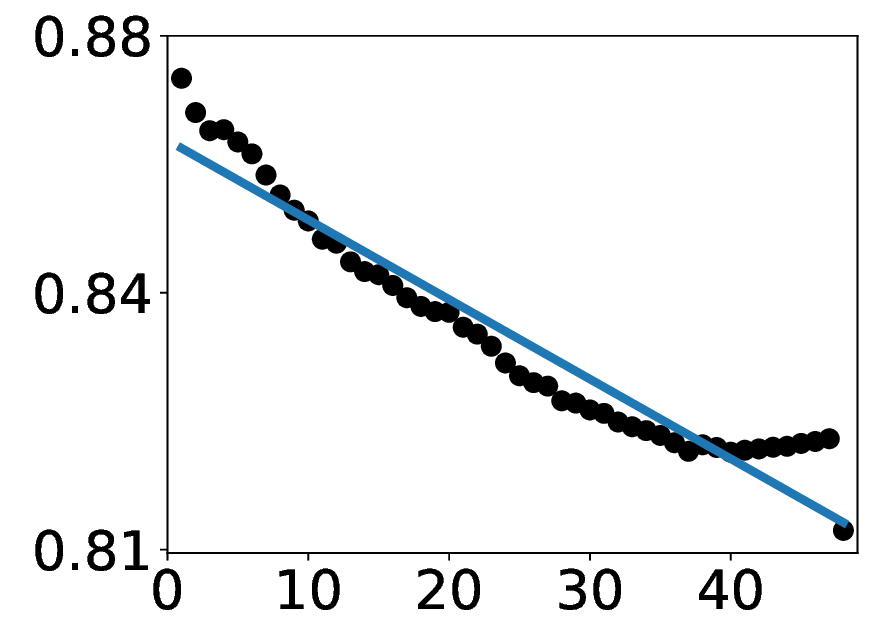}}
     \label{fig:gpt2-xl-no-LN}\hfill
     	\subfloat[Llama-3-8B-Instruct]{
			\centering
		\includegraphics[scale=0.22]{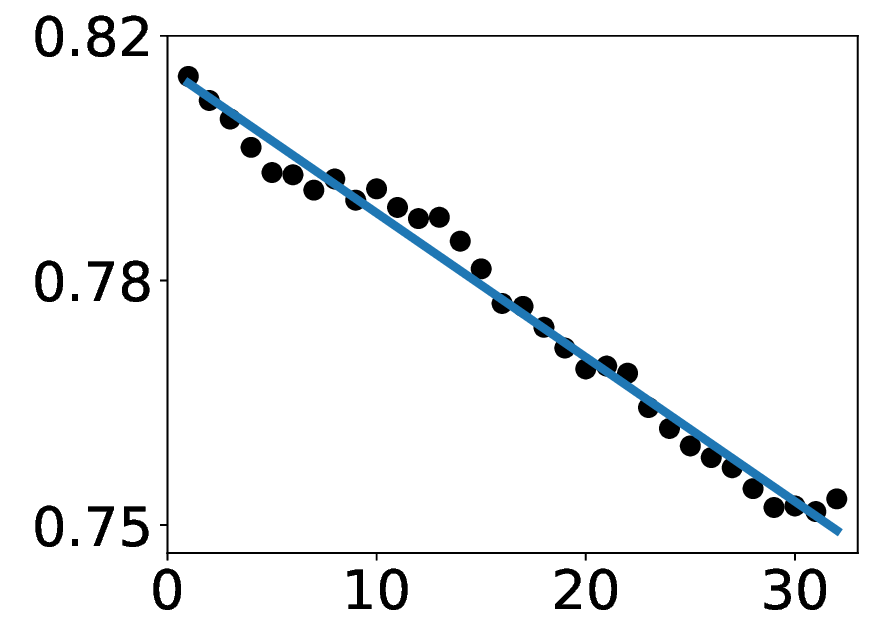}}
            \label{fig:llama-3-8b-it-no-LN}\hfill
        \subfloat[RWKV-Raven-14B]{
			\centering
		\includegraphics[scale=0.22]{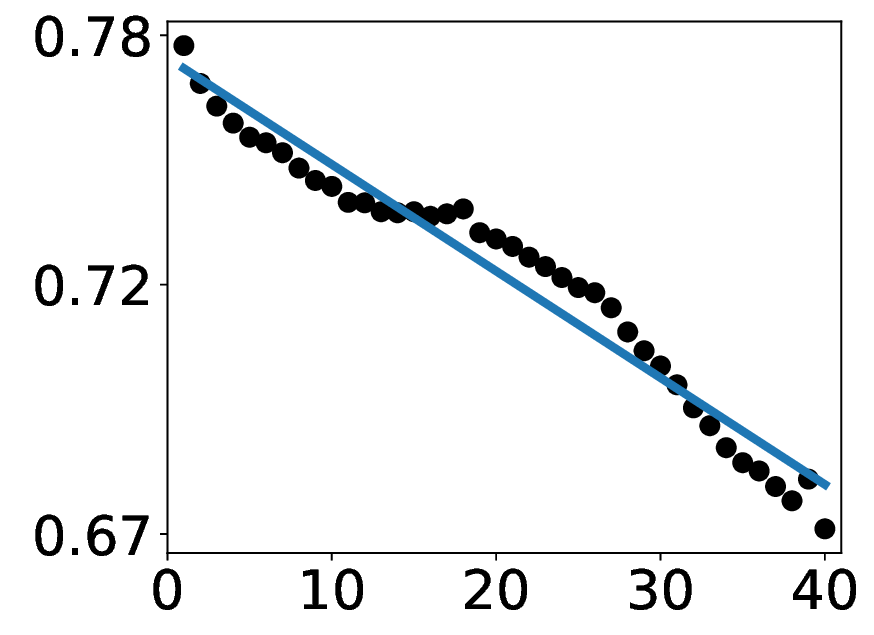}}
            \label{fig:rwkv-raven-14b-no-LN}\hfill
            \subfloat[Mamba-2.8B]{
			\centering
		\includegraphics[scale=0.22]{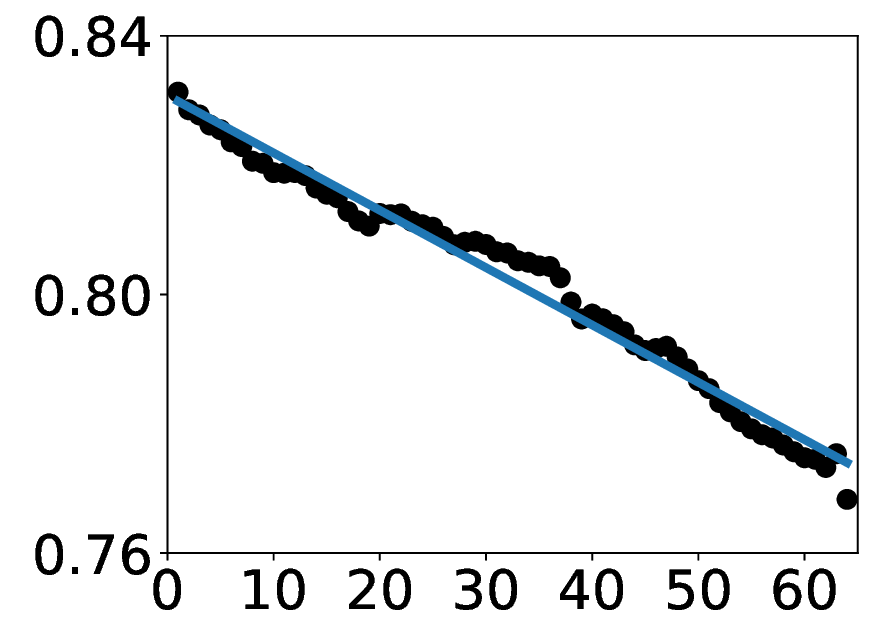}}
            \label{fig:mamba-2.8b-no-LN}\hfill

       \rotatebox[y=1.3cm]{90}{\footnotesize Standardization}\quad
        \subfloat[GPT-2 XL]{
			\centering
		\includegraphics[scale=0.22]{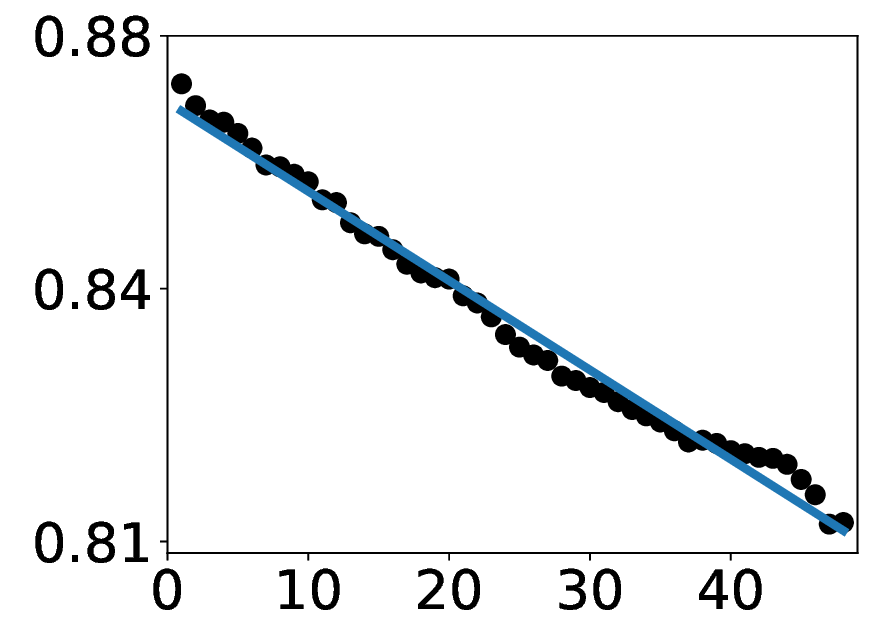}}
     \label{fig:gpt2-xl-standardization}\hfill
     	\subfloat[Llama-3-8B-Instruct]{
			\centering
		\includegraphics[scale=0.22]{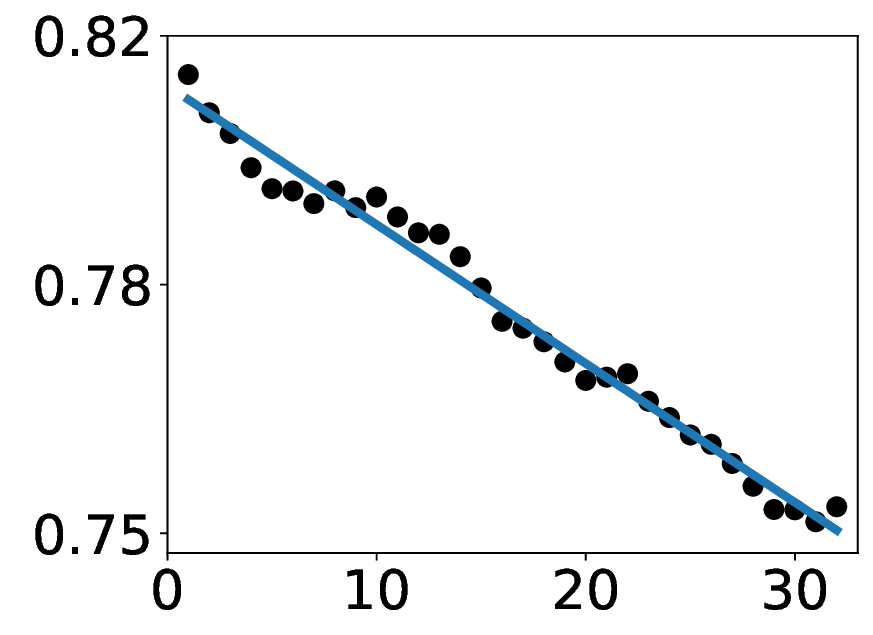}}
            \label{fig:llama-3-8b-it-standardization}\hfill
        \subfloat[RWKV-Raven-14B]{
			\centering
		\includegraphics[scale=0.22]{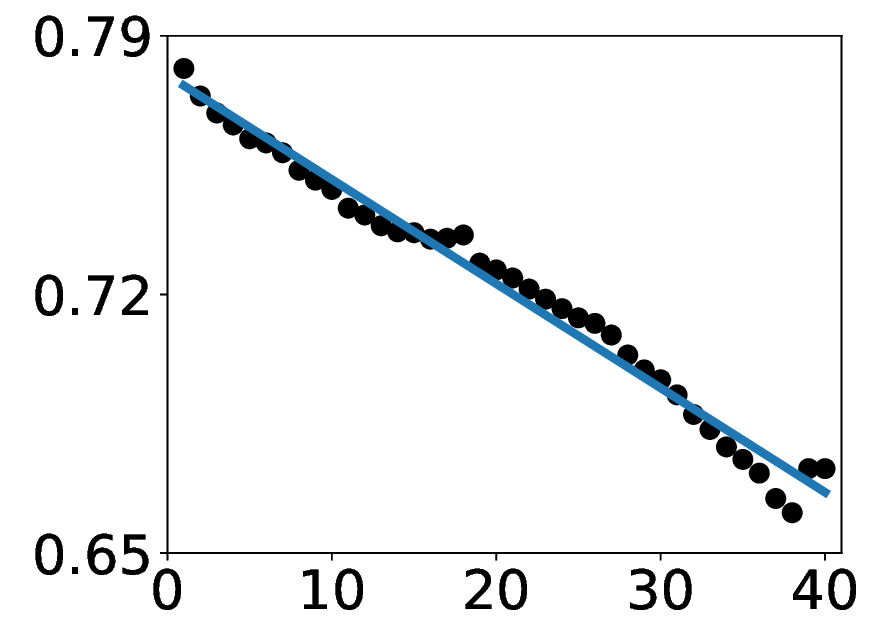}}
            \label{fig:rwkv-raven-14b-standardization}\hfill
            \subfloat[Mamba-2.8B]{
			\centering
		\includegraphics[scale=0.22]{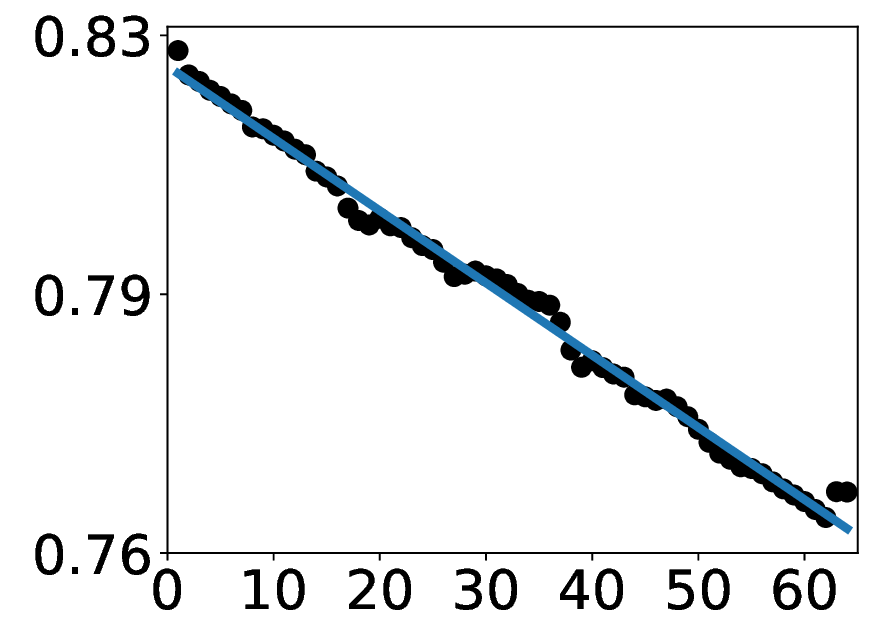}}
            \label{fig:mamba-2.8b-standardization}\hfill

           \rotatebox[y=1.3cm]{90}{\footnotesize Default}\quad
             \subfloat[GPT-2 XL]{
			\centering
		\includegraphics[scale=0.22]{figures/gpt2-xl_bookcorpus_pretrained_features_size=3000_seed=666.eps}}
     \label{fig:gpt2-xl-default}\hfill
     	\subfloat[Llama-3-8B-Instruct]{
			\centering
		\includegraphics[scale=0.22]{figures/Meta-Llama-3-8B-Instruct_bookcorpus_pretrained_features_size=3000_seed=666.eps}}
            \label{fig:llama-3-8b-it-default}\hfill
        \subfloat[RWKV-Raven-14B]{
			\centering
		\includegraphics[scale=0.22]{figures/rwkv-raven-14b_c4_pretrained_features_size=200_seed=666.eps}}
            \label{fig:rwkv-raven-14b-default}\hfill
            \subfloat[Mamba-2.8B]{
			\centering
		\includegraphics[scale=0.22]{figures/mamba-2.8b-hf_bookcorpus_pretrained_features_size=3000_seed=666.eps}}
            \label{fig:mamba-2.8b-default}\hfill
		\caption{For pre-LN models, omitting the additional initialized layer normalization (no LN) for contextualized token embeddings leads to a noisier manifestation of the law of equi-learning compared to the default approach with initialized layer normalization. This normalization effect can also be achieved through standardization, rather than relying on initialized layer normalization. The x-axis denotes the layer index, while the y-axis (log scale) shows the prediction residual (PR) as defined in Eq.~\ref{eq:PR}.
		}
		\label{fig:layer-normalization}
\end{figure*}

\textbf{Layer normalization analysis.} As illustrated in Fig.~\ref{fig:layer-normalization}, layer normalization plays a critical role in pre-LN models, as the absence of additional layer normalization makes the law noisier. Notably, this normalization effect can also be attained through standardization, specifically $z = \frac{x - \mu}{\sigma}$ across the embedding dimension. For simplicity, we selected four models from Fig.~\ref{fig:law}—GPT-2 XL, Llama-3-8B-Instruct, RWKV-Raven-14B, and Mamba-2.8B—and analyzed the impact of layer normalization.  This finding aligns with observations related to the equi-separation law in MLPs \cite{he2023law}, where batch normalization is crucial for its emergence. It is important to note that batch normalization does not impact the PR of contextualized token embeddings in our case.

\begin{table*}[t]
\centering
\scalebox{0.74}{
\begin{tabular}{l|c|c|c|c|c|c|c|c}
\Xhline{2\arrayrulewidth} \hline
  & BookCorpus & C4 & OpenWebText & Wikipedia & peS2o & Pile & Redpajama & OSCAR  \bigstrut[t] \bigstrut[b]   \\ \hline
 GPT-1 & {\bf -0.997} & -0.951  & -0.959 & -0.943  & -0.972 & -0.917 & -0.884 & -0.969 \\ \hline
GPT-2 XL & {\bf -0.994}  & -0.960  & -0.982  & -0.962 & -0.963 & -0.965 & -0.962 & -0.957 \\ \hline
 Llama-1-13B & -0.956 & -0.993  & -0.990 & -0.986 & {\bf -0.994} & -0.992 & -0.993 & -0.993 \\ \hline
 Llama-2-13B & -0.913 & -0.985  & -0.977 & -0.966 & {\bf -0.988} & -0.973 & -0.983 & -0.985 \\ \hline
  Llama-2-13B-Chat & -0.879 & -0.984  & -0.967 & -0.962 & {\bf -0.983} & -0.964 & -0.978 & -0.980 \\ \hline
 Llama-3-8B & {\bf -0.993} & -0.981  & -0.977 & -0.979 & -0.938 & -0.941 & -0.948 & -0.969  \\ \hline
 Llama-3-8B-Instruct & {\bf -0.992} & -0.981   & -0.974 & -0.979 & -0.940 & -0.940 & -0.948 & -0.967  \\ \hline
 Mistral-7B-v0.1 & -0.874 & {\bf -0.988}  & -0.950 & -0.968 & -0.956 & -0.941 & -0.940 & -0.979 \\ \hline
  Mistral-7B-Instruct-v0.1 & -0.850  & {\bf -0.991}  & -0.958 & -0.971 & -0.961 & -0.939 & -0.948 & -0.985  \\ \hline
 Mistral-7B-v0.2 & -0.863 & {\bf -0.989}  & -0.952 & -0.967 & -0.953 & -0.936 & -0.941 & -0.979  \\ \hline
 Mistral-7B-Instruct-v0.2 & -0.874 & {\bf -0.988}  & -0.952 & -0.967 & -0.958 & -0.931 & -0.936 & -0.979 \\ \hline
 Mistral-7B-v0.3 & -0.863 & {\bf -0.989}  & -0.952  & -0.967 & -0.953 & -0.936 & -0.941 & -0.979  \\ \hline
  Mistral-7B-Instruct-v0.3 & -0.863  & {\bf -0.988}   & -0.951  & -0.966 & -0.951 & -0.932 & -0.938 & -0.979 \\ \hline
 phi-1.5 & {\bf -0.994}  & -0.967  & -0.986 & -0.976 & -0.974 & -0.958 & -0.967& -0.971  \\ \hline
 phi-2 & {\bf -0.993} & -0.984  & -0.978  & -0.983  & -0.890 & -0.954 & -0.930 & -0.987 \\ \hline
 phi-3-medium-4k-instruct & -0.894 &  {\bf -0.992}  & -0.961 & -0.955 & -0.977 & -0.972 & -0.975 & -0.991 \\ \hline
  phi-3-medium-128k-instruct & -0.902 & {\bf -0.992}  & -0.962 & -0.959 & -0.975 & -0.972 & -0.974 & -0.991 \\ \hline
 RWKV-14B & -0.984 & {\bf -0.991}  & -0.983 & -0.987 & -0.967 & -0.984 & -0.952 & -0.983 \\ \hline
RWKV-Raven-14B & -0.984 & {\bf -0.991}  & -0.982 & -0.984 & -0.964 & -0.984 & -0.952 & -0.981 \\ \hline
 Mamba-2.8B & {\bf -0.997} & -0.981  & -0.989 & -0.987 & -0.992 & -0.981 & -0.966 & -0.986 \bigstrut[t] \bigstrut[b]   \\ \hline
\Xhline{2\arrayrulewidth}
\end{tabular}}
\caption{Pearson correlation coefficients between the logarithm of PR values (Eq.~\ref{eq:PR}) and layer indices for all LLMs in Fig.~\ref{fig:law}, evaluated across eight probing datasets. For each LLM, the probing dataset with the strongest Pearson correlation is highlighted in bold.
}
\label{table:probing-data}
\end{table*}

\begin{figure*}[!htp]
    \captionsetup[subfigure]{labelformat=empty}
		\centering
     \rotatebox[y=1.3cm]{90}{\footnotesize RedPajama}\quad
            \subfloat[GPT-2 XL]{
			\centering
		\includegraphics[scale=0.22]{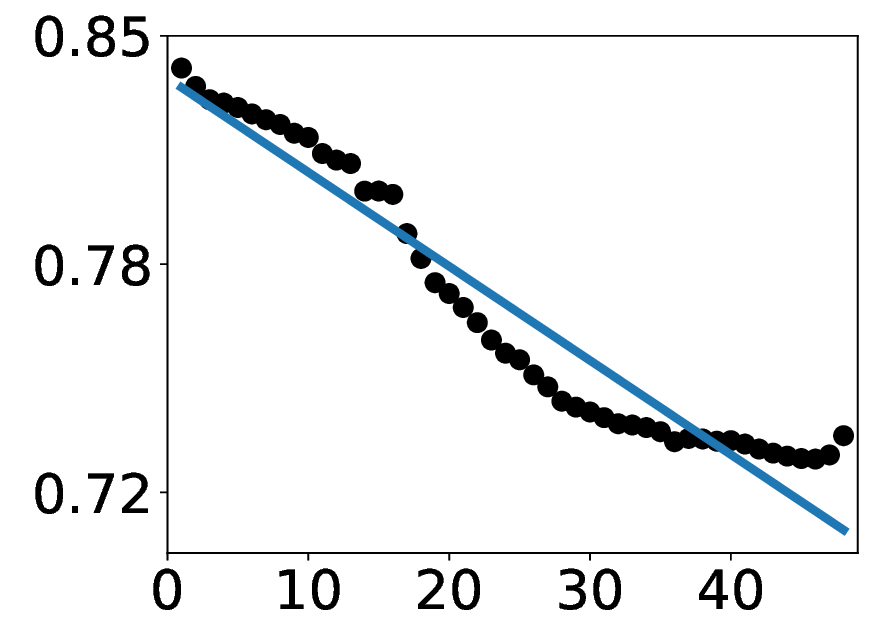}}
     \label{fig:gpt2-xl-redpajama}\hfill
     	\subfloat[Llama-3-8B-Instruct]{
			\centering
		\includegraphics[scale=0.22]{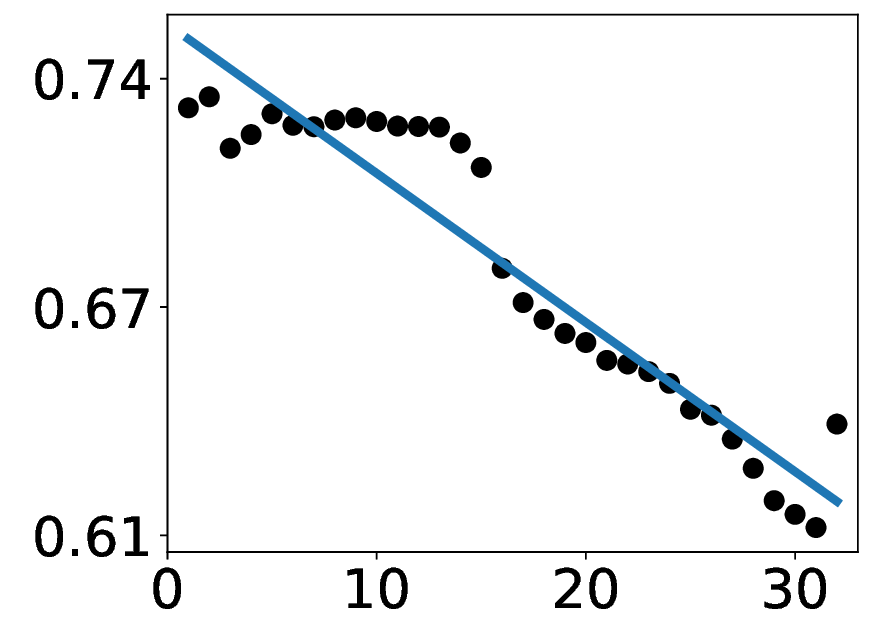}}
            \label{fig:llama-3-8b-it-redpajama}\hfill
        \subfloat[RWKV-Raven-14B]{
			\centering
		\includegraphics[scale=0.22]{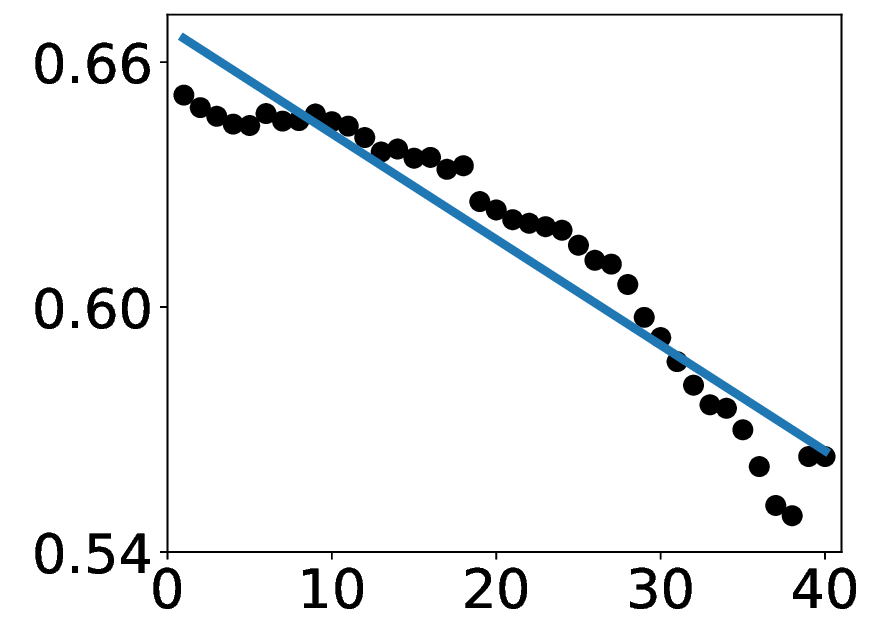}}
            \label{fig:rwkv-raven-14b-redpajama}\hfill
            \subfloat[Mamba-2.8B]{
			\centering
		\includegraphics[scale=0.22]{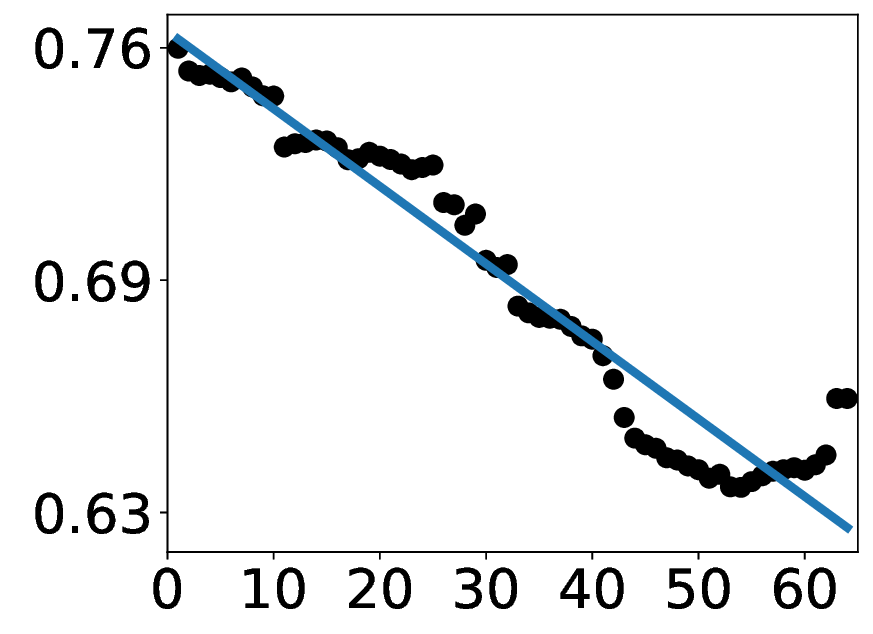}}
            \label{fig:mamba-2.8b-redpajama}\hfill
		\caption{With unsuitable probing dataset (e.g., RedPajama here), the law of equi-learning is not clear, though a descending trend is still observed. The x-axis denotes the layer index, while the y-axis (log scale) shows the prediction residual (PR) as defined in Eq.~\ref{eq:PR}.
		}
		\label{fig:probing-data}
\end{figure*}

\textbf{Probing data analysis.} Table \ref{table:probing-data} highlights that the probing datasets exhibiting the strongest Pearson correlations for various LLMs in Fig.\ref{fig:law} are BookCorpus, C4, and peS2o among the eight evaluated datasets. This trend may be attributed to the high quality of these datasets and their resemblance to the pre-training data used for these models. Notably, the optimal probing dataset among the eight evaluated transitions from peS2o to BookCorpus as models progress from Llama-1 and Llama 2 (including its chat variant) to Llama 3 (and its instruct variant). This shift suggests that higher-quality pre-training data may require correspondingly higher-quality probing datasets to facilitate the emergence of the law of equi-learning, likely reflecting the impact of carefully designed pre-processing and curation pipelines used to enhance the pre-training data quality in Llama 3 \cite{dubey2024llama}. Furthermore, as demonstrated in Fig.~\ref{fig:probing-data}, the observed law's clarity diminishes when an inappropriate probing dataset is selected, although a general descending trend remains evident. For simplicity, we selected four models from Fig.~\ref{fig:law}—GPT-2 XL, Llama-3-8B-Instruct, RWKV-Raven-14B, and Mamba-2.8B—and presented the PR of their token embeddings using the RedPajama dataset as the probing dataset. These findings underscore the critical importance of selecting appropriate probing data for the emergence of the law of equi-learning.

\begin{figure*}[!htp]
    \captionsetup[subfigure]{labelformat=empty}
		\centering
     \rotatebox[y=1.8cm]{90}{\footnotesize C4}\quad
            \subfloat[GPT-2 2.8B]{
			\centering
		\includegraphics[scale=0.32]{figures/gpt2-2b855b55bc4_global_step52452_openwebtext_pretrained_features_size=100_seed=666.eps}}
            \label{fig:c4-2.8b-openwebtext}\hfill
            \subfloat[GPT-2 4.2B]{
			\centering
		\includegraphics[scale=0.32]{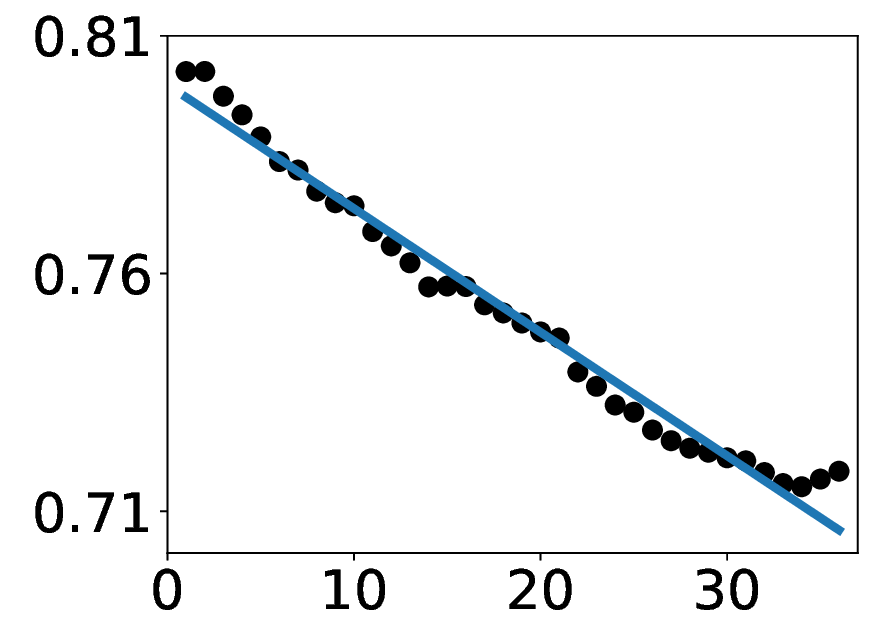}}
            \label{fig:c4-4.2b-openwebtext}\hfill
     	\subfloat[GPT-2 8.7B]{
			\centering
		\includegraphics[scale=0.32]{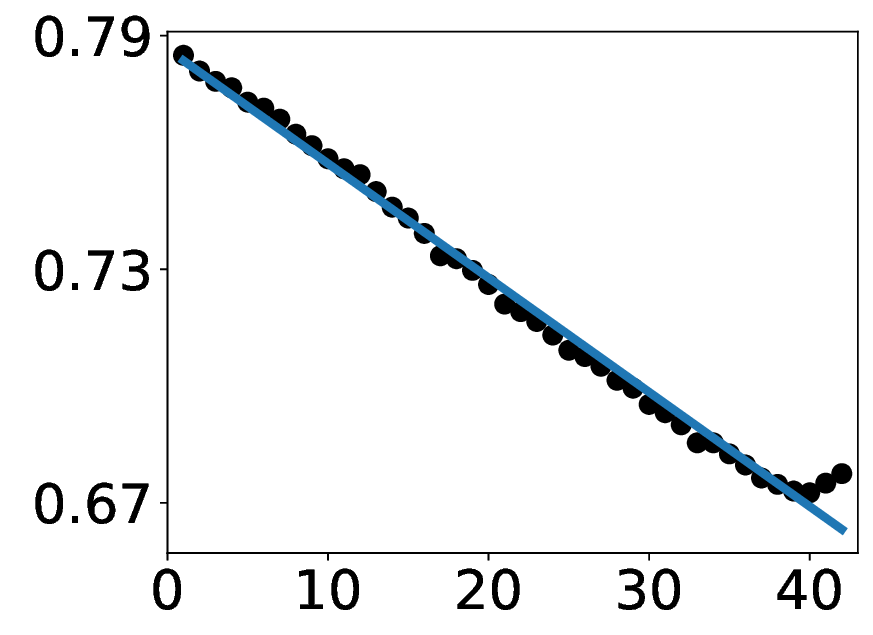}}
            \label{fig:c4-8.7b-openwebtext}\hfill

       \rotatebox[y=1.8cm]{90}{\footnotesize OSCAR}\quad
        \subfloat[GPT-2 2.8B]{
			\centering
		\includegraphics[scale=0.32]{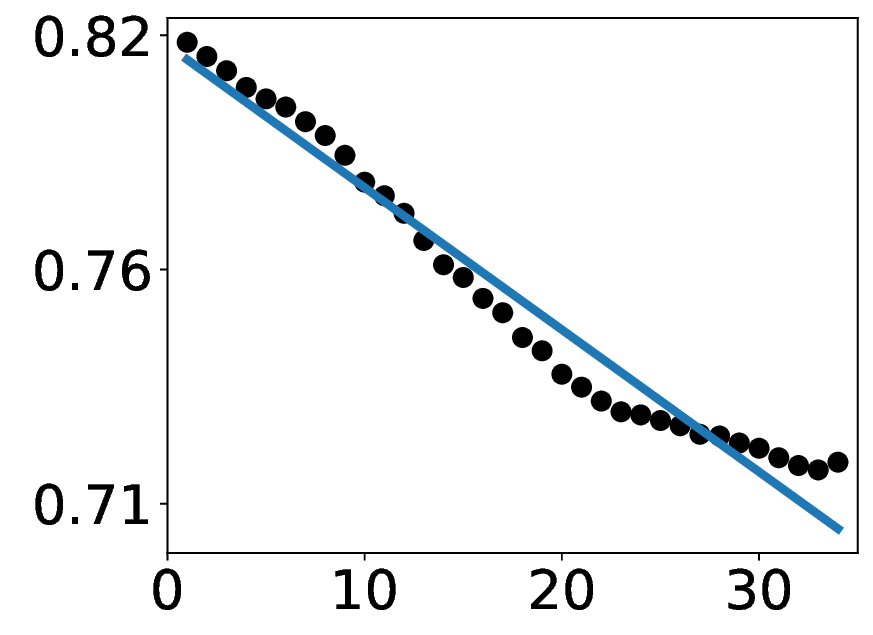}}
            \label{fig:oscar-2.8b-openwebtext}\hfill
            \subfloat[GPT-2 4.2B]{
			\centering
		\includegraphics[scale=0.32]{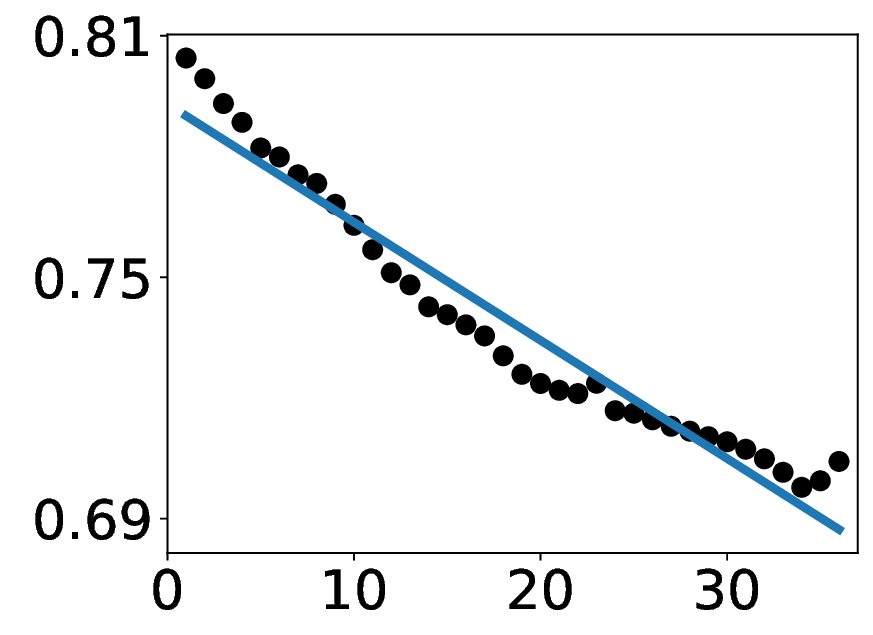}}
            \label{fig:oscar-4.2b-openwebtext}\hfill
     	\subfloat[GPT-2 8.7B]{
			\centering
		\includegraphics[scale=0.32]{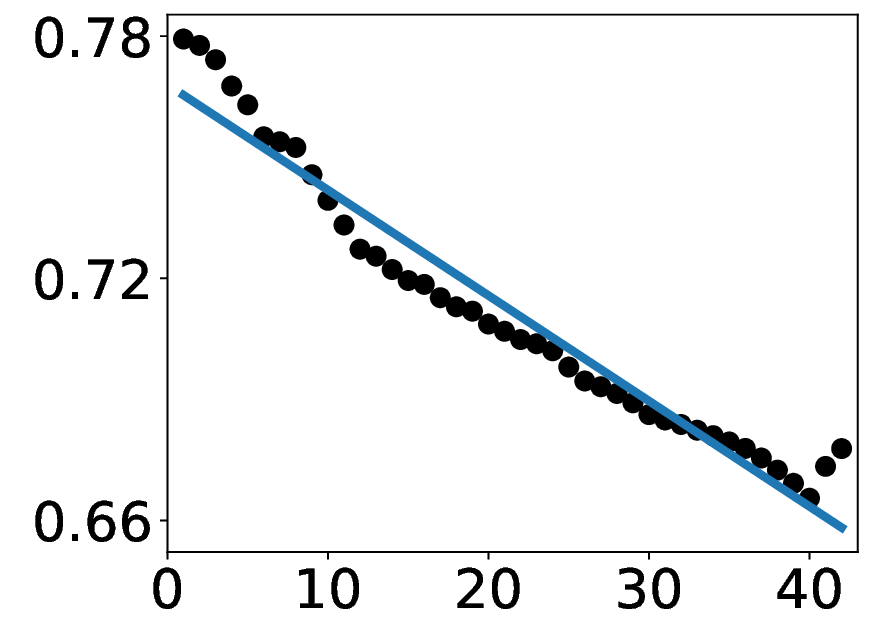}}
            \label{fig:oscar-8.7b-openwebtext}\hfill
		\caption{Pre-training data can affect how the law of equi-learning behaves. The same GPT-2 models pre-trained on C4 and OSCAR are probed with the OpenWebText dataset. The x-axis denotes the layer index, while the y-axis (log scale) shows the prediction residual (PR) as defined in Eq.~\ref{eq:PR}.
		}
		\label{fig:pretraining-data}
\end{figure*}

\textbf{Pre-training data analysis.} As illustrated in Fig.~\ref{fig:pretraining-data}, GPT-2 models pre-trained on two different datasets, C4 and OSCAR, exhibit distinct behaviors regarding the emergence of the law of equi-learning when evaluated on the same probing dataset (i.e., OpenWebText). Specifically, models pre-trained with OSCAR demonstrate more noise in the law's emergence compared to those pre-trained with C4. This discrepancy is likely attributable to the higher noise levels in OSCAR, stemming from its less stringent deduplication. These models were released by \cite{muennighoff2024scaling}. Our findings underscore the significant impact of pre-training data quality on the emergence of the law of equi-learning, suggesting that higher quality pre-training data may result in a more pronounced manifestation of this law.

\end{document}